\documentclass{article}

\PassOptionsToPackage{numbers, compress}{natbib}

 \usepackage[preprint]{neurips_2026}

\usepackage[utf8]{inputenc} 
\usepackage[T1]{fontenc}    
\usepackage{hyperref}       
\usepackage{url}            
\usepackage{booktabs}       
\usepackage{amsfonts}       
\usepackage{nicefrac}       
\usepackage{microtype}      
\usepackage{xcolor}         
\usepackage{graphicx}
\usepackage{amsmath}
\usepackage[percent]{overpic}
\usepackage{comment}
\usepackage{multirow}
\usepackage{float}
\usepackage{caption}
\usepackage{wrapfig}

\title{Rare Events, Real Signals: Functional Ensembles as Units of Computation in Deep Spiking Networks}

\author{%
  \textbf{Aditi Aravind$^{1,2,*}$ \quad
  Konstantinos Ladakis$^{1,2,*}$ \quad
  Mario Alexios Savaglio$^{1,2}$} \\
 \textbf{Stelios M. Smirnakis$^{4, 5}$ \quad
  Maria Papadopouli$^{1,2,3,\dag}$ }\\
  \AND
   $^{1}$University of Crete, Greece \\
  $^{2}$Foundation of Research \& Technology - Hellas (FORTH), Greece \\
    $^{3}$ Archimedes Research Unit, Athena Research Center, Athens, Greece\\
  $^{4}$Harvard Medical School, USA \quad 
  $^{5}$Brigham and Women's Hospital, USA \\ 
  \AND
 $\dag$  Corresponding author: Maria Papadopouli (maria@csd.uoc.gr)\\
  $^{*}$ Equal contribution\\
}

\begin{document}

\maketitle

\begin{abstract}
We investigate how internal representations emerge across hierarchical processing systems by introducing a neuroscience-inspired framework for analyzing deep spiking neural networks (SNN) through the lens of functional connectivity. Drawing on concepts from systems neuroscience and information theory, we form the first-order functionally connected (1FC) group of a neuron based on its statistically significant pairwise correlations with neurons from the previous layer of a trained SNN architecture. We then track its response properties during inference under various conditions. Our analysis shows that several principles of functional connectivity previously observed in biological cortex are preserved in spiking ResNet architectures. These 1FC ensembles display interesting properties: their aggregate cofiring reliably predicts downstream neuronal responses through a robust, ReLU-like input–output relationship, whose gain scales systematically with ensemble size. Reliable encoding of the presented class emerges only during high 1FC cofiring events, which themselves occur infrequently, indicating that informative representations are concentrated in rare but highly coordinated activity patterns. Under uniform random noise or adversarial perturbations, these response profiles are disrupted, particularly in early and intermediate layers. This enables a targeted high-resolution interrogation at specific nodes and pathways. We showed that the functional connectivity structure is shaped by learning and this structure breaks under weight permutation. These establish 1FC ensembles as a functionally meaningful substrate for input encoding and information transfer, with potential implications in designing targeted fine-grained diagnostics on the information flow.
\end{abstract}
\vspace{-0.1in} 
\section{Introduction}
\vspace{-0.1in} 
Neural networks are not merely fitting data or minimizing training error; they are learning structured internal representations, exhibiting consistent patterns across tasks and scales \cite{simon2026there, huh2024platonic, bau2017network, kornblith2019similarity, bengio2013representation}.
However understanding how 
representations emerge across hierarchical processing systems remains a central challenge in computational neuroscience and machine learning.  
Inspired by systems neuroscience, we study this question in deep spiking neural architectures, with an emphasis on the functionally emergent connectivity.
In neuroscience, both spontaneous activity—observed in the absence of strong sensory input—and stimulus-evoked responses exhibit structured, non-random patterns, indicating the presence of organized collective dynamics.
Neuronal ensembles—groups of neurons that \textit{co-activate more frequently than expected by chance}—are increasingly viewed as fundamental units of cortical computation, enabling efficient encoding and transmission of shared information. A substantial body of work has identified structured patterns in both spontaneous and evoked population activity, supporting this ensemble-based view of neural processing \cite{de1949cerebral, Meister1991, Abeles1993, Grinvald2003, hebb1949first, Ringach2009, CarrilloReid2016, panzeri2022structures, averbeck2006neural, yuste:2015, Cossart:2003Attractor, PMID:25201983, russo2017cell, Yuste2024}. 
In neocortex, cortico-cortical connectivity is weak, necessitating that ensembles of neurons coordinate to form computational units. 
Functional connectivity analysis offers a principled approach to identifying such ensembles, revealing networks of neurons linked through shared anatomical and statistical dependencies, forming canonical elements of information processing \cite{Kenet2003, Luczak2009, Palagina282988,Papadopouli2024,Savaglio25:Direction,Savaglio25:Temporal}. 
Building on this idea, we have proposed that all neurons of a layer in the visual cortex with statistically significant pairwise correlations with a \textit{target neuron} compose its \textit{first-order functionally connected (1FC)} ensemble in that layer, forming elementary cortical computational units (see attached manuscript in supplement). We then characterized their properties and couplings.
%
%
%
This framework yields mechanistic insight into cortical circuit organization, complementary to structural connectivity,  helping us also to bridge the study of cortical circuit organization with deep-learning architectures for assessing neuro-computing hypotheses.

Spiking neural networks (SNNs) consist of layers of simplified (computational) neurons, that mimic the hierarchical, integrative properties and sparse firing of biological circuits\cite{mathis2024:Decoding,tavanaei2019deep}.
Here we apply this framework to study the structure of these representations 
across the processing hierarchy in SNNs. 
This framework enables a targeted high-resolution interrogation, placing  "magnifying lens" at critical nodes and pathways within the architecture, \textit{during inference}, offering direct insight into the information flow.
%
Unlike conventional interpretability approaches that summarize model behavior through attribution maps/scores \cite{Achtibat2023:Attribution,Dha18,Bau20}, feature visualizations \cite{ola18, ola20}, mechanistic approaches like causal tracing \cite{meng2022ROME}, or post hoc saliency analyses \cite{Zho15,Sel16c,Des20}, this framework provides a node-level, pathway-specific view of computation. Rather than asking only which input features or units are important, it identifies \textit{when} and \textit{through which functional routes} information is propagated within the architecture during inference.

\textbf{Innovation}: We apply systems network neuroscience and information theory to reverse engineer the structure of learned  \textit{circuit} representations during inference.
We do this by studying: \textbf{1)} the architecture of 1FC ensembles that emerge from identifying significant pairwise functional connections between neurons in consecutive layers during testing, and \textbf{2)} the \textit{response properties} of these ensembles across layers that shape the information flow during inference, and how they change under different input testing conditions.
While existing interpretability approaches predominantly analyze individual neuron activations or population-based layer-wise representations, this work takes a distinct approach by examining \textit{functional} connectivity based on \textit{pairwise correlations} of neurons between \textit{consecutive layers}, to characterize the structure of information flow in SNNs.

\textbf{Contributions:} Although, we do not intend to claim a precise correspondence between SNNs and visual cortex, which have substantial differences \cite{Bowers2023}, we find it interesting that several principles of the functional connectivity identified in mouse visual cortex (see attached manuscript in supplement) are conserved in SNNs. 
Both uniform random noise and adversarial perturbations induce changes in these response properties, with the most significant impact observed in early and intermediate processing layers. 
This provides an opportunity for targeted, high-resolution probing of specific nodes and pathways, particularly in regimes characterized by strong 1FC co-firing events.
As expected, functional connectivity architecture emerges during learning and degrades markedly under weight permutation. Taken together, these observations support the interpretation of 1FC ensembles as a functionally relevant substrate for input encoding and information transmission, enabling targeted fine-grained diagnostics on the information flow. 

\textbf{Limitations:} So far we tested this approach in
ResNet18 SNNs and spiking-neural units of spiking-layers that extend a ResNet32 feature extractor\cite{He2016:Deep}, pre-trained on CIFAR100\cite{CIFAR100} and ResNet9 SNNs trained on DVS Gesture Dataset. We also have only tested the framework for the Fast Gradient Sign Method (FGSM) adversarial attack\cite{Goodfellow2014ExplainingAH} and additive noise. It is part of our ongoing effort to test it with other architectures, activation functions, and adversarial conditions and analyze its diagnostic power in efficiently detecting anomalies using test datasets of different size, richness, and noise levels. 
Going forward, integrating this framework with causal perturbations will be key to clarifying the mechanistic role of specific ensembles in learning, as well as to probing how these mechanisms fail under adversarial conditions. 
\section{Related Work} \label{sec:rel}
\vspace{-0.12in}
Research on internal network interpretability in ANNs and SNNs has developed along several connected lines: probing intermediate representations for class-relevant information, comparing representational geometry across layers and models, attributing predictions to spatial regions or channels, and identifying influential hidden units \cite{Klabunde2025-sh, pathpatching, somvanshi-26}. 
One approach treats intermediate layers as objects to be probed. \citet{Ala16} showed that linear classifier probes can track where task-relevant information becomes linearly separable across depth, turning internal representations into diagnostic objects. This framework summarizes a layer by its ability to decode a target class, as opposed to the internal organization of neurons that support it. Representational comparison methods enable cross-layer and cross-model comparisons. For example, SVCCA \cite{Rag17} enabled comparisons by factoring out irrelevant basis changes, while \citet{kornblith2019similarity} identified limitations of CCA-based approaches in high-dimensional settings and advocated for centered kernel alignment as a more reliable statistic. These methods work as tools for investigations that remain whole-layer summaries; they characterize what a representation contains, not how its constituent neurons are functionally organized.

A parallel branch explains predictions through class-discriminative spatial attribution. CAM \cite{Zho15} demonstrated that class-specific weights can be projected onto convolutional feature maps to identify decision-relevant image regions. Grad-CAM \cite{Sel16c} generalized this using gradients into any chosen convolutional layer, removing architectural constraints. Ablation-CAM \cite{Des20} proposed a gradient-free alternative based on feature-map ablation, reducing sensitivity to gradient pathologies. These methods make class importance spatially visible and are central in visualization of unit receptive fields, but they rely on predictions of models to ascertain feature representation of individual nodes or units.
Previous work on information flow within the network concerns tracking the activations of the nodes. For example, \citet{Wan18b} identifies critical data paths which are learnt by gating over channels to recover the semantically selective pathway most responsible for a prediction. This frames explanation in terms of internal pathways rather than input saliency. At the unit level, conductance \cite{Dha18} tracks how attribution flows through individual neurons, outperforming raw activation measures. \citet{Bau20} showed that ablating small subsets of class-aligned units can strongly impair class-specific performance suggesting that relevant structure is often concentrated in ensembles rather than isolated neurons. For SNNs, \citet{Kim21c} adapted the traditional visual explanation through Spike Activation Maps based on inter-spike intervals, providing a gradient-free, SNN-native attribution baseline, though still focused on individual neuron salience over time.\\
While existing interpretability approaches primarily focus on individual neuron activations or layer-wise representations, we instead examine functional connectivity through pairwise correlations between neurons in consecutive layers, enabling a structured characterization of information flow in SNNs.
\section{Functional Connectivity Methodology} \label{sec:methods}
 \vspace{-0.12in}
\textbf{Network Neuroscience Background.} We recently found that biological neurons, along with their functionally-connected neighbors form elementary multi-neuronal ensembles that act as putative information-processing “primitives” both within and across granular and supragranular layers in mouse primary visual cortex\cite{Papadopouli2024}. Such ensembles, being flexible and multiplexing, may support task-specific processing, adaptation, and learning through dynamic reconfiguration of their interactions.
%
We found that the firing probability of layer 2/3 ("putative output") neurons in mouse primary visual cortex exhibits a ReLU-like nonlinearity, emerging when 13\% of L4-1FC “putative inputs” co-fire, yielding sparse yet reliable responses. Moreover, their responses depend on the count (N), not the identity, of co-active L4-1FC partners, with response sensitivity scaling as a power law in N. 
The corresponding L4-1FC and L2/3-1FC ensembles form information pathways, a view reinforced by their shared representational content.
Our approach is inherently local and contrasts with prior methods that use regression or dimensionality reduction to model inter-areal communication, behavior, or stimulus encoding \cite{kohn2016correlations,Stringer2019,Semedo2019,See2018}. 
Recent advances have mapped anatomical connectivity with unprecedented resolution \cite{MICrONS2024}.
However, structural connectivity alone does not specify 
how functional ensembles interact across cortical layers. In contrast, our approach focuses on functional connectivity, capturing coordinated, time-varying interactions among specific neurons to reveal ensemble organization at single-cell resolution. 


\textbf{Deep Learning Spiking Model}. 
To address our objective to study how spiking representations emerge across hierarchical processing, we selected the \textit{convolutional ResNet18} that supports both layer-wise analysis, biologically meaningful inductive biases, and practical considerations (e.g., sufficient depth and computational tractability).
Moreover, the hierarchical stages of CNNs and ResNet have been extensively mapped onto the ventral visual stream in prior work\cite{Yamins2016-ry, Wen2018-ho}, further supporting the use of SEW-ResNet as a suitable model system for probing the development of spiking representations. We focused on the main residual branch through successive layers.
We analyzed the spiking responses of a convolutional SNN trained on CIFAR100 \cite{CIFAR100}, using the Spike-Element-Wise (SEW) ResNet\cite{Fang_21_SEW}. The SEW-ResNet employs a Parametric Leaky Integrate-and-Fire (PLIF) neuron with a learnable membrane time constant \cite{Fang_2021_ICCV}, and is structured as 8 residual blocks each containing two convolutional spiking layers in the order of Convolution→Batch Normalization(BN)→Spiking Neuron (CONV$\rightarrow$BN$\rightarrow$PLIF), yielding 16 feature-extracting layers in total. 
Final representations are aggregated via adaptive average pooling across spatial dimensions, before being passed to a linear classification head. The resulting logits are averaged across the $T=4$ timesteps (chosen as per \cite{Fang_21_SEW}), and a single softmax and loss computation is performed on this averaged output to produce the final class prediction.
As the architecture was originally designed for ImageNet, we adapt the stem convolution from a $7\times7$, 128-channel configuration to a $3\times3$, 64-channel kernel to match CIFAR100 inputs. Crucially, the network operates at a mean firing rate of $\sim 0.072$ spikes per sample (i.e. image) per neuron (Fig. \ref{fig:firing_rates_SNN}) which is consistent with the sparse activity observed in the biological visual cortex and thus provides a biologically plausible regime in which to study sparse functional structure. 
The SEW-ResNet-18 comprises 11.2 million parameters and achieves 72.18$\%$ accuracy when trained and evaluated on CIFAR100 (Sec. \ref{sec:SNN_arch_train}).  

\textbf{Image Presentation and Recorded Spike Trains.} Each image is presented to the network $T=4$ times within a single pass, both during training and inference stages; the network produces one classification output per image, with the temporal repetition allowing neurons to accumulate sufficient spiking activity for reliable encoding rather than generating independent predictions at each timestep. 
All results below are computed on a subset of neurons from each examined layer. Approximately 7,500 - 9000 neurons are sampled per layer, drawn across layers while restricting sampling to a central spatial region for each channel. In the final four convolutional spiking layers (corresponding to the last two blocks), each containing 8,192 neurons, all neurons are included in the analysis. For earlier layers, only neurons located in the central region are selected. 

\textbf{Spike Trains}. Passing the full CIFAR100 test set (10,000 images) through the network yields, for each neuron, a binary spike train of length 40,000—-\textit{four spike- or no-spike responses} per image—-with each neuron's activity arranged spatially as a tensor of shape $(C_l, H_l, W_l)$ corresponding to channel, height, and width at layer $l$. These spike trains will be used in the pairwise correlation analysis for the identification of the 1FC groups. (These pairwise correlations are the equivalent signal correlations.)
The CIFAR100 images, with no distortions, will be called \textit{original input}. 
Throughout this paper, the membership of 1FC remains fixed, as computed under the reference testing set ("original input"). 


\textbf{Pairwise Correlations}. The pairwise correlations between neurons are calculated using the spike time tiling coefficient (STTC)\cite{cutts2014detecting}, a symmetric measurement that is bounded in [-1, 1] (Sec. \ref{sec:A_STTC}). 
STTC was selected because it is largely
independent of firing rate and performs favorably relative
to thirty-three other commonly used correlation measures.
STTC is defined with respect to a measurement window of synchronization $\Delta t$. 
Since in SNNs the exact response timing is known, we set $\Delta t=0$. 
\begin{figure}[h]
  \centering
  \begin{overpic}[width=\linewidth]{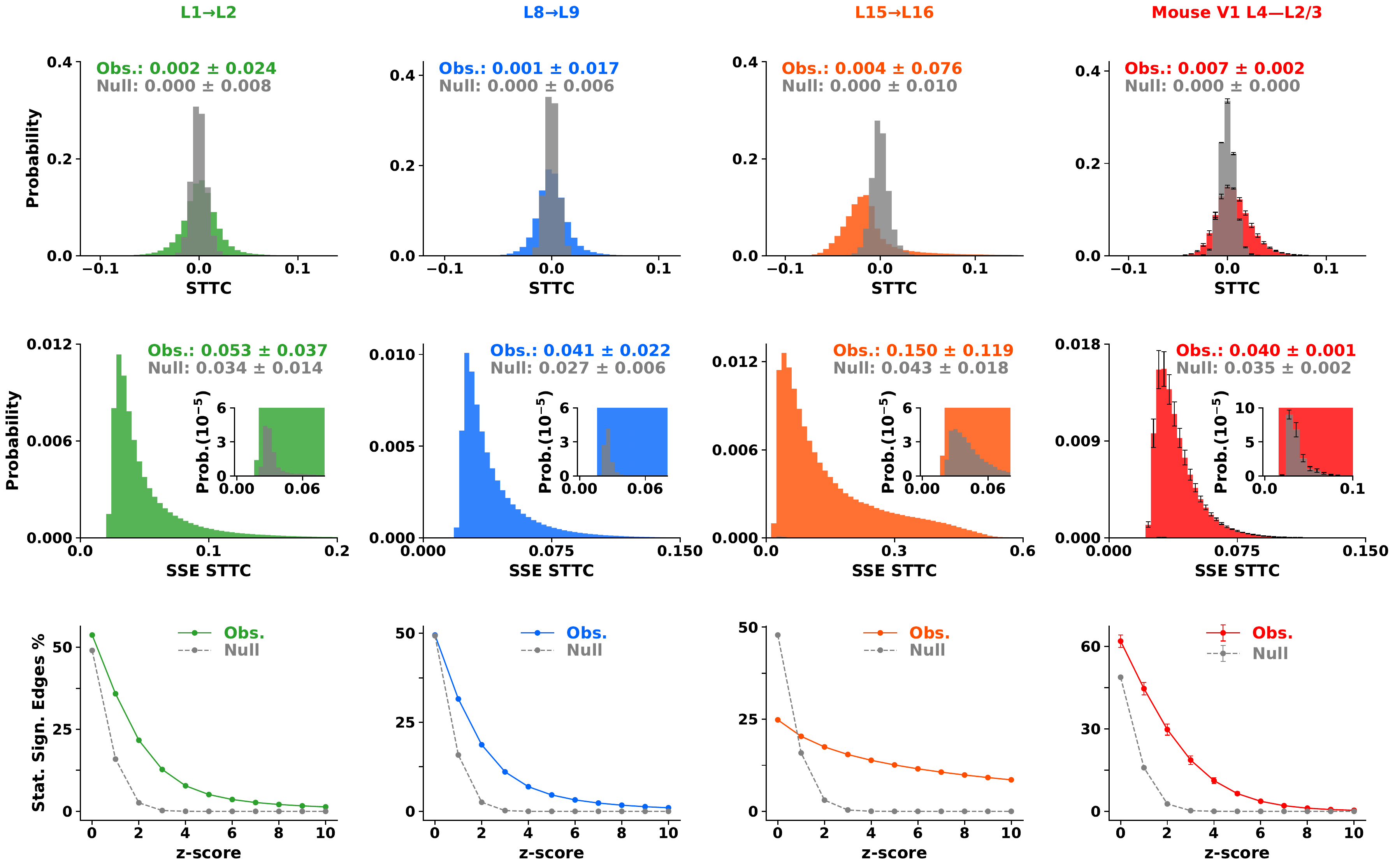}
  \put(2,60){\textbf{A}}
  \put(27,60){\textbf{B}}
  \put(51,60){\textbf{C}}
  \put(75,60){\textbf{D}}
  \put(2,39){\textbf{E}}
  \put(27,39){\textbf{F}}
  \put(51,39){\textbf{G}}
  \put(75,39){\textbf{H}}
  \put(2,18){\textbf{I}}
  \put(27,18){\textbf{J}}
  \put(51,18){\textbf{K}}
  \put(75,18){\textbf{L}}
  \end{overpic}
  \caption{\textbf{Inter-Layer Functional Connectivity through Pairwise STTC}. Pairwise STTC was computed between the spiking activity of neurons in consecutive layer pairs during inference over the test set of CIFAR100 dataset ("original" data). Columns correspond to layer pairs L2$\rightarrow$L3 (green), L9$\rightarrow$L10 (blue), and L16$\rightarrow$L17 (orange); The final column shows the same analysis in mouse primary visual cortex (V1), computed across cortical layers L4 to L2/3 and aggregated over 5 mice (red). For each panel, a separate null distribution (grey) was computed by randomly circular shifting the spiketrains 150 times. \textbf{(A–D)} Distributions of pairwise STTC values overlaid with null distributions derived from circularly shifted controls. \textbf{(E–H)} Distributions of statistically significant positively correlated edges, defined by a z-score threshold of $z > 4$. \textbf{(I-L)} Percentage of statistically significant edges as a function of z-score. For the mouse data, histograms reflect the distribution of mean values across mice, with error bars denoting SEM; z-scores for all data were computed using the mean and standard deviation of the circular-shift null distributions. The pairwise correlations are weak but statistically significant.}
\label{fig:STTC_PercSSE}
\end{figure}
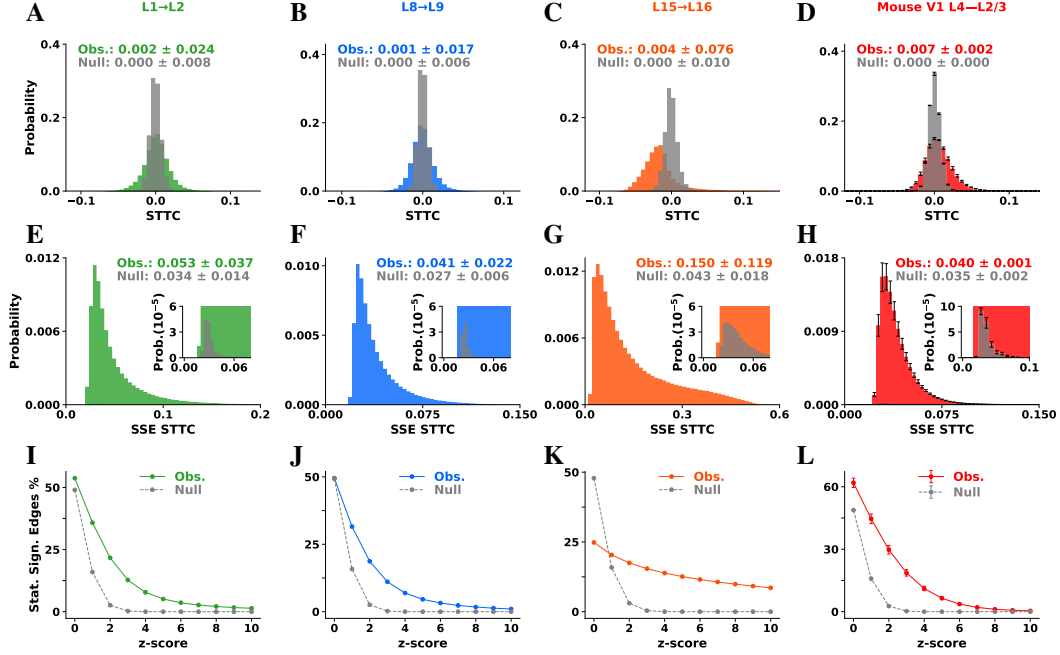

\textbf{First-Order Functional-Connectivity (1FC) Groups}. 
To extend the analysis beyond pairwise correlations, a natural next step is to consider the graph formed by a node (neuron) and its \textit{statistically significantly} correlated (functionally connected) neighbors at \textit{one degree of separation}. We refer to this set as the neuron’s \textit{first-order functional connectivity (1FC)} group. The 1FC group thus represents the simplest multi-neuronal ensemble beyond a single neuron and therefore provides a principled starting point for higher-order analysis. We summarize the extent of each neuron's participation in functional interactions via its degree of connectivity (DoC), i.e., the size of its 1FC group. Here we focus on the 1FC group of a neuron at layer $L$, based on its \textit{positive} \textit{statistically significant functional connections} with neurons from the \textit{previous }layer ($L-1$), in a ResNet architecture trained on CIFAR100. The 1FC groups identified at testing using the original dataset becomes our functional connectivity reference throughout the paper.
%

\textbf{Response Function of a Neuron.}
We computed the response of a neuron as a function of the aggregate co-firing within its input 1FC group. We then examine how aggregate co-firing within an input 1FC group relates to the activity of its downstream "output" neuron. We characterize this response function and examine its behavior under different conditions. The membership of its 1FC group remains fixed (as computed under the original test). The spike trains that will be used for estimating the response will be the ones recorded under the corresponding test input. Thus, the response of a neuron may change under different conditions (despite keeping the membership of its 1FC membership as the one identified during original input--no distortion).

\textbf{Informativeness of a Neuron.} To quantify the informativeness of individual neurons, we computed the mutual information (MI) between each neuron's spike train and the true class labels across the test set. 
MI was calculated between the spike counts (number of spikes out of a maximum possible 4) per image and both the categorical class labels and their one-hot encoded representations, the latter yielding a class-wise MI profile for each neuron.
MI is defined as 
$I(X;Y) = H(X) - H(X|Y),$
where $H(X)$ denotes the Shannon entropy of the neuron's spiking activity and $H(X|Y)$ is the conditional entropy given the class label, such that MI reflects the reduction in uncertainty about a neuron's firing upon observation of the true class. 

\textbf{Distorted Image Presentation.}
For analysis under \textit{additive noise}, each test image is corrupted with noise (Sec.\ref{sec:add_noise}), prior to inference, and the resulting spiking activity is recorded and analyzed equivalently. For the \textit{adversarial perturbation} condition specifically, analysis is conducted on a representative subset of 30 classes, owing to the computational overhead of per-sample gradient computation required by FGSM (Fig.\ref{fig:noise_example_images}). The spike trains produced under these image presentations will be used for computing the response of the neurons under distorted image presentation, keeping however the membership of their 1FC groups that was computed on the original image set.

\textbf{Exclusion Criteria.} Neurons with a mean firing rate below 0.01 spikes per sample are designated as \textit{silent} and are excluded from all analyses ($\sim 9.5 \%$ excluded per layer). Of the sampled neurons per layer, approximately $90.5\%$ meet the activity threshold and are designated as \textit{active} neurons; all subsequent analyses are conducted exclusively on this active population (See Fig. \ref{fig:firing_rates_SNN}).
\vspace{-0.15in}
\section{Results}\label{sec:res}
 \vspace{-0.2in}
\textbf{Statistically significant but weak pairwise correlations.} We first analyze the functional connectivity of the model on the original test set. We observed significant functional coupling between neurons of each layer of the network as well as across source-target layer pairs (Figs \ref{fig:STTC_PercSSE}). Observed pairwise STTC values extended well-beyond their corresponding null distributions in both positive
and negative directions, with a stronger positive tail (Fig. \ref{fig:STTC_PercSSE}, A-C), with the structural edges contributing to the tails (Fig. \ref{fig:ALL_STR_nSTR} A-F) at the first layers. At a conservative threshold of z-score>4, approximately $7.8\%$ of neuronal pairs from L1$\rightarrow$L2 exhibit significant functional correlations compared to $\sim7\%$ in L8$\rightarrow$L9 and $\sim 13.9\%$ edges from L15$\rightarrow$L16 (Fig.\ref{fig:STTC_PercSSE} I-L). 

\begin{figure}[H]
\centering
\begin{tabular}{cccc}

\begin{overpic}[width=0.22\linewidth]{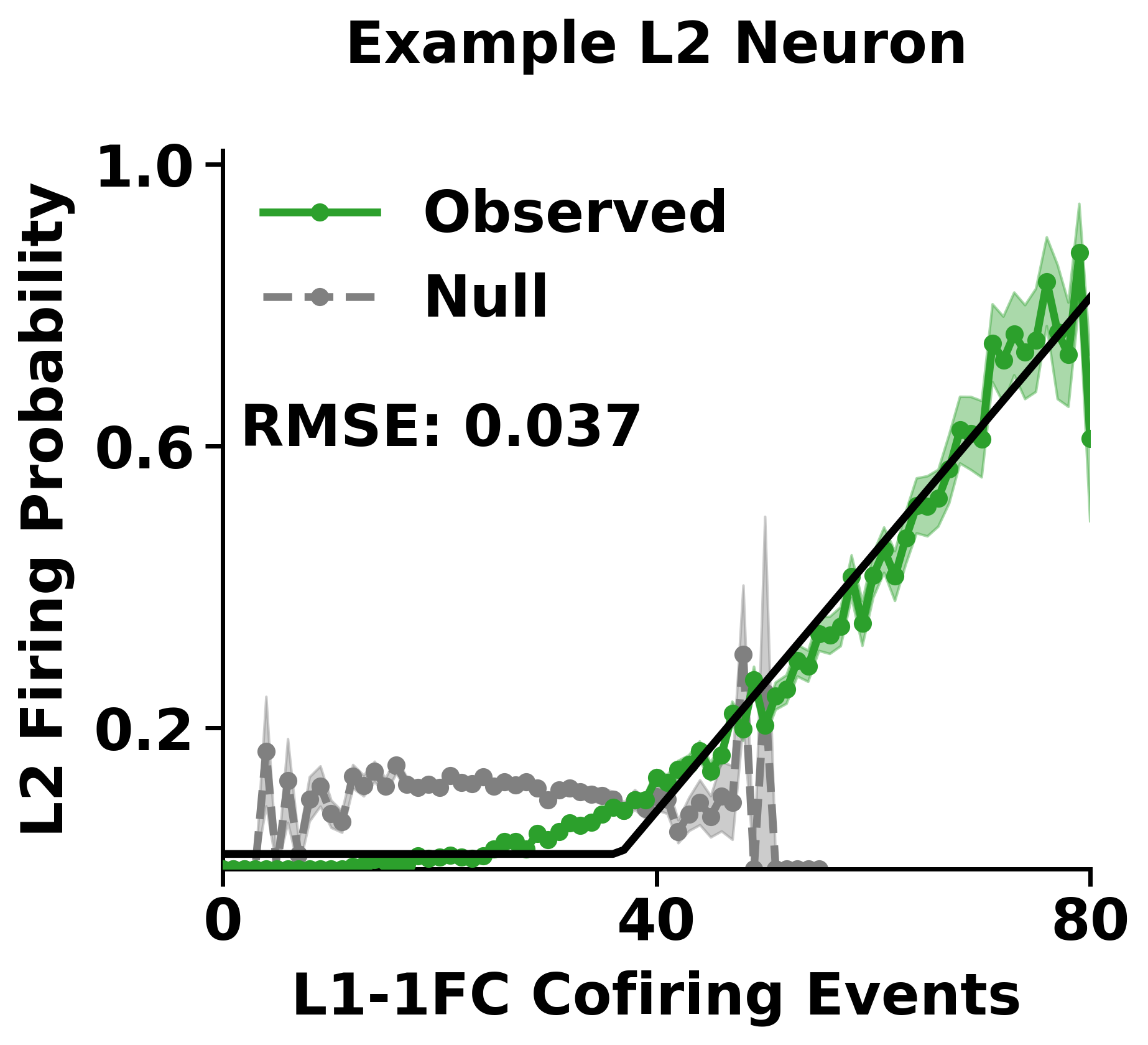}
\put(2,90){\textbf{A}}
\end{overpic} &
\begin{overpic}[width=0.22\linewidth]{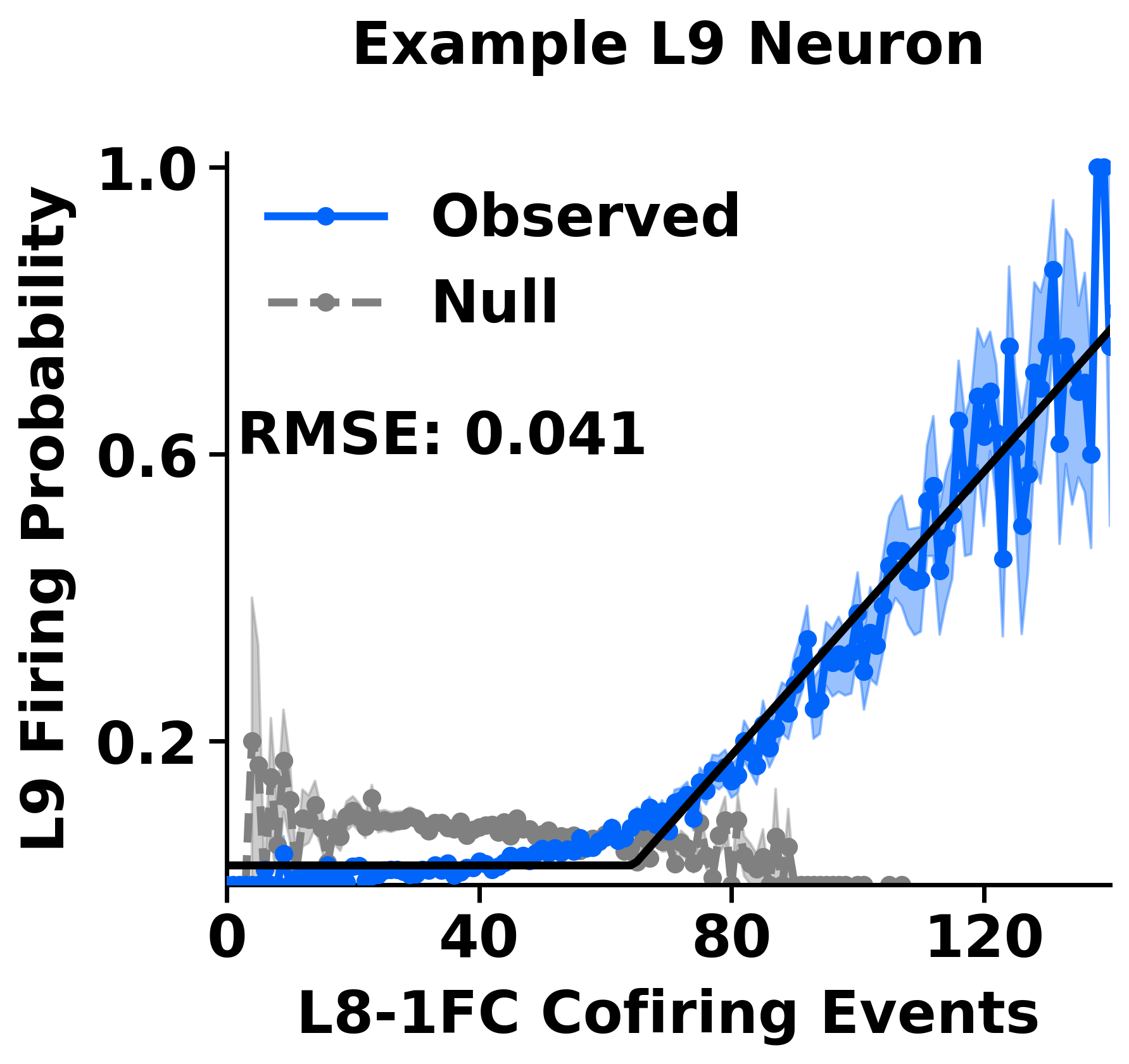}
\put(2,90){\textbf{B}}
\end{overpic} &
\begin{overpic}[width=0.22\linewidth]{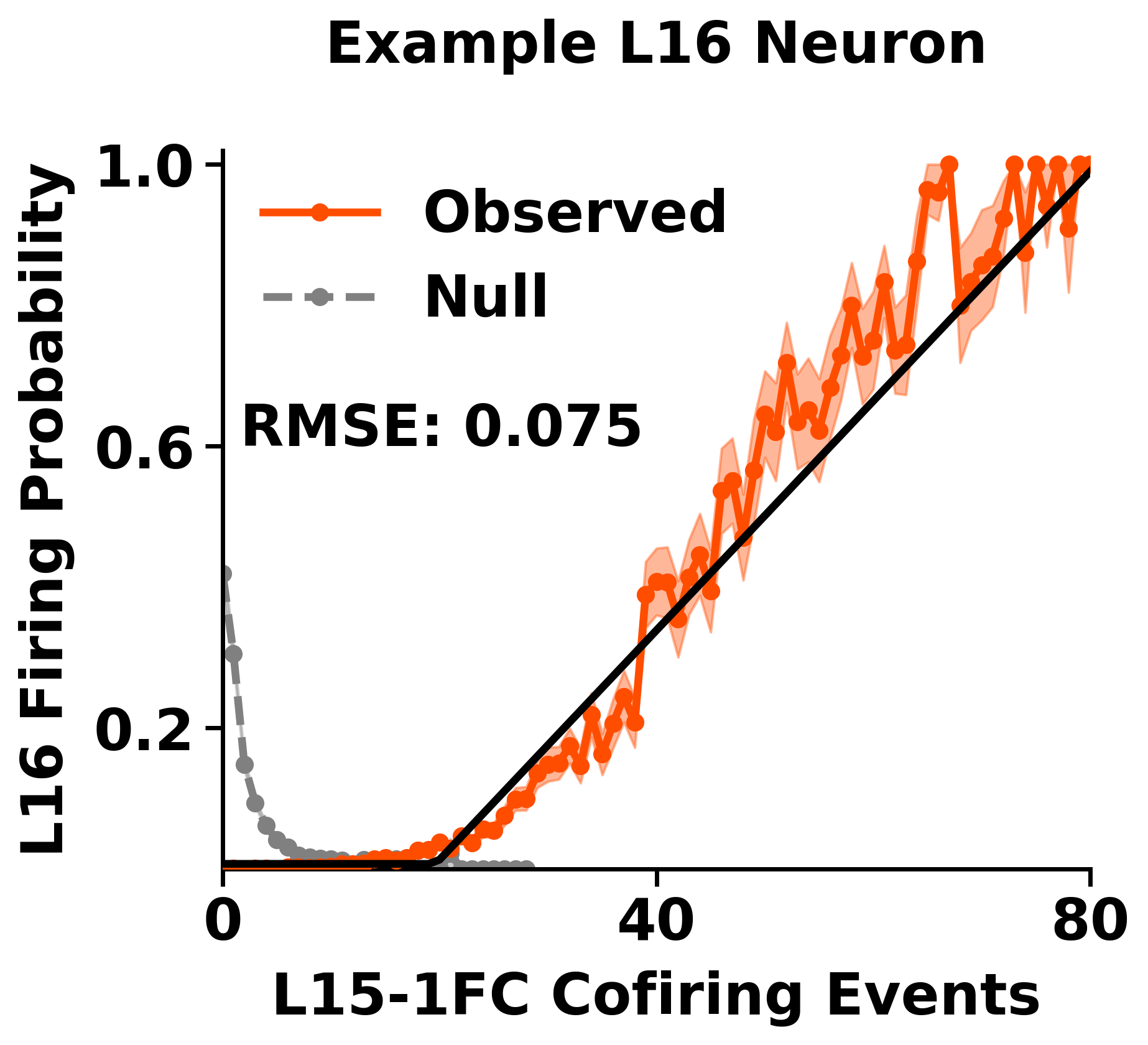}
\put(2,90){\textbf{C}}
\end{overpic} & 
\begin{overpic}[width=0.22\linewidth]{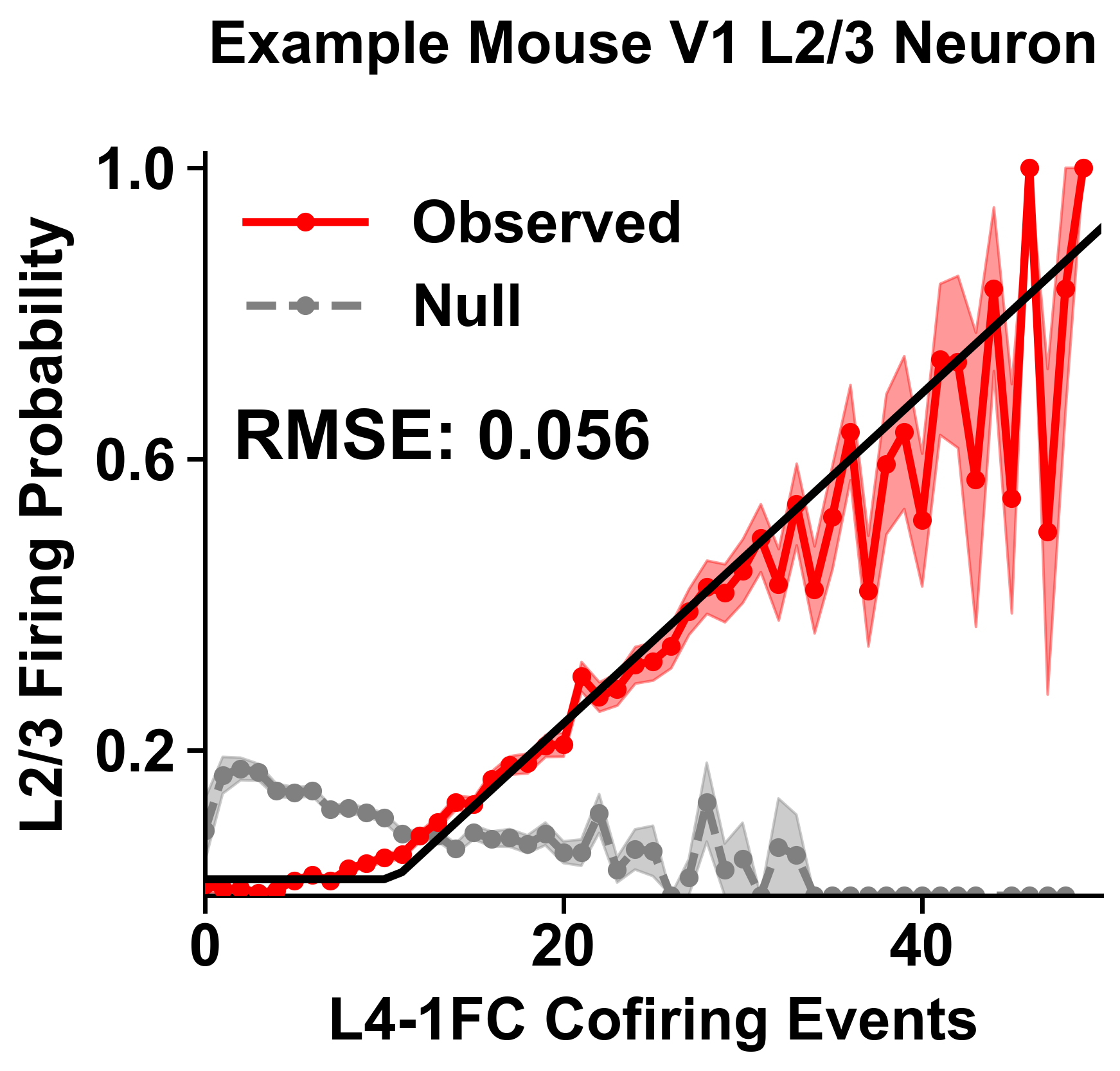}
\put(2,90){\textbf{D}}
\end{overpic} \\

\begin{overpic}[width=0.22\linewidth]{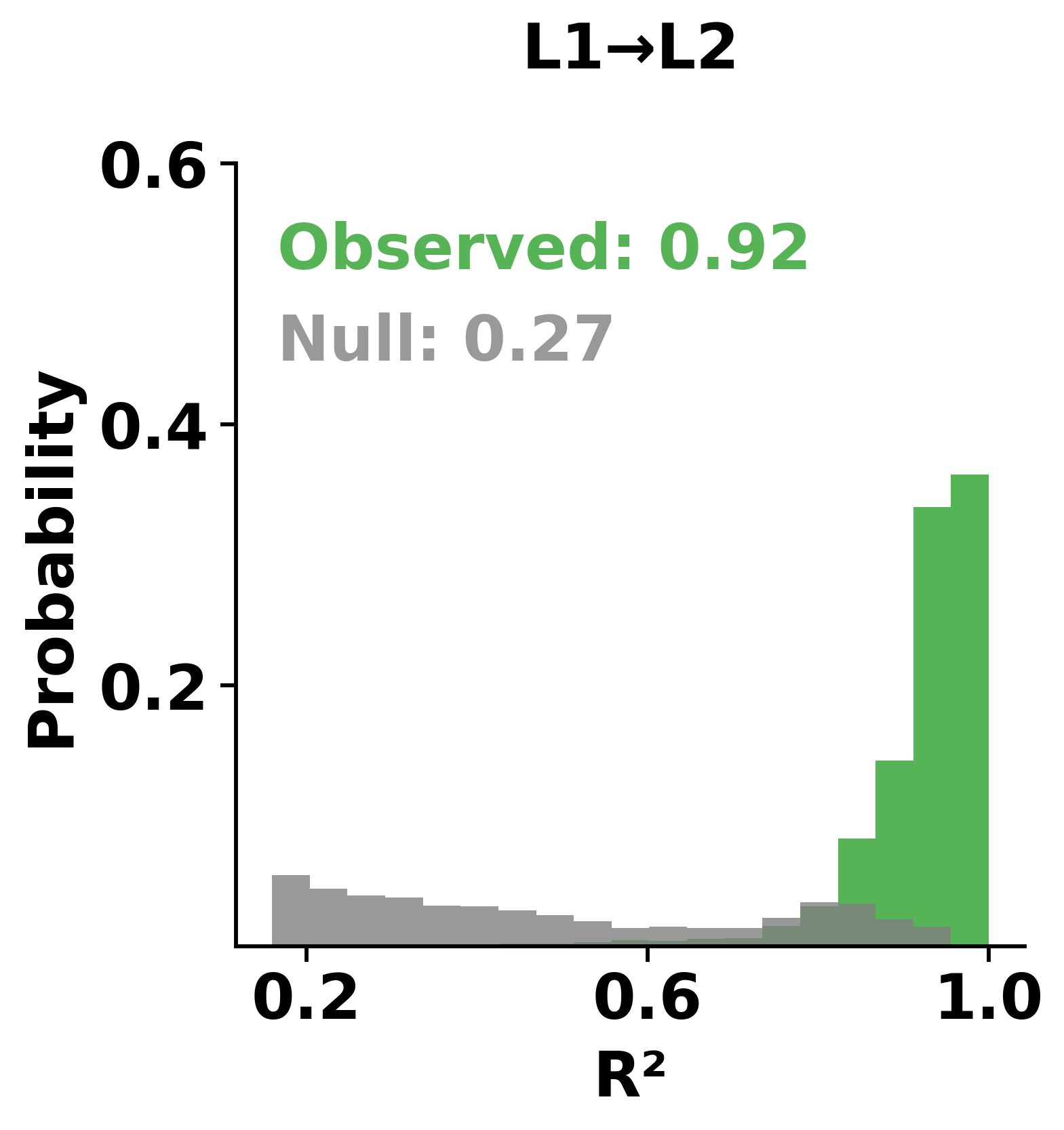}
\put(2,90){\textbf{E}}
\end{overpic} &
\begin{overpic}[width=0.22\linewidth]{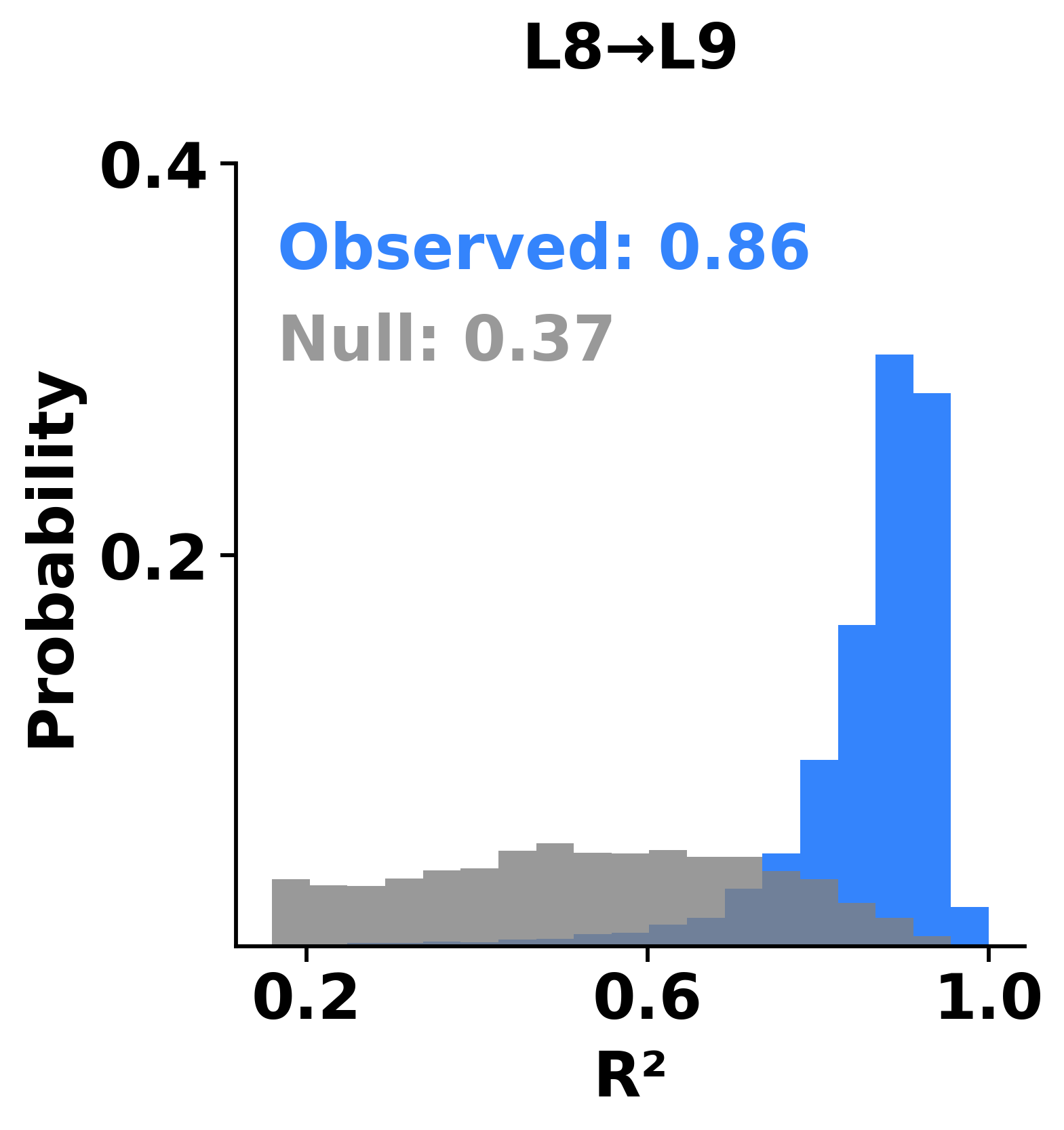}
\put(2,90){\textbf{F}}
\end{overpic} &
\begin{overpic}[width=0.22\linewidth]{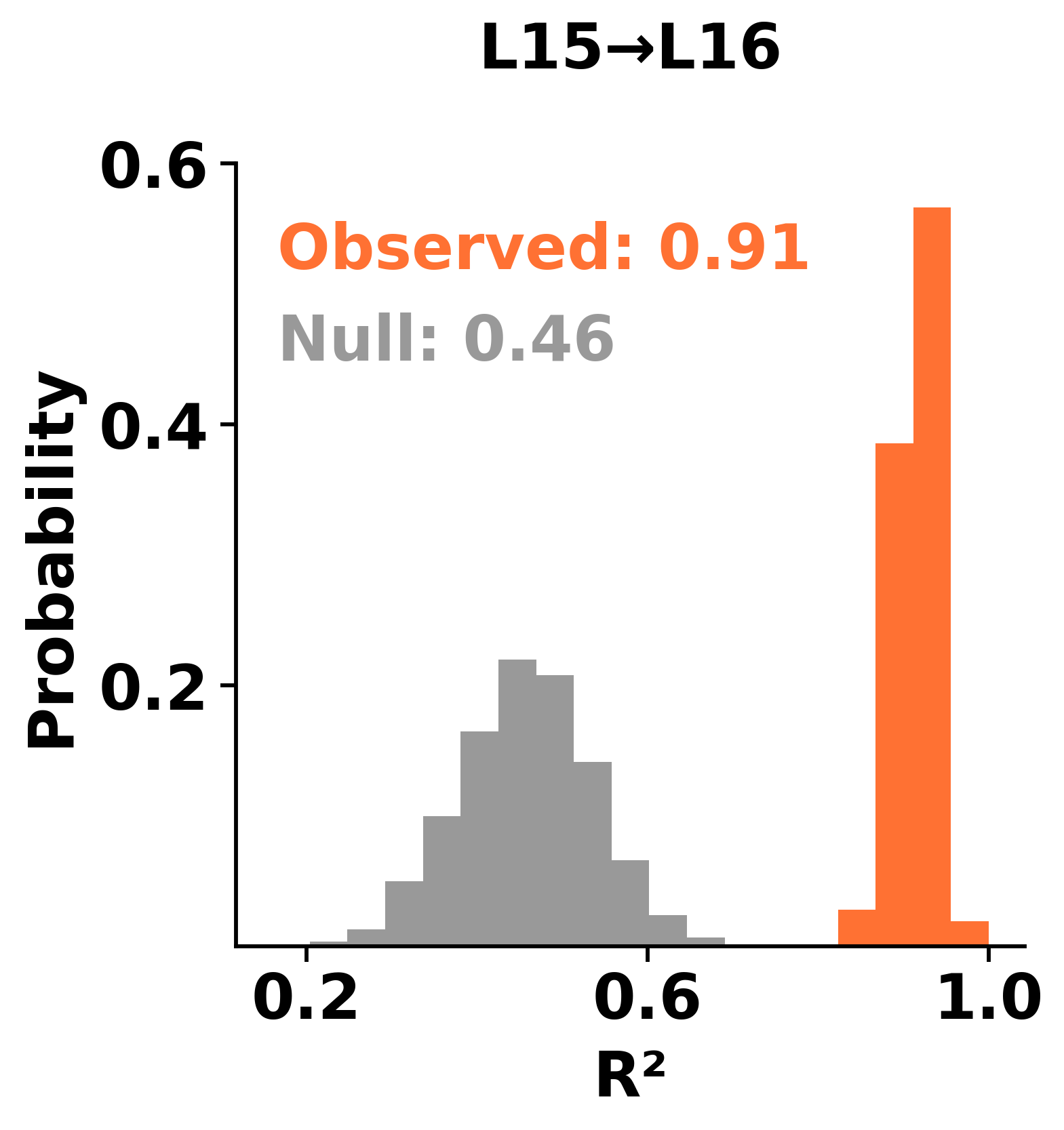}
\put(2,90){\textbf{G}}
\end{overpic} &
\begin{overpic}[width=0.22\linewidth]{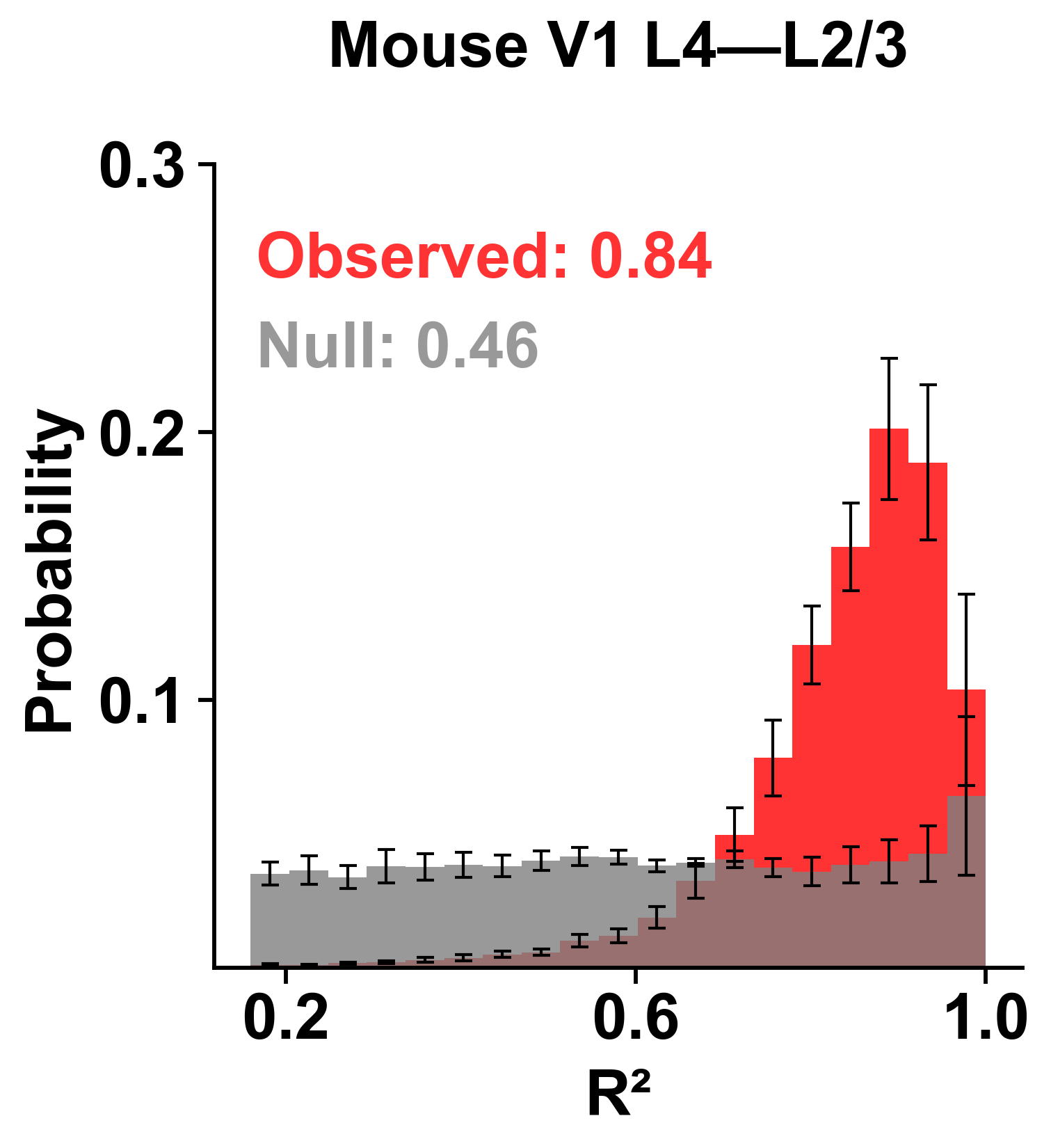}
\put(2,90){\textbf{H}}
\end{overpic} \\

\begin{overpic}[width=0.22\linewidth]{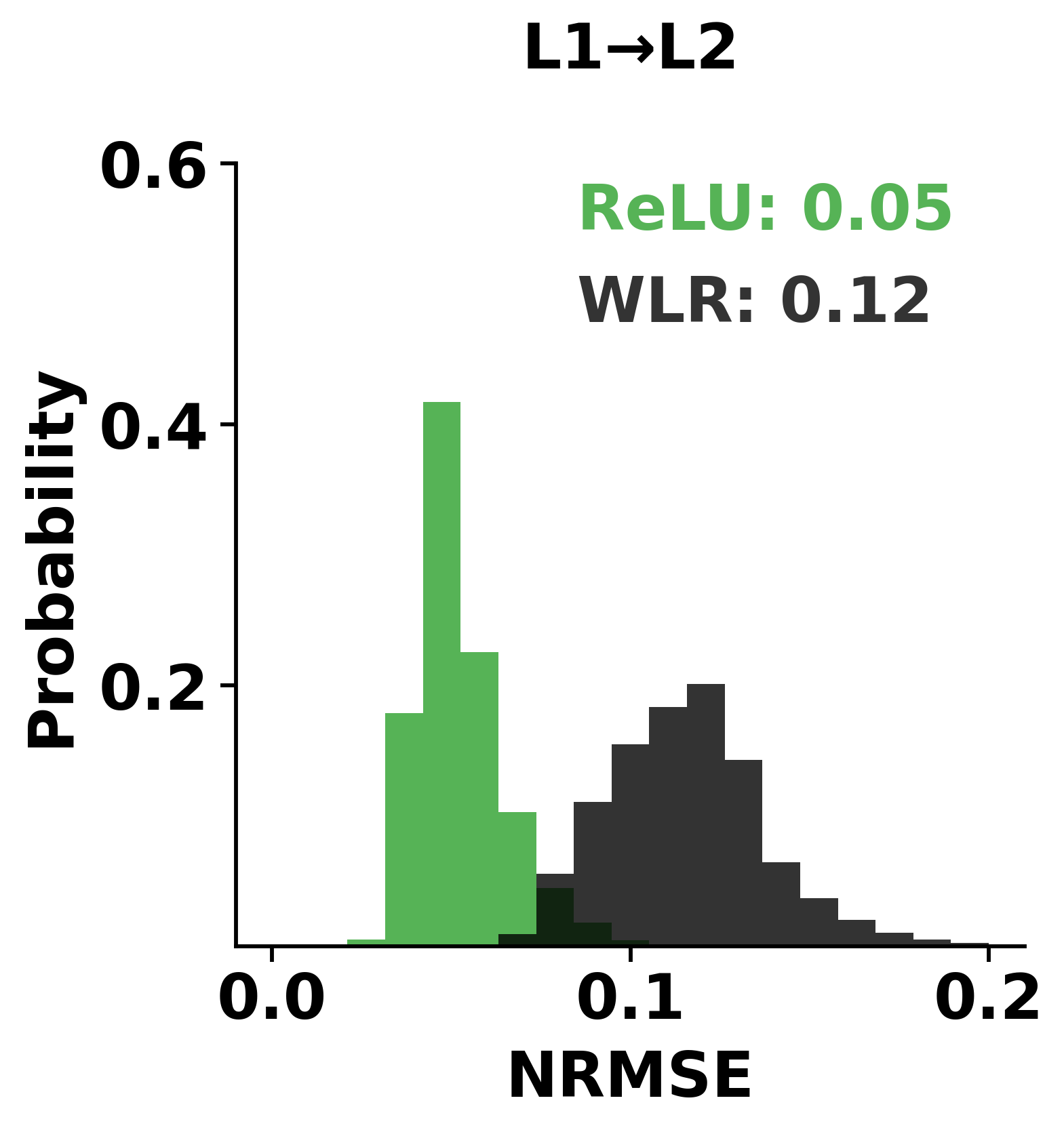}
\put(2,90){\textbf{I}}
\end{overpic} &
\begin{overpic}[width=0.22\linewidth]{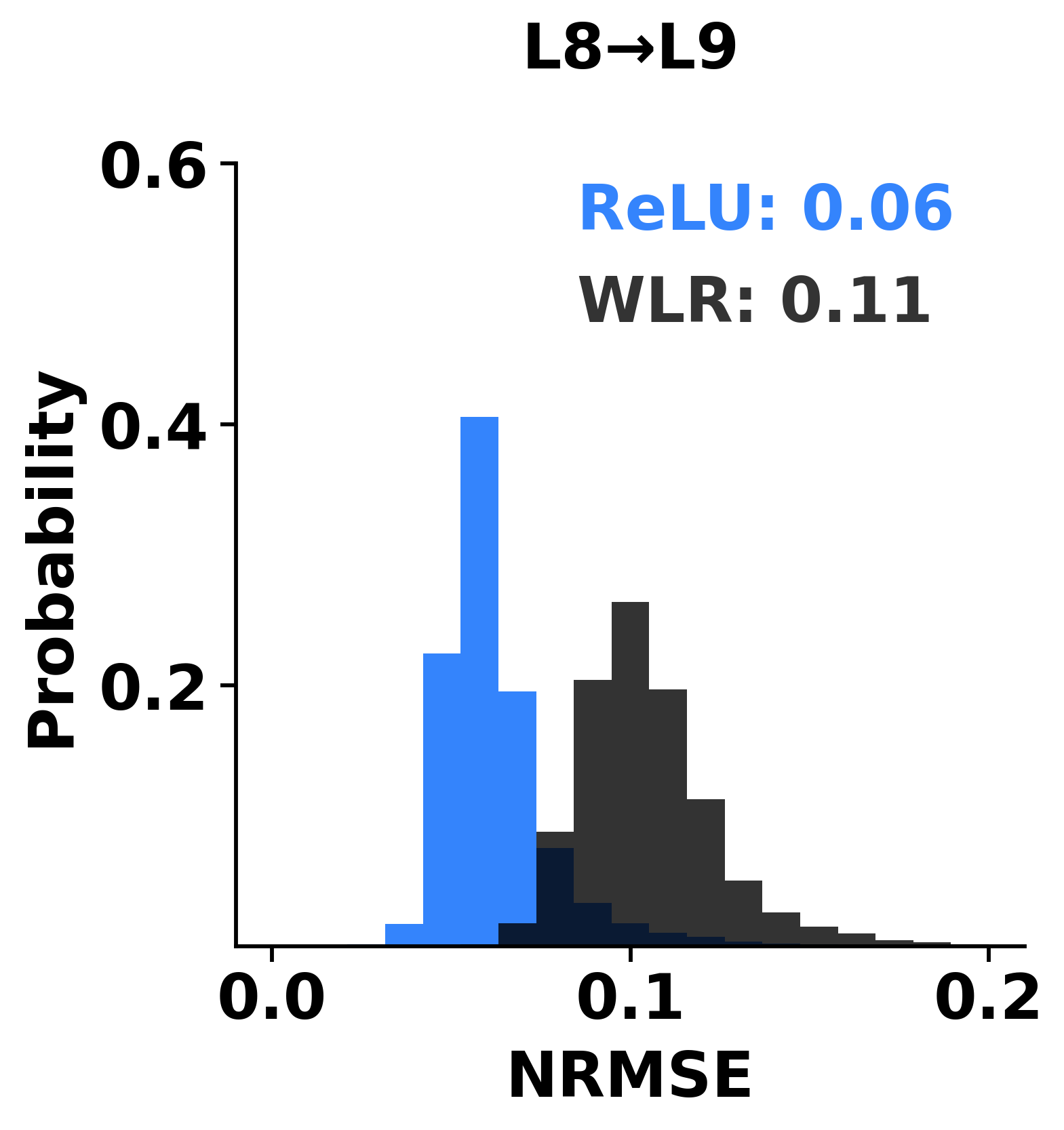}
\put(2,90){\textbf{J}}
\end{overpic} &
\begin{overpic}[width=0.22\linewidth]{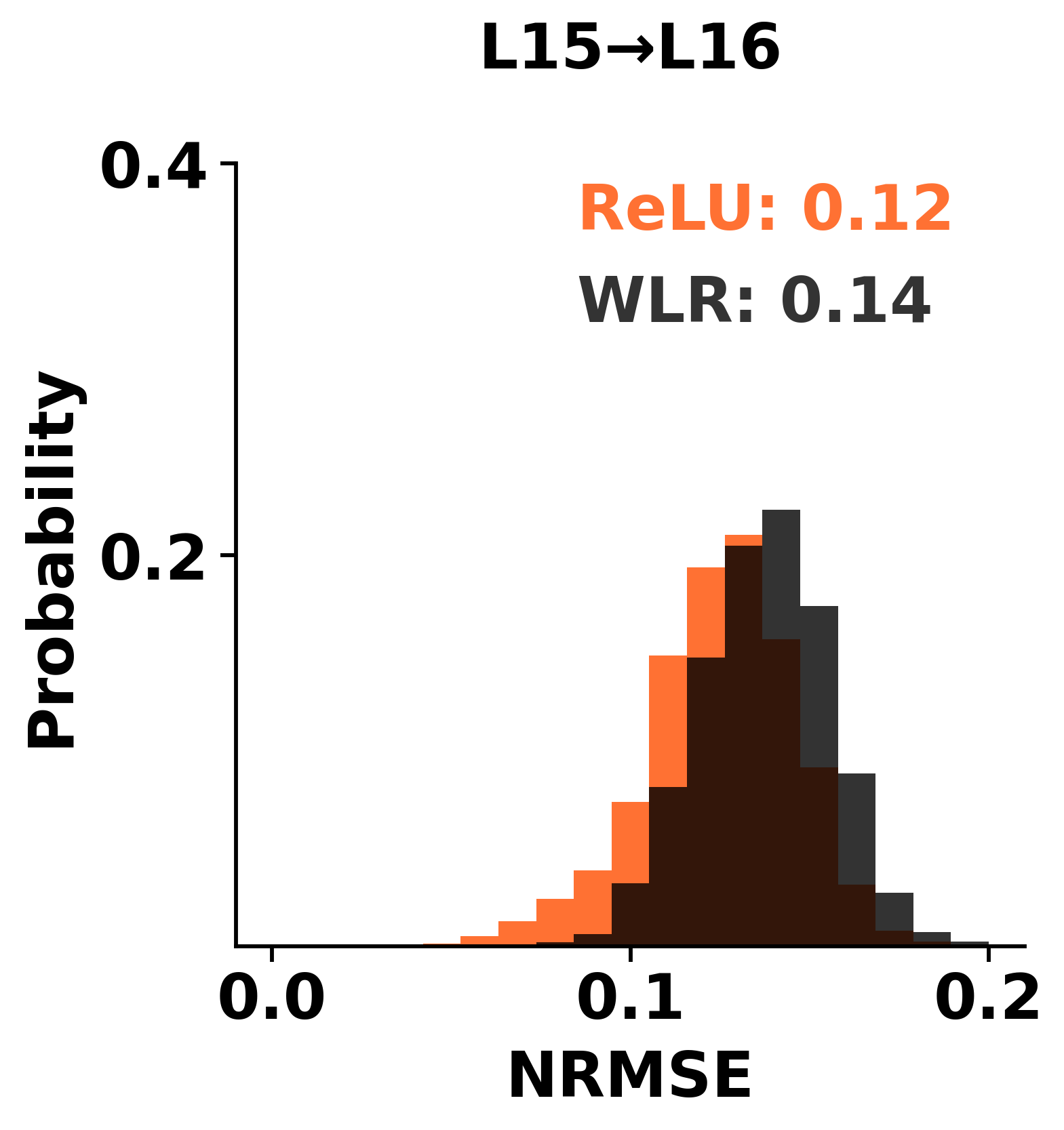}
\put(2,90){\textbf{K}}
\end{overpic} &
\begin{overpic}[width=0.22\linewidth]{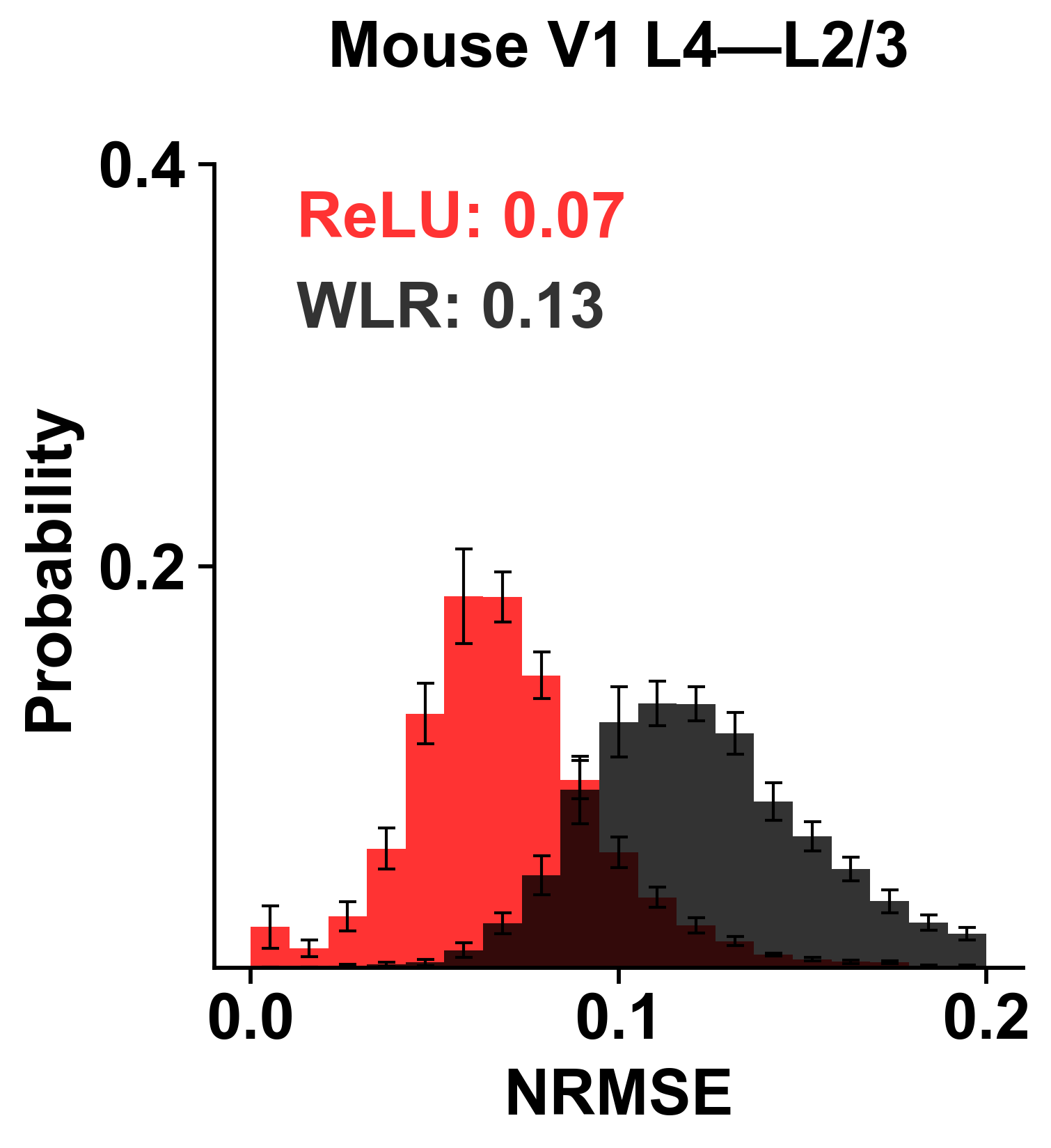}
\put(2,90){\textbf{L}}
\end{overpic}\\ 

\begin{overpic}[width=0.22\linewidth]{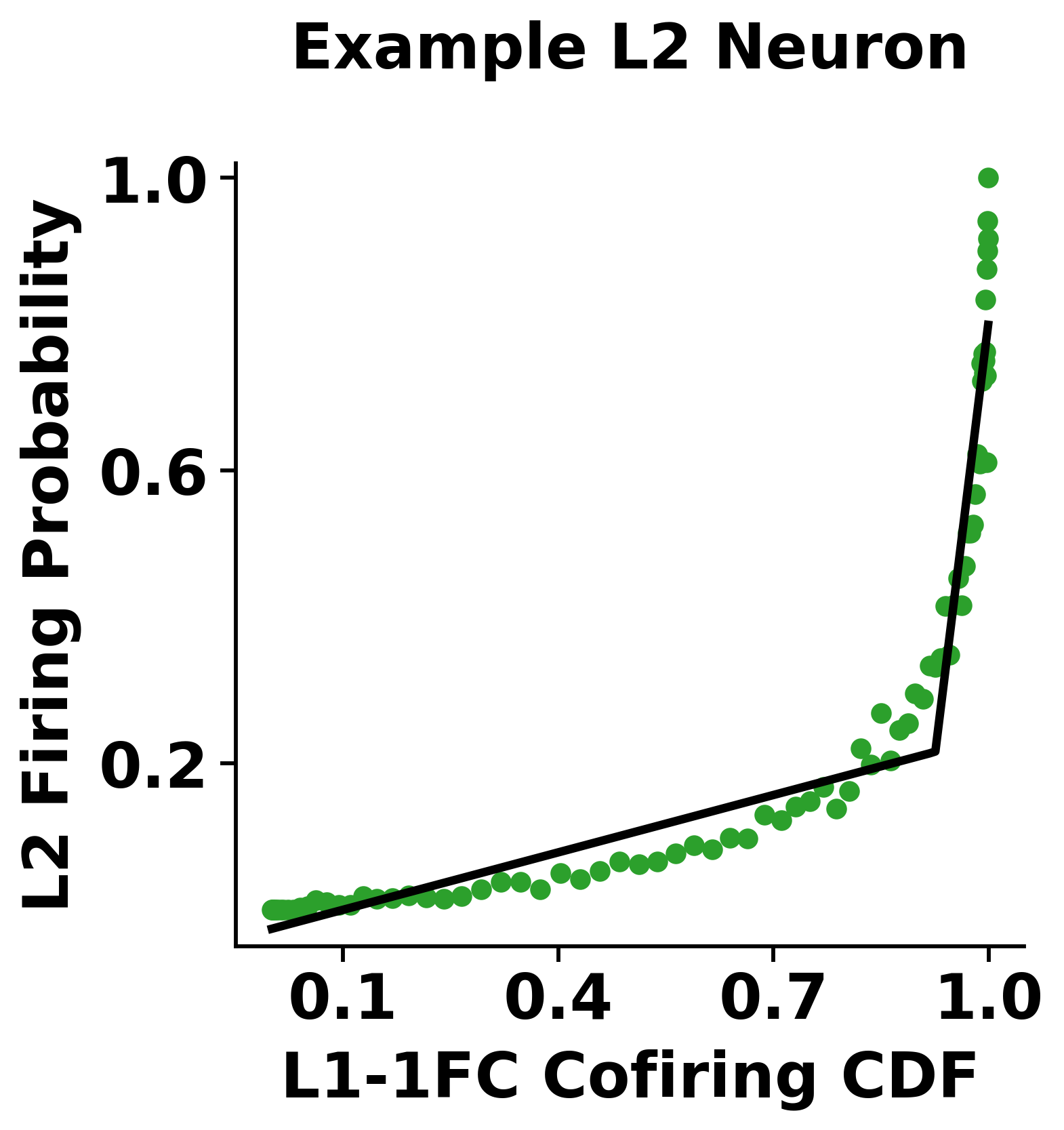}
\put(2,90){\textbf{M}}
\end{overpic} &
\begin{overpic}[width=0.22\linewidth]{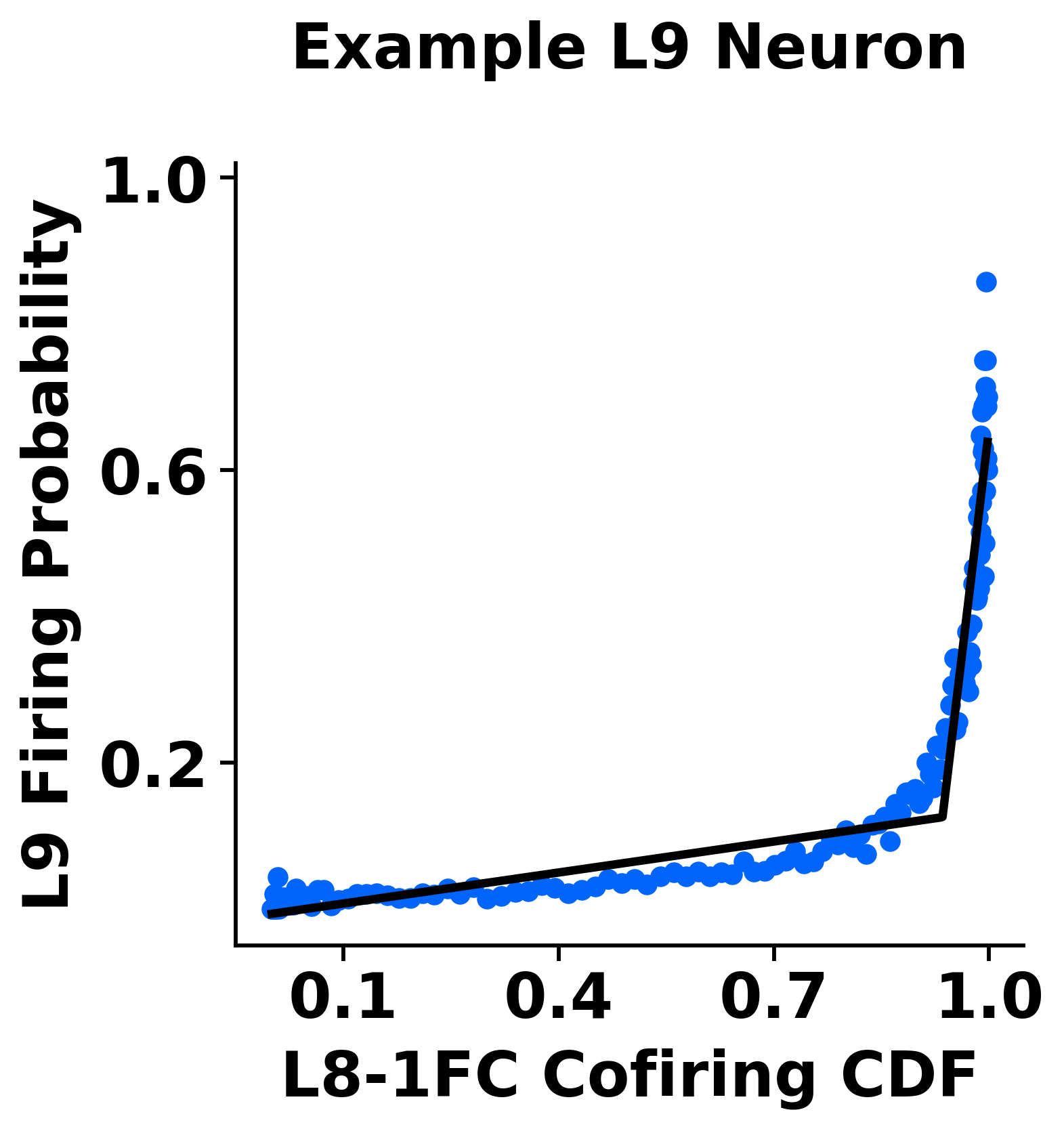}
\put(2,90){\textbf{N}}
\end{overpic} &
\begin{overpic}[width=0.22\linewidth]{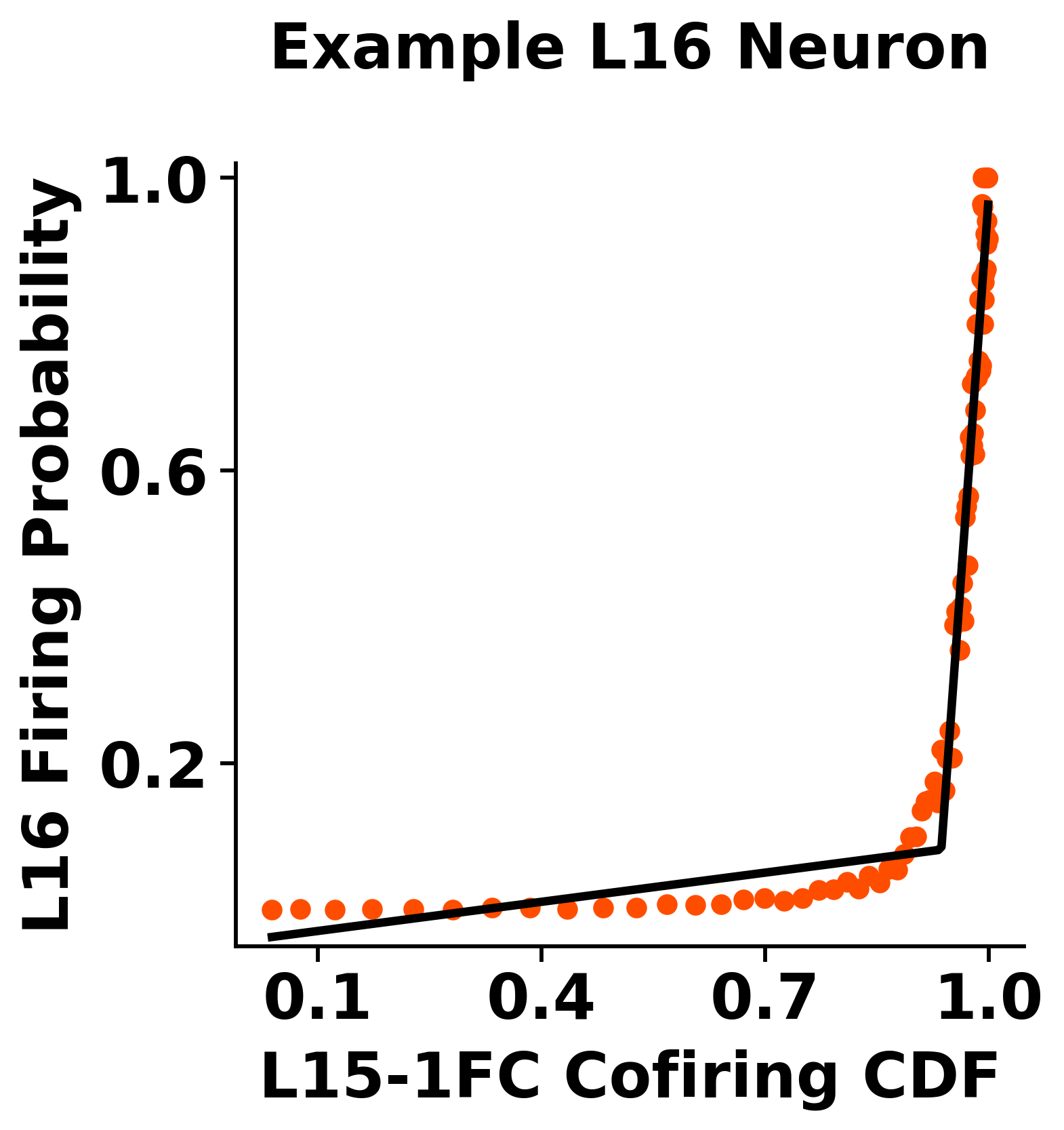}
\put(2,90){\textbf{O}}
\end{overpic} &
\begin{overpic}[width=0.22\linewidth]{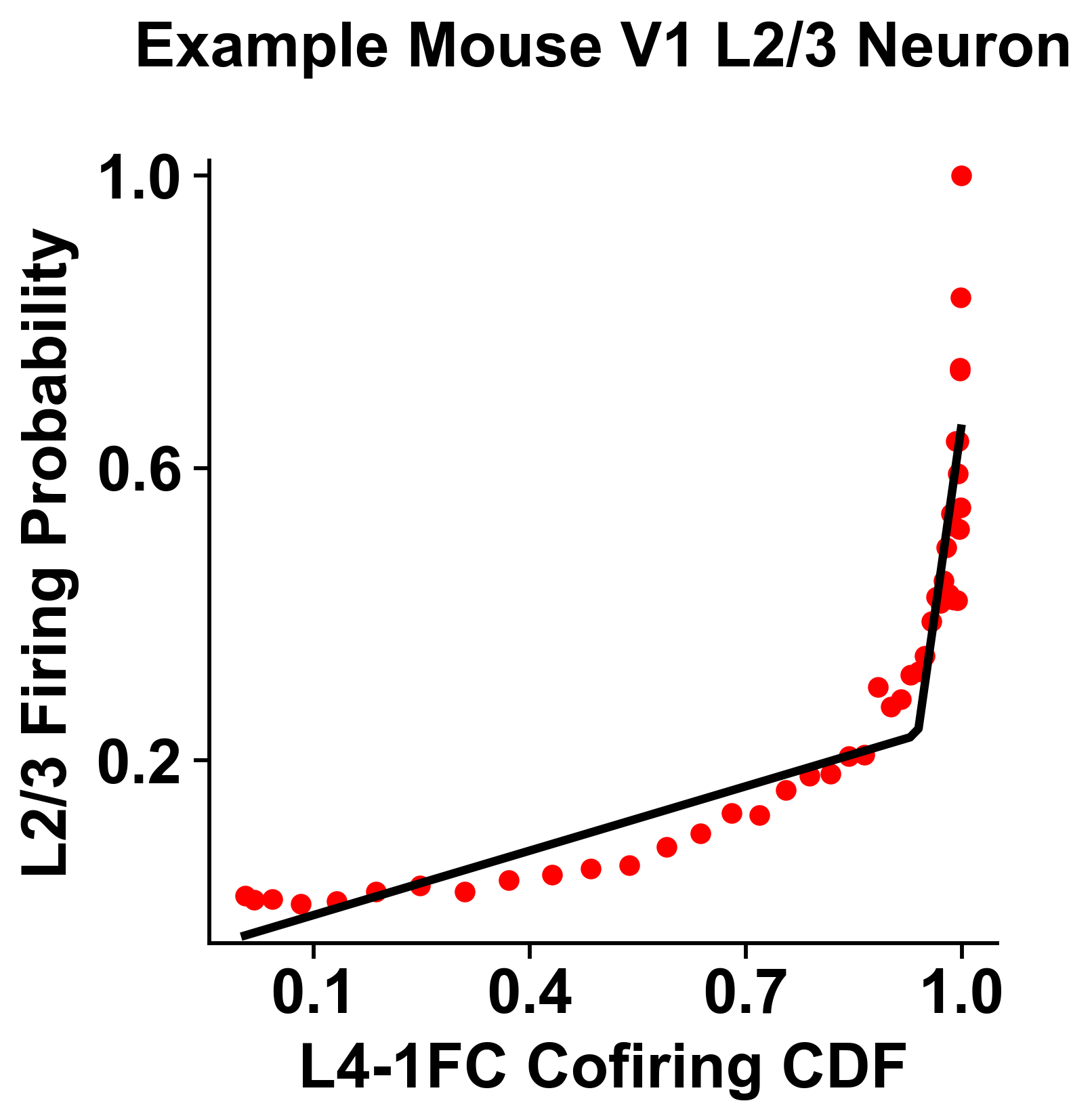}
\put(2,90){\textbf{P}}
\end{overpic}\\ 

\begin{overpic}[width=0.22\linewidth]{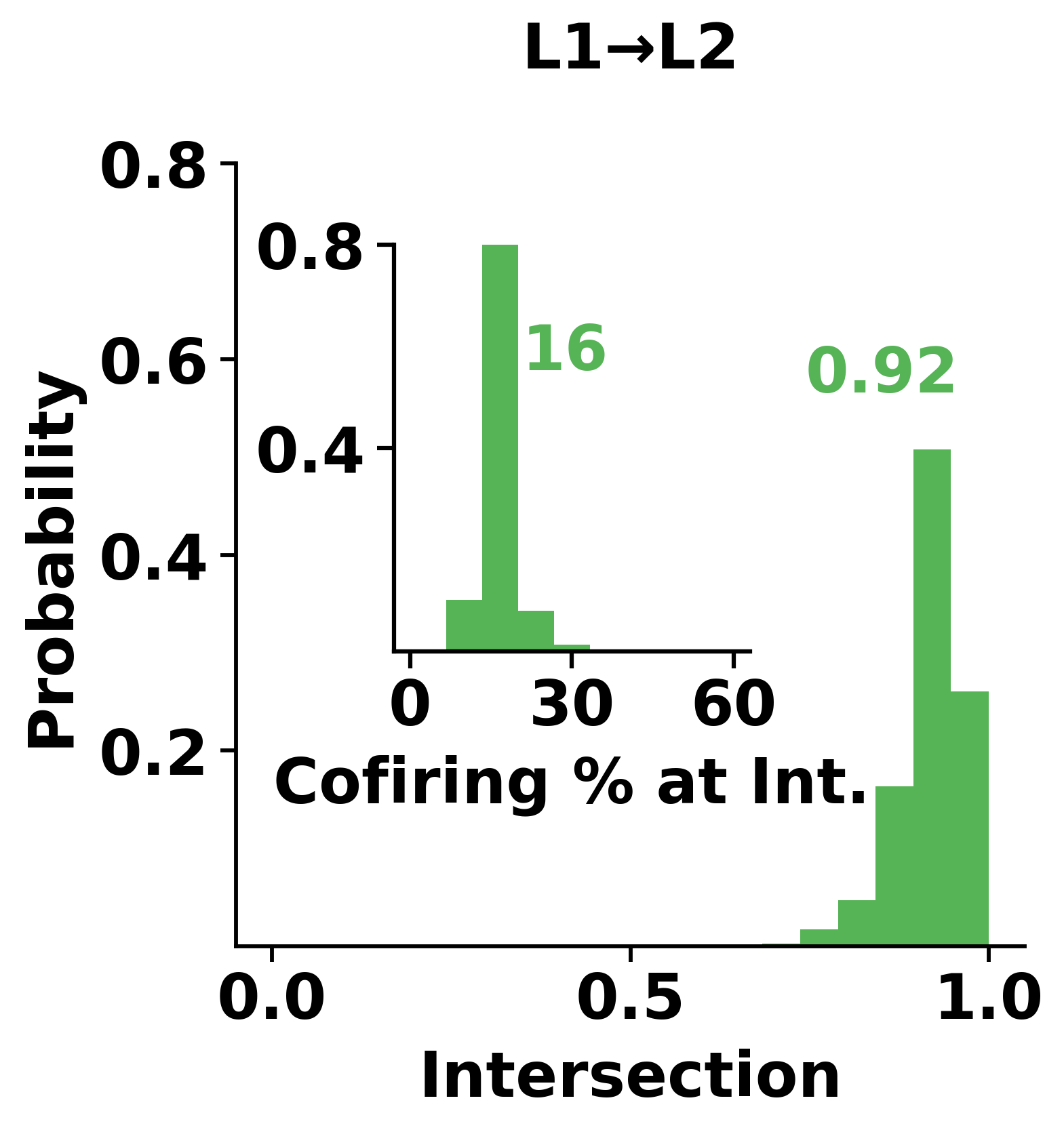}
\put(2,90){\textbf{Q}}
\end{overpic} &
\begin{overpic}[width=0.22\linewidth]{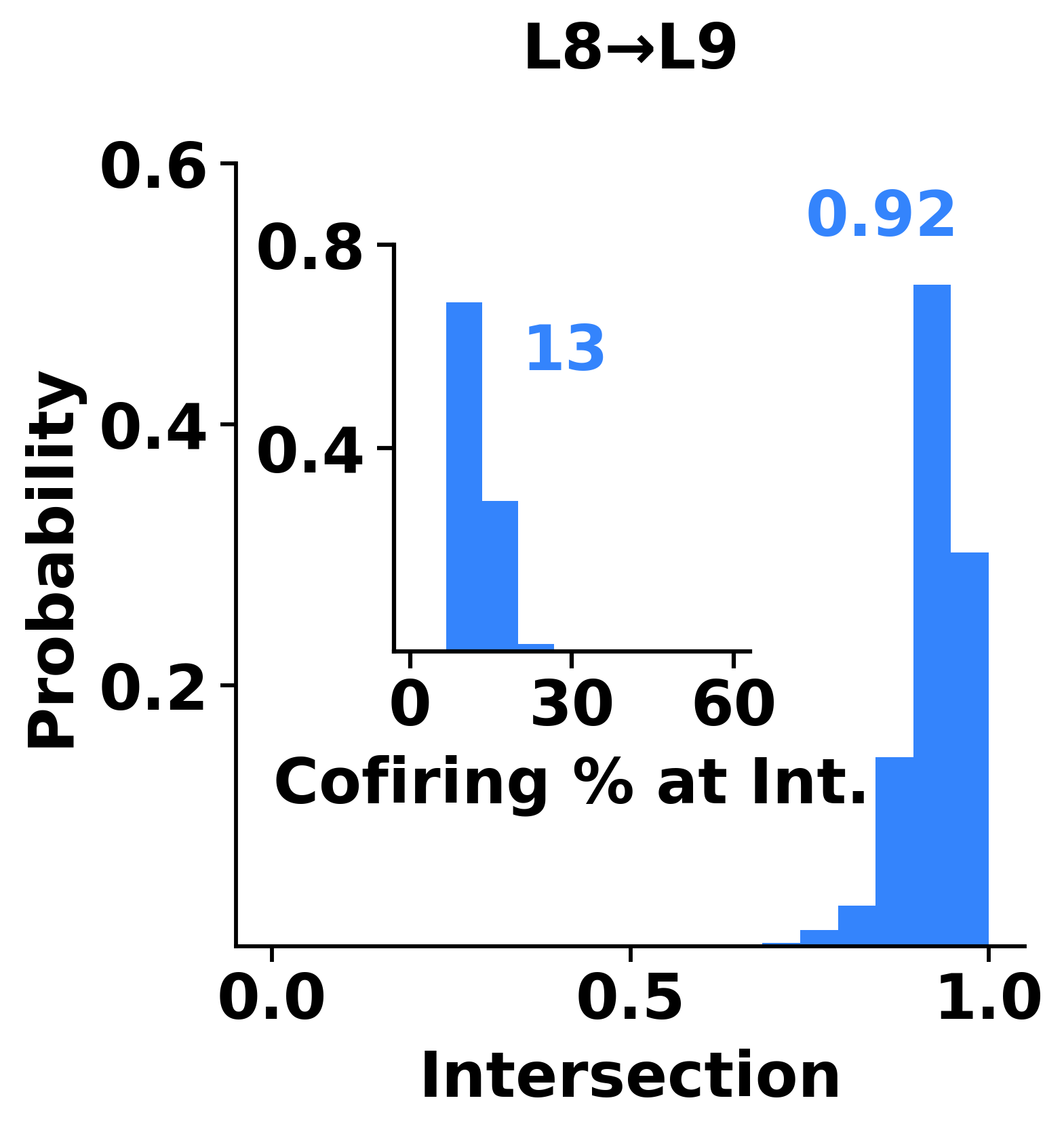}
\put(2,90){\textbf{R}}
\end{overpic} &
\begin{overpic}[width=0.22\linewidth]{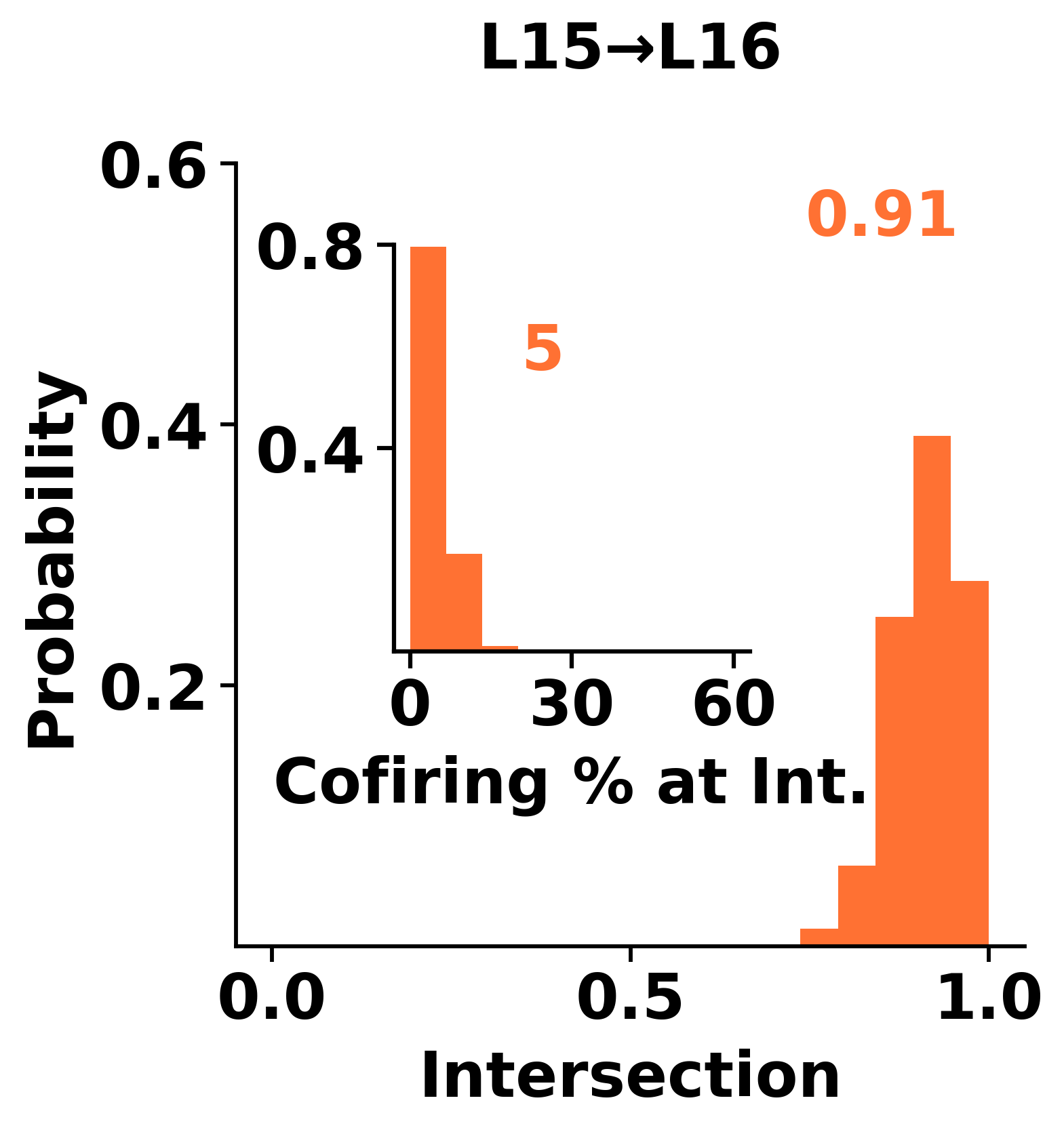}
\put(2,90){\textbf{S}}
\end{overpic} &
\begin{overpic}[width=0.22\linewidth]{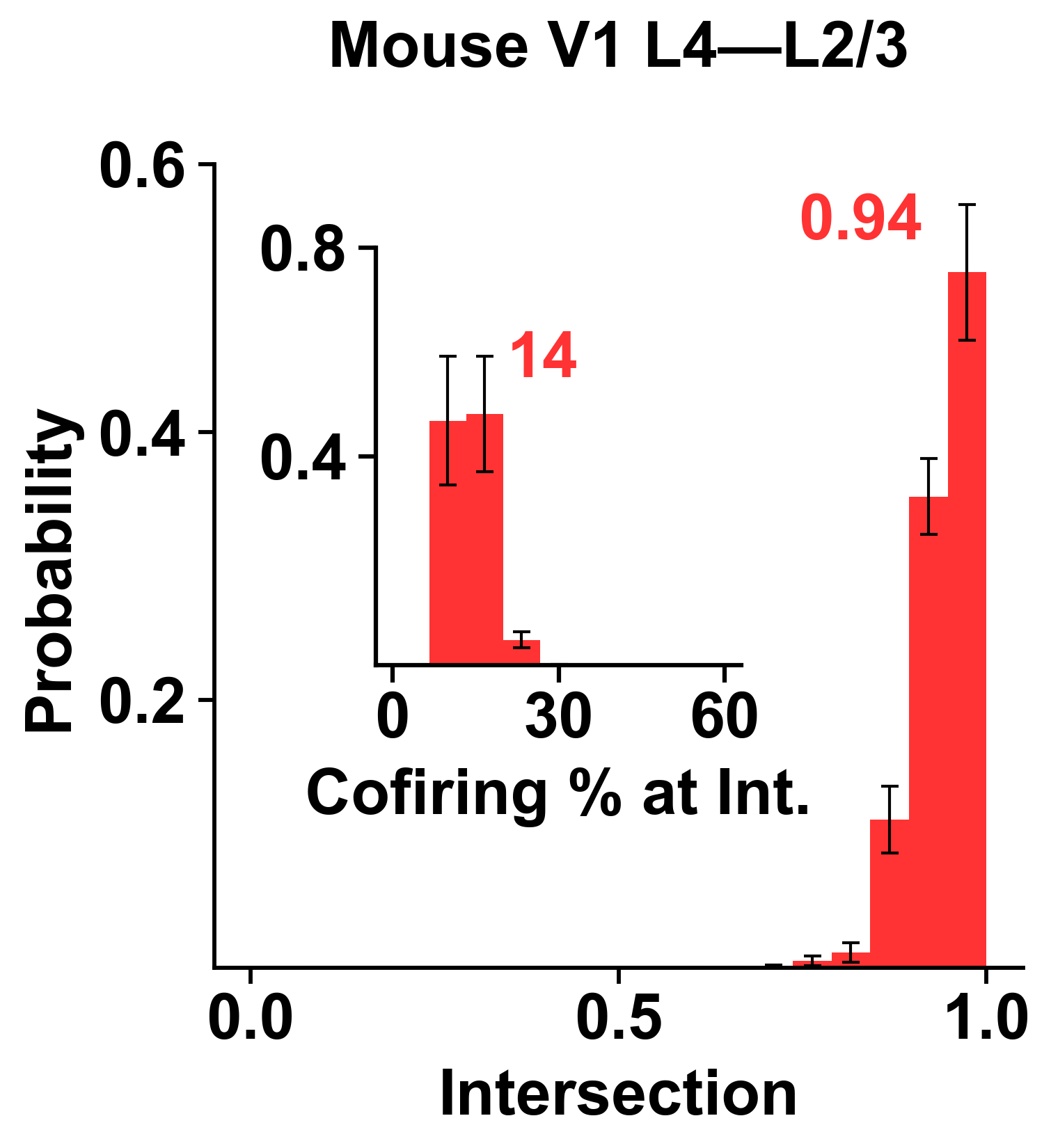}
\put(2,90){\textbf{T}}
\end{overpic}
\end{tabular}
\caption{\textbf{Response of a Neuron as a Function of the Number of Cofiring of its Input 1FC group.} Columns 1–3 show results from the SNN (for layer pairs L1$\rightarrow$L2, L8$\rightarrow$L9, L15$\rightarrow$L16), and column 4 shows corresponding analyses from mouse primary visual cortex (V1). 
\textbf{(A–D) }Example response functions of single output neurons, showing firing probability as a function of the number of cofiring events in their respective input-layer 1FC groups. \textbf{(E–H)} $R^2$ scores of the ReLU fits (black lines in A–D). 
\textbf{(I–L)} Normalized root mean square error (NRMSE) comparing linear (black) and ReLU (red) fits. \textbf{(M–P)} Firing probability as a function of the cumulative distribution (CDF) of group cofirings, highlighting sensitivity to rare, high-cofiring events. 
\textbf{(Q–T)} Cofiring percentages at the intersection points defining the transition to the fast-rising response regime, with insets showing the corresponding distributions of input-layer 1FC cofiring percentages. For the mouse data (column 4), error bars in \textbf{(H)}, \textbf{(L)}, \textbf{(P)}, and \textbf{(T)} indicate SEM across $n=5$ mice. On average, the transition to the fast-rising regime occurs at CDF $\approx 0.92$, corresponding to $\sim33\%$ of the input group firing together. Analyses include only neurons with firing probability $\geq 0.6$ and ReLU fit $R^2 \geq 0.8$; neurons with response-region slope $< 1$ were excluded ($\sim 58 \pm 13\%$ across mice). The slope was estimated as the change in response probability between two cofiring levels divided by the change in their corresponding ECDF values. The slope histograms are presented in Fig.\ref{fig:slopes_violin}A.}
\label{fig:response}
\end{figure}

\textbf{Response of a Neuron as a Function of its input 1FC Cofiring well-fitted by ReLU}. We examined how the activity of a downstream "output" neuron at layer $L$ relates to the aggregate co-firing of its input ($L-1$) 1FC group. 
Across the population, the firing probability of “output” neurons was well described by rectified linear unit (ReLU)-like input–output relationships as a function of their corresponding input-1FC group co-firing. Notably, some neurons reached firing probabilities approaching unity (Fig. \ref{fig:response}). This functional form is of particular interest, as ReLU nonlinearities confer key computational advantages—including sparsity, efficient learning, and representational efficiency—in both biological and artificial neural systems \cite{lecun2015deep,Glorot:ReLU2011}. We therefore focused on neurons whose responses were well fit by a ReLU model ($R^2 \geq $  0.8).
The slope of the response functions scaled
inversely with the size of the corresponding input-1FC groups: neurons associated with smaller groups exhibited steeper input–output functions, whereas those linked to larger groups showed shallower slopes. 
We found that when cofiring was rescaled by $N^{\alpha}$, where N is input 1FC group
size, and $\alpha$  $\sim 0.6-1$ depending the depth, the response
functions collapsed onto a common curve (\ref{fig:Resp_Group_Sizes}).


\textbf{Two Regimes in the Response.} We also observed the existence of weak and strong response regimes: 
Figures \ref{fig:response} M shows a neuron whose response (green) is plotted alongside the cumulative probability distribution of cofiring events within its input-1FC group. The firing probability clearly begins to increase only for events located deep in the tail of the cofiring event size distribution. In general, neurons remain inactive for most cofiring events in their 1FC groups, with their firing probability \textit{rising sharply} only once the event size exceeds approximately 8\% of the largest and \textit{rarest} events. This threshold corresponds to at least 5-16\% of 1FC neurons firing \textit{synchronously} (Fig. \ref{fig:response} Q-S inset).

\textbf{Representational Significance}. To assess the representational significance of the first-order functional connectivity modules in encoding the image being presented, we plot the similarity between the class of the image being presented at a frame with the preferred class of the “index” neuron as a function of the number of its input 1FC cofiring events (y-axis). 
The larger the number of cofiring, the higher the similarity. That is, at the strong response regime, the neurons of the deeper layer encode more reliably the image being presented.
As we ascend the layers, neurons exhibit higher MI for a class, consistent with the improved class discrimination trend in CNNs\cite{Zeiler2014:CNN}. Especially in last layer (L16), the higher the degree of connectivity, the larger the MI. A cluster of hub neurons with high degree of connectivity and MI is prominent (Fig. \ref{fig:MI_vs_DoC}). 
To further examine the relationship between  the index neuron firing at a given layer (say layer $L$) and the cofiring events of its layer $L-1$ 1FC group, we applied several classifiers to predict on a frame by frame basis, whether the neuron fires from the cofiring events that occurred in the preceding layer's 1FC group. Preserving the identities of individual neurons within input–1FC groups does not provide significant additional predictive value beyond the aggregate co-firing signal when modeling the activity of the corresponding output neuron (see \ref{sec:membership-1FC}, Fig. \ref{sfig:identity_barplots}). 
This suggest that ensemble co-firing magnitude may play a more prominent role in shaping downstream responses than the specific identities of the contributing neurons.

\textbf{Temporal Robustness of the Response.} To verify that the 1FC groups are not an artifact of the specific images used to estimate functional connectivity, we assess the temporal robustness of response functions across held-out input samples. 
The consistency of response function slopes across batches and across the full range of input images confirm that the 1FC groups reflect stable functional structure rather than sample-specific co-activation (Figs. \ref{fig:Resp_Temp_Robust}, ~\ref{fig:Slopes_Temp_Robust}, Sec. \ref{sec:temp_robustness}).

\textbf{Properties of a Neuron.} We then characterize a neuron based on the main properties of its response on a certain test set, namely the \textit{slope} of the ReLU fit, the 1FC-cofiring \% at the intersection of the two response regimes, the response rise intersection (CDF of the 1FC-cofiring), and the semantic rank similarity. For each neuron, we also compute the membership and size of its 1FC group (i.e., degree of connectivity) and its informativeness with respect to the original test set. 

\begin{figure}[h]
    \centering
    \begin{overpic}[width=\linewidth]{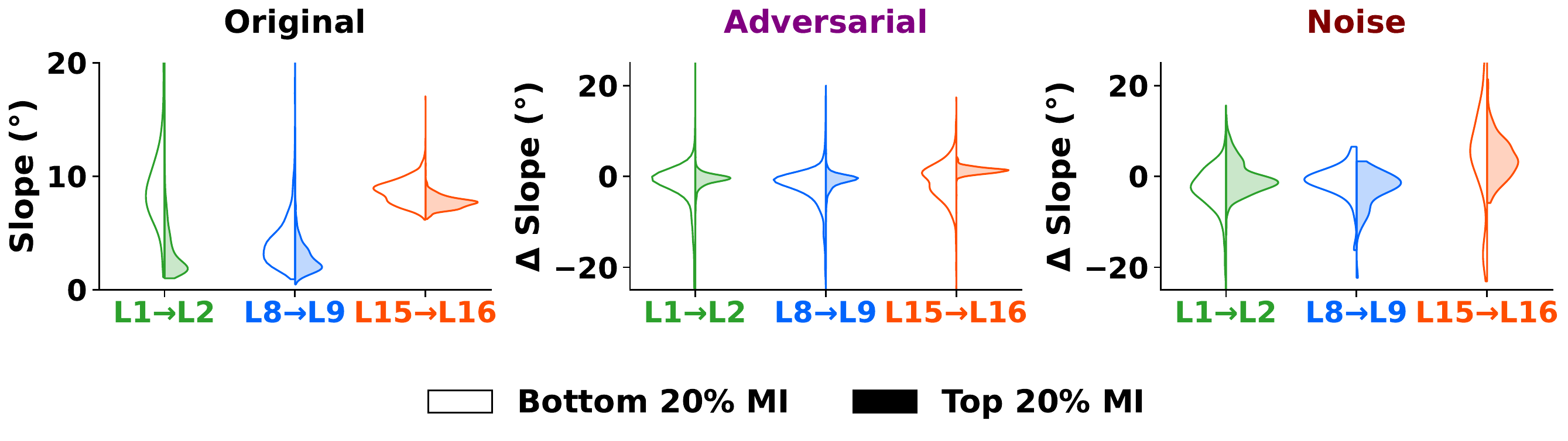}
        \put(2,28){\textbf{A}}
        \put(35,28){\textbf{B}}
        \put(68,28){\textbf{C}}
    \end{overpic}
\caption{\textbf{Slopes of response functions under different input conditions} \textbf{(A)} Violin plots of neuronal response slopes for layer pairs L1$\rightarrow$L2, L8$\rightarrow$L9, and L15$\rightarrow$L16. For each violin, the left half corresponds to neurons with MI below the 20th percentile, and the right half to neurons with MI above the 80th percentile. \textbf{(B)} Same as \textbf{(A)}, but showing the \textit{difference} in slopes between \textit{adversarial} and original test, respectively. 
\textbf{(C)} Same as \textbf{(B)} but on the \textit{difference} in slopes between \textit{uniform noise} and original test.(Fig. \ref{fig:MI_Adv_Scatter})}
\label{fig:slopes_violin}
\end{figure}

\begin{figure}[h]
    \centering
    \begin{overpic}[width=\linewidth]{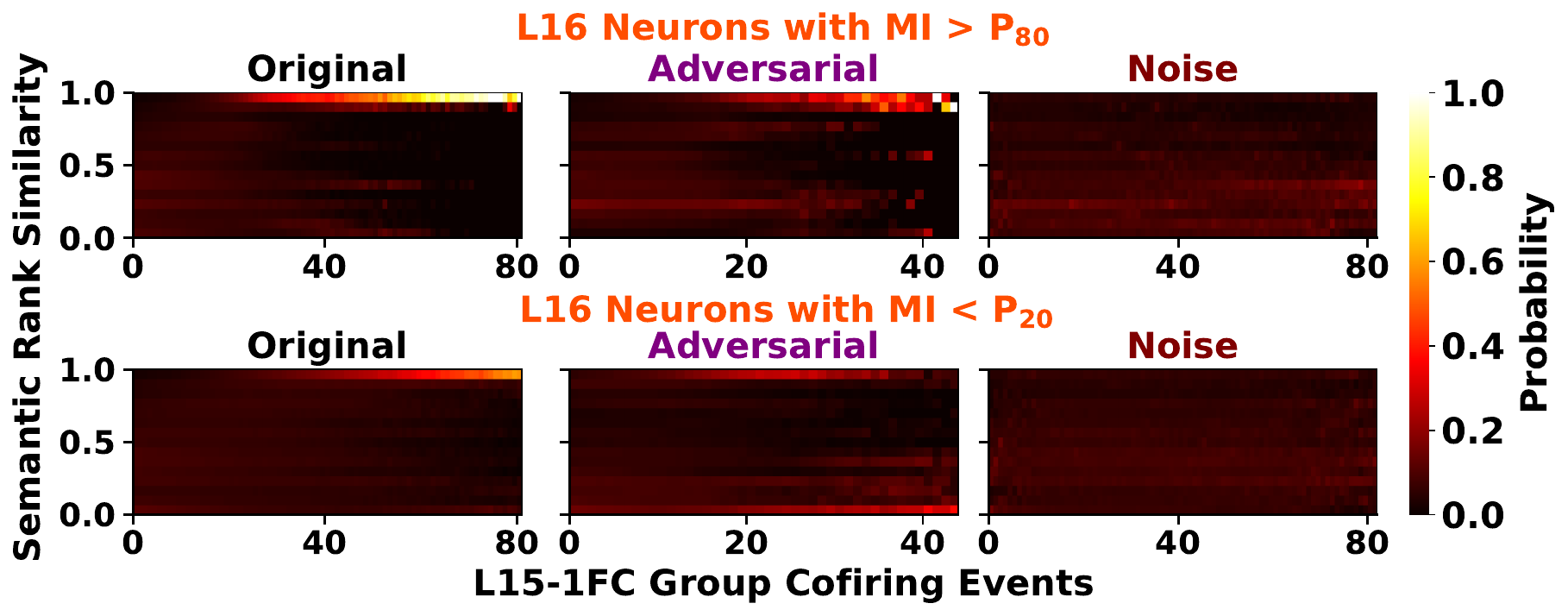}
        \put(4,35){\textbf{A}}
        \put(34,34.5){\textbf{B}}
        \put(62,34.5){\textbf{C}}
    \end{overpic}

\caption{\textbf{Information encoding of neurons as a function of their 1FC cofiring.} \textbf{(A-Top)} Heatmap of semantic rank similarity for L16 neurons in the top 20th percentile of class-wise mutual information (MI), shown as a function of L15-1FC group co-firing events under original input. Each bin indicates the probability that co-firing frames match the neuron’s preferred class. \textbf{(A) }The larger the L15-1FC cofiring, the higher the semantic similarity of L16-neurons. \textbf{(B-C Top)} Same as (A) under adversarial and random noise input, respectively, with L15-1FC groups fixed (formed under original input). High-MI neurons remain relatively informative under adversarial perturbations. Their higher degree of connectivity (Fig.~\ref{fig:MI_vs_DoC}) may support robustness via input aggregation. \textbf{(A-C Bottom)} Same analysis for neurons in the lowest 20th percentile of MI. For each co-firing bin, probabilities are normalized to sum to 1 across semantic rank similarity (Fig.~\ref{fig:semantic_rank_sim_metric}).}
\label{fig:semantic_heatmaps_all}
\end{figure}

\textbf{Robustness of Neuronal Response Functions Under Additive Uniform Noise.} Here we analyze the response of the neurons on the test set with random noise 
(Sec. \ref{sec:add_noise}). For each neuron, we consider its 1FC group as formed during the original test set.  All neurons exhibited lower firing rates under noise input (Fig. \ref{sfig:noise_firing_rates}),  
L15-1FC groups of L16 neurons exhibited infrequent co-firing events (Fig. \ref{fig:semantic_heatmaps_all}C) and had large deviations in their response slopes (Fig. \ref{fig:slopes_violin}C).


\begin{figure}[h]
\centering
    \begin{tabular}{ccc}
    \begin{overpic}[width=0.3\linewidth]{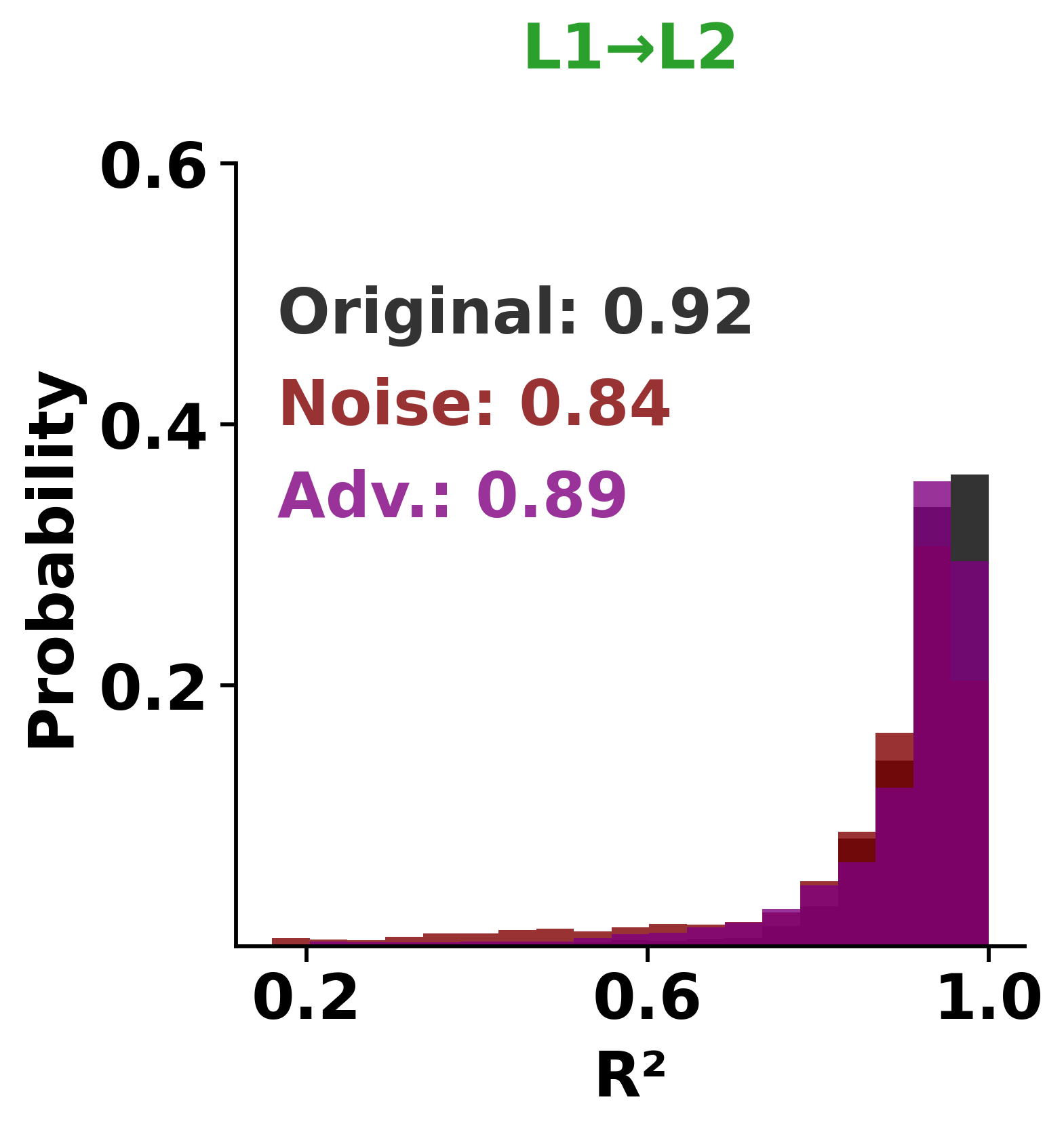}
    \put(1,90){\textbf{A}}
    \end{overpic} &
    \begin{overpic}[width=0.3\linewidth]{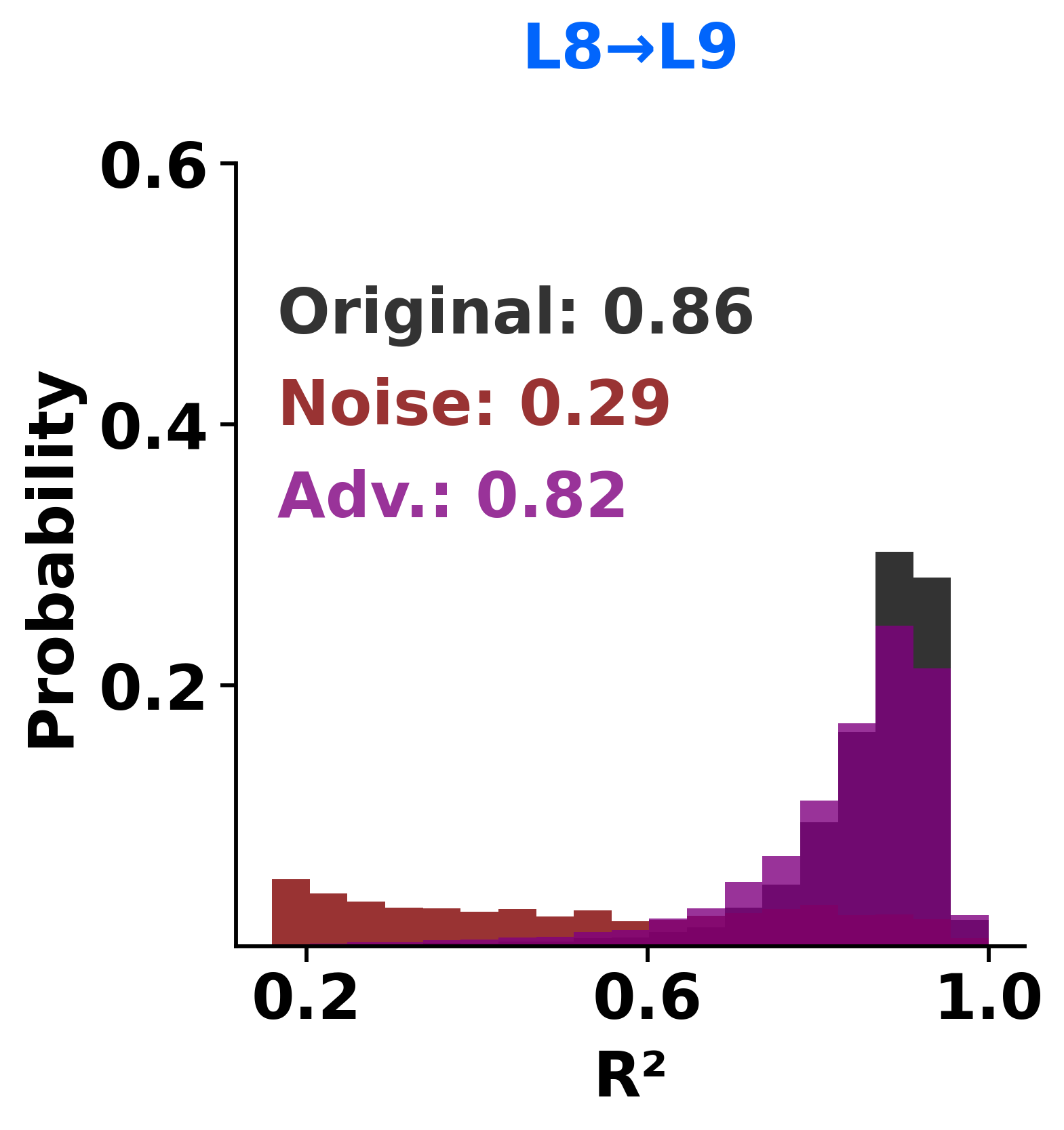}
    \put(1,90){\textbf{B}}
    \end{overpic} &
    \begin{overpic}[width=0.3\linewidth]{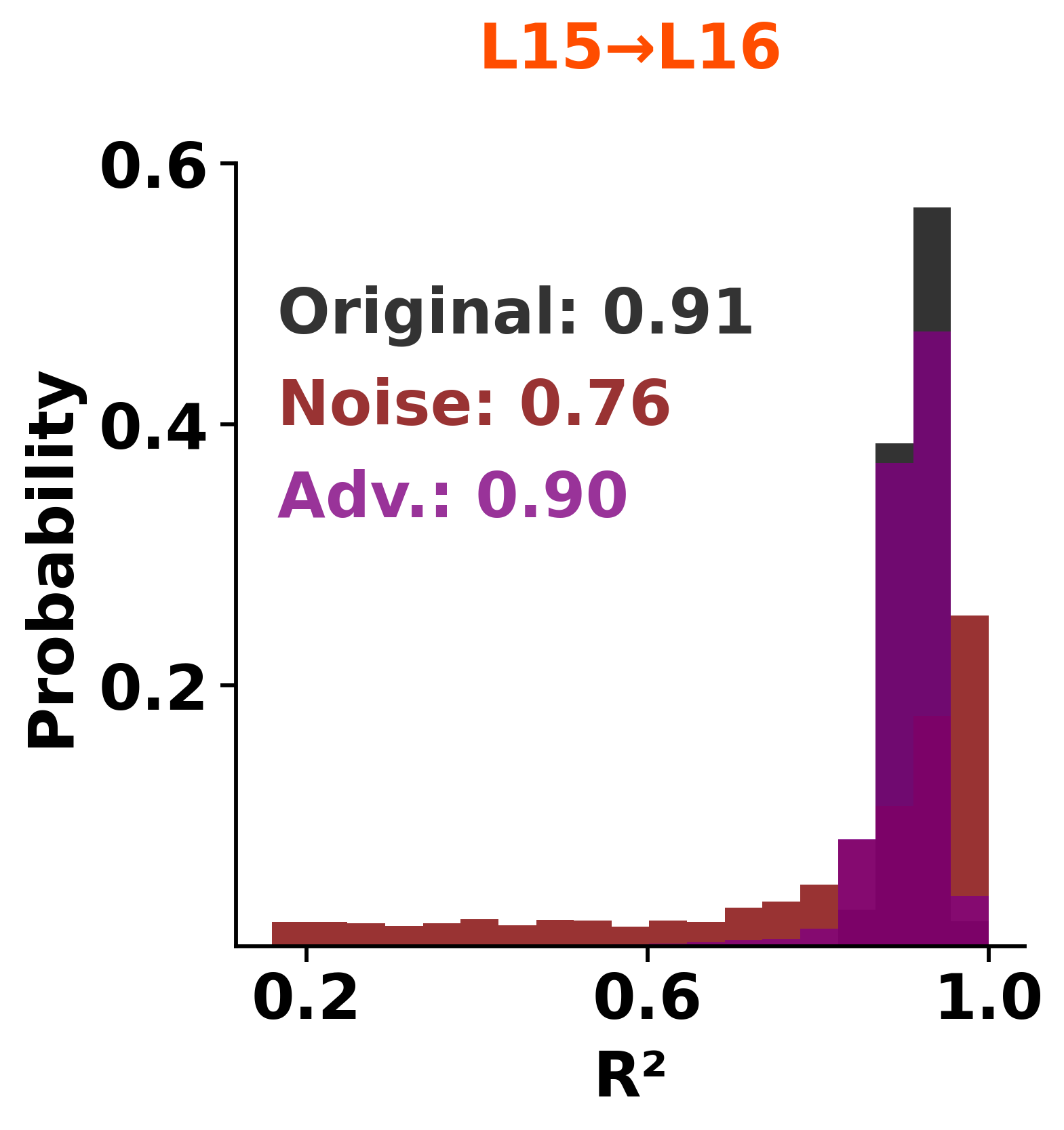}
    \put(1,90){\textbf{C}}
    \end{overpic} \\
    \begin{overpic}[width=0.3\linewidth]{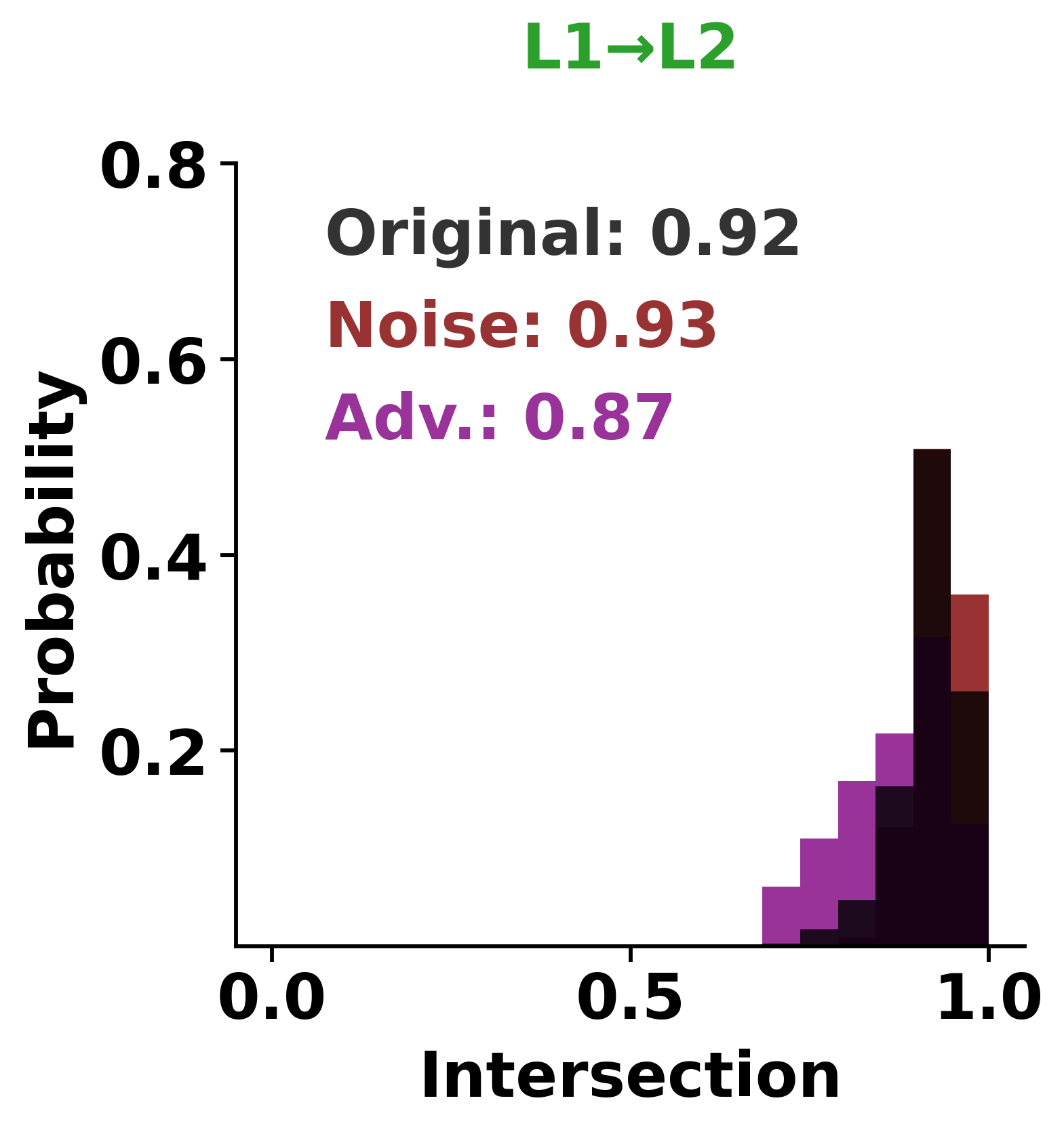}
    \put(1,90){\textbf{D}}
    \end{overpic} &
    \begin{overpic}[width=0.3\linewidth]{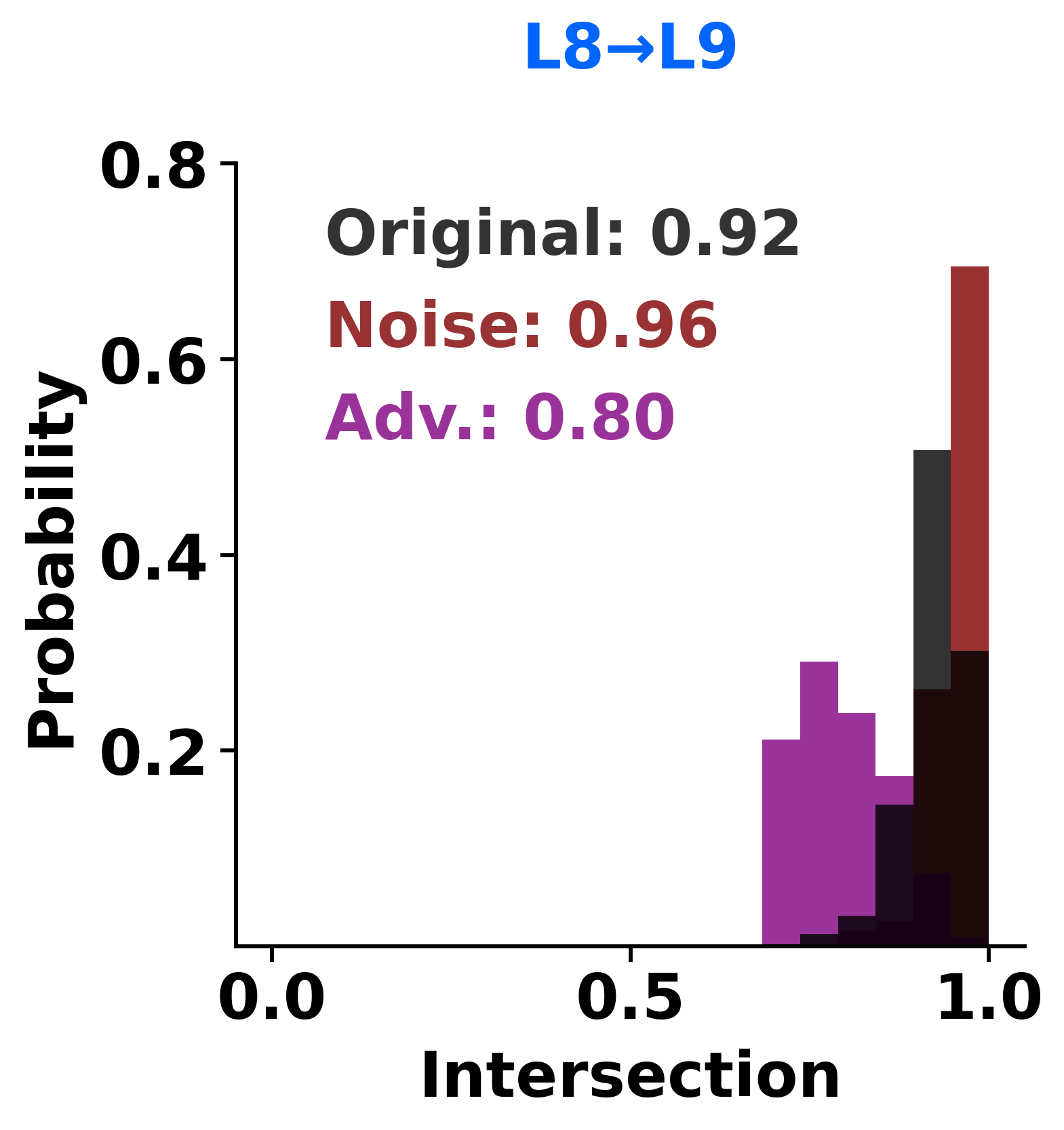}
    \put(1,90){\textbf{E}}
    \end{overpic} &
    \begin{overpic}[width=0.3\linewidth]{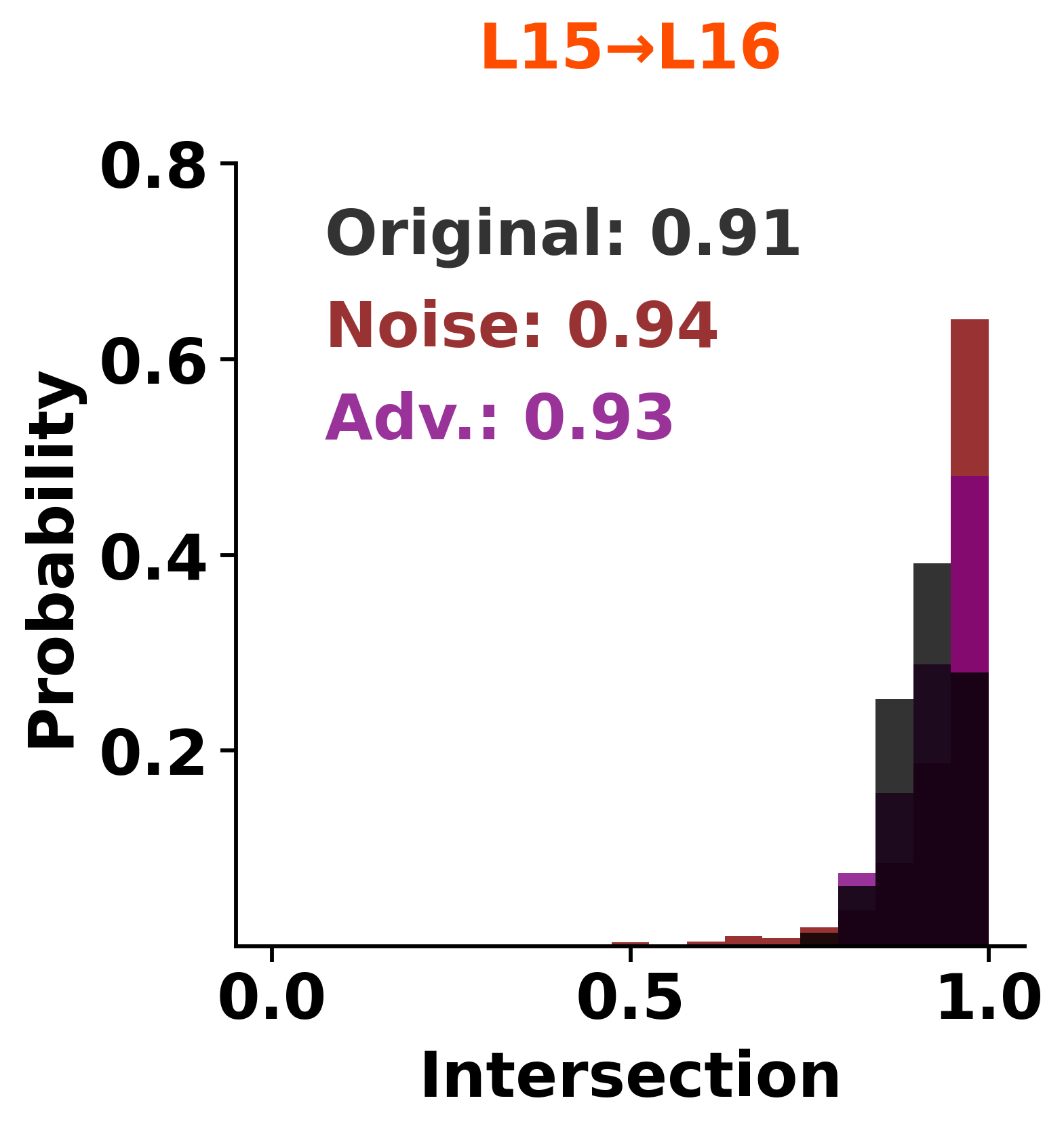}
    \put(1,90){\textbf{F}}
    \end{overpic} \\
    \begin{overpic}[width=0.3\linewidth]{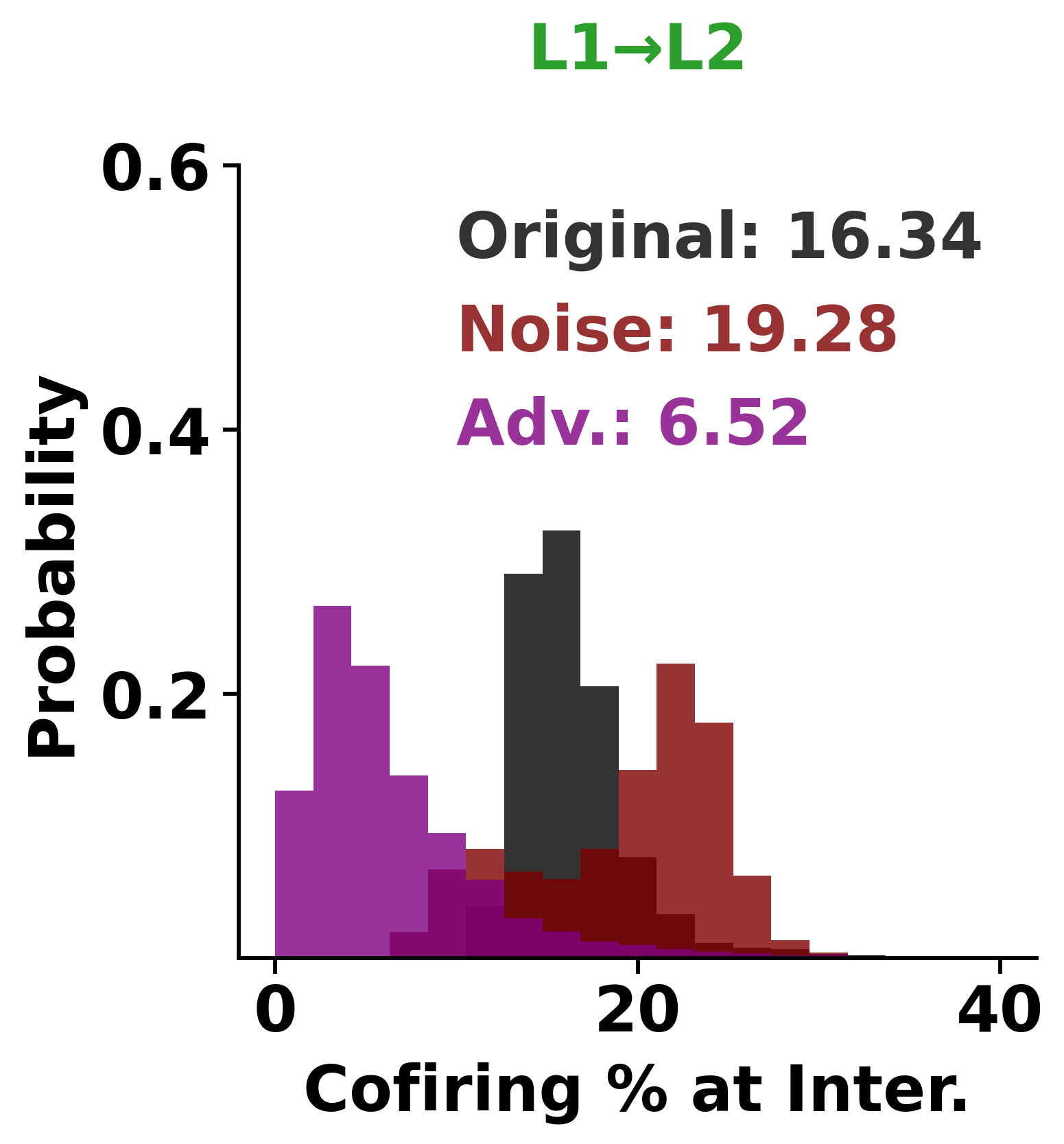}
    \put(1,90){\textbf{G}}
    \end{overpic} &
    \begin{overpic}[width=0.3\linewidth]{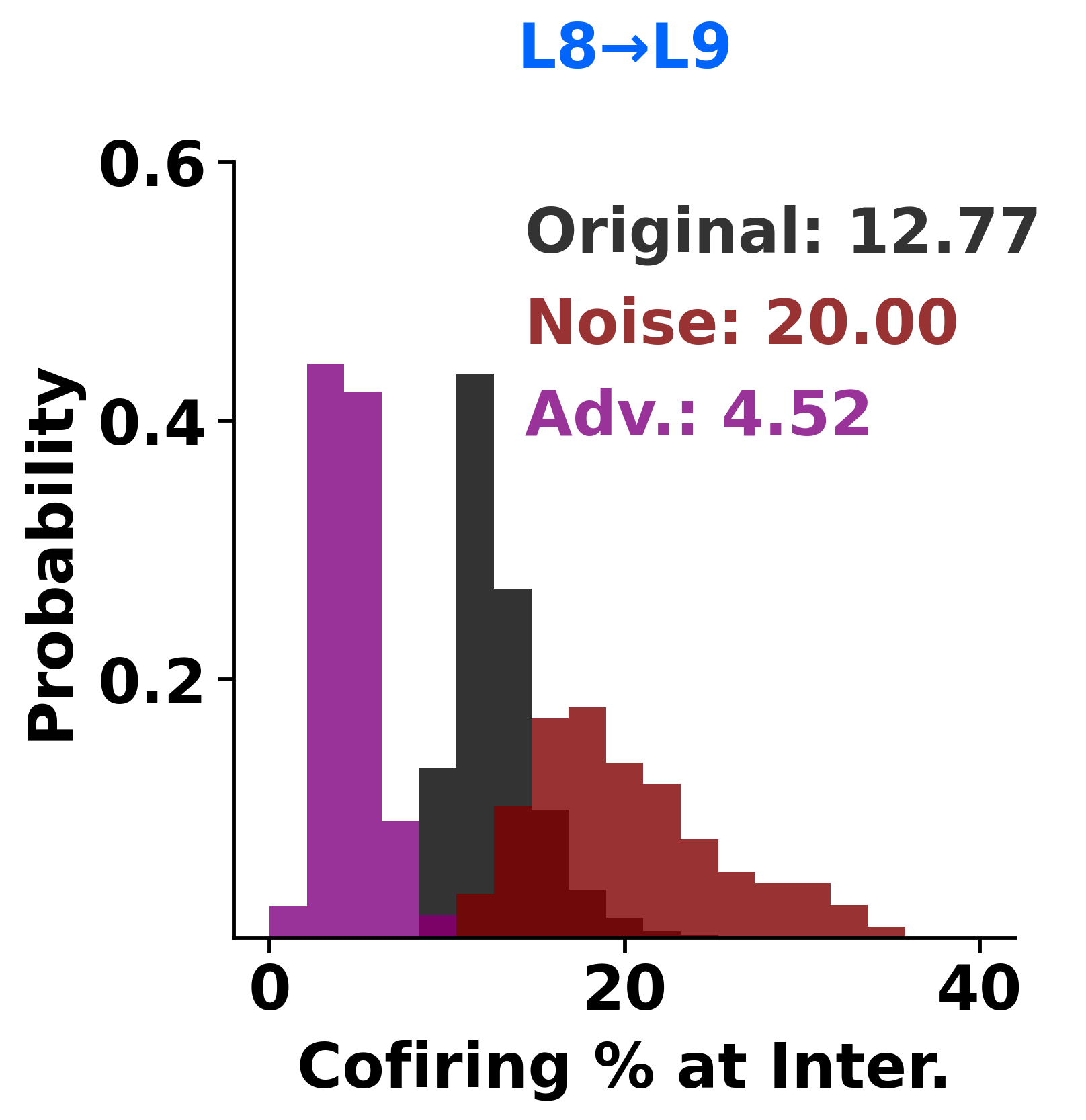}
    \put(1,90){\textbf{H}}
    \end{overpic} &
    \begin{overpic}[width=0.3\linewidth]{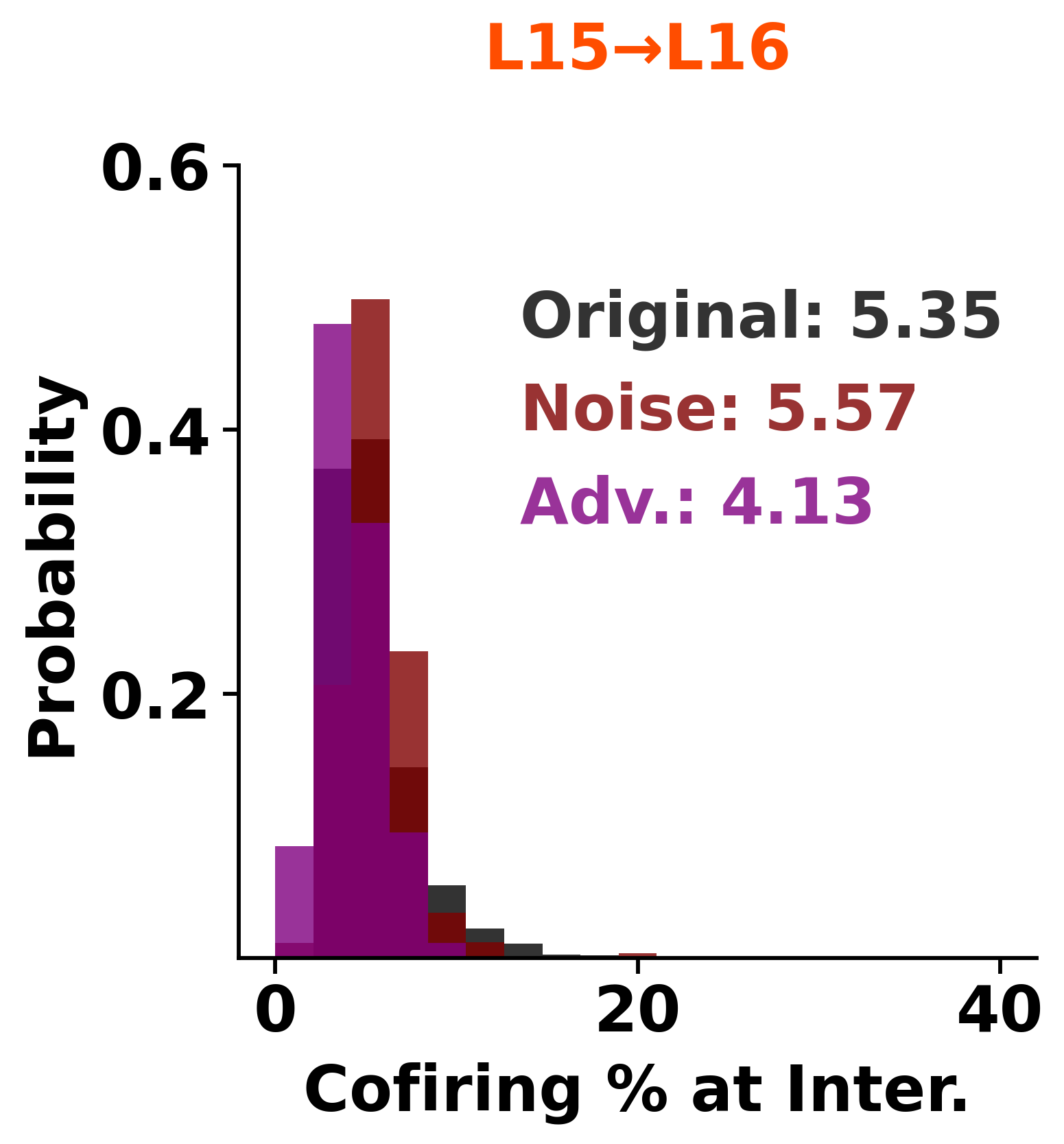}
    \put(1,90){\textbf{I}}
    \end{overpic} \\
    \end{tabular}
    \caption{\textbf{Response characteristics as they change under different conditions}. \textbf{(A-C)} Histograms of $R^2$ values resulting from ReLU fits of neuronal responses as a function of their respective input-layer 1FC group cofirings, for layer pairs L1$\rightarrow$L2, L8$\rightarrow$L9, L15$\rightarrow$L16 for \textit{original} input (black), \textit{random noise} (maroon), and \textit{adversarial} input (purple). \textbf{(D-F)} Histograms of the intersection points corresponding to the transition to the fast-rising response regime for the same three layer pairs. \textbf{(G-I)} Distributions of input-layer 1FC group cofiring percentages at these intersection points. On average, under the original input, this transition occurs at CDF $\approx 0.92$, corresponding to $\sim16\%$ of the 1FC cofiring. Under noise and adversarial conditions, these histograms of the first and intermediate layers change significantly. Neurons with $R^2 < 0.8$ or firing rate $< 0.01$ Hz were excluded.}
    \label{fig:unoise_intersection}
\end{figure}
\textbf{Robustness of the Response under Adversarial Noise.}
To probe the sensitivity of individual neuronal response functions under targeted input perturbation (Sec. \ref{sec:add_noise}), we applied the Fast Gradient Sign Method (FGSM) \cite{Goodfellow2014ExplainingAH} adversarial attack to the test set. FGSM generates adversarial examples by perturbing the input in the direction of the gradient of the loss with respect to the input.
Under adversarial input, individual neuronal response functions preserved qualitatively similar trends to those observed under clean input, suggesting a degree of structural robustness in the population-level response organization. To quantify this at the single-neuron level, we computed the difference in the slopes of the response functions between adversarial and original inputs, and examined this quantity as a function of each neuron's mutual information with respect to the output class (Fig. \ref{fig:MI_Adv_Scatter}). Neurons exhibiting high MI showed minimal deviation in response function slope between original and adversarial conditions across all layers, suggesting that informationally relevant neurons maintain stable tuning under perturbation. In contrast, neurons with low MI displayed substantially greater variability in slope differences, 
suggesting that weakly informative neurons are more susceptible to adversarial perturbation, with this susceptibility attenuating towards deeper layers.
In the presence of adversarial perturbations, L16 neurons with high MI remain relatively informative (see Figs. \ref{fig:semantic_heatmaps_all}B, \ref{fig:semantic_rank_sim_metric} B). Notably these L16 high-MI neurons also exhibit a high degree of connectivity (Fig. \ref{fig:MI_vs_DoC}), which allows them to integrate input from multiple upstream nodes. This dense connectivity may confer a degree of robustness, enabling these neurons to partially mitigate the impact of adversarial perturbations through aggregation and implicit noise averaging. Nevertheless,
overall model performance (Fig. \ref{fig:noise_example_images}A) is still degraded due to the propagation of noisy information from nodes with smaller degree of connectivity to the output layer. The adversarial input yields images with imperceptible degradation, whereas images with additive random noise input 
introduce substantial visual corruption Fig. \ref{fig:noise_example_images}B, noticable in the response function (e.g., Fig.\ref{fig:semantic_heatmaps_all}B, \ref{fig:semantic_heatmaps_with_MI_original}, \ref{fig:semantic_heatmaps_with_MI_noise}).

\textbf{Diagnostics: Detection of Adversarial Frames}
We developed a diagnostic framework to detect adversarial perturbations using cofiring patterns of 1FC groups in ResNet output layers. 1FC groups were identified across layers L1–L16 from 10,000 training examples of CIFAR100. Next, spike trains (T=4, length 40,000) were generated for clean and adversarial inputs from the full CIFAR-100 test set. Finally, a logistic regression classifier was trained on the resulting cofiring statistics to distinguish adversarial from clean inputs.

\section{Discussion, Conclusions, and Future Work}\label{sec:disc}
Our results support the view that first-order functionally connected (1FC) ensembles act as computational modules that mediate information processing across hierarchical layers.
In a convolutional architecture, neurons receiving common input through the same convolutional kernel may exhibit correlated activity by virtue of its input and structural organization
(Fig. \ref{fig:BP_vs_STTC_weights}). 
To contrast direct vs. indirect connections, we partition each neuron's 1FC group into \textbf{i)} the \textit{structural} subset (\textit{STR}), comprising source-layer neurons whose activations are routed to the reference neuron via the convolutional kernel, representing (\textit{direct} connections); and \textbf{ii)} the  subset (\textit{nSTR}) of \textit{all remaining} 1FC members, whose co-activation with the reference neuron cannot be attributed to direct information flow and must therefore reflect feature-level correlations learned during training (in part via indirect structural connections). For each neuron, we estimate all its statistically significant functional connections with all neurons in the previous layer (Fig. \ref{fig:ALL_STR_nSTR}). As expected, the first layer (L1) is driven by the spatial locality of the input, which shapes the structural connections, resulting to higher STTC weights and larger normalized degree of connectivity. Interestingly, at the last layer (L15$\rightarrow$L16), the pairwise correlations exhibit similar structure (Figs. \ref{fig:ALL_STR_nSTR} C,F, I). 
Note that structural connectivity alone
does not capture the complete view of the coordinated cofiring at a\textit{ wider level} across the hierarchical processing pipeline, which the proposed functional connectivity analysis reflects (Fig. \ref{fig:Resp_STR_nSTR}).

The 1FC ensembles display interesting properties: their aggregate cofiring reliably predicts downstream neuronal responses through a robust, ReLU-like input–output relationship, whose gain scales systematically with ensemble size. Notably, neuronal responses are characterized by two regimes—a largely inactive or weak-response regime and a strong-response regime. Reliable encoding of the presented class emerges only during high 1FC cofiring events, which themselves occur infrequently, indicating that informative representations are concentrated in rare but highly coordinated activity patterns.
Under uniform random noise or adversarial perturbations, keeping fixed the 1FC groups identified during the original testing, the response regimes of these neurons are disrupted, particularly in early and intermediate layers, with downstream degradation in class-specific encoding at later stages (Figs. \ref{fig:semantic_heatmaps_with_MI_original},\ref{fig:semantic_heatmaps_with_MI_noise},\ref{fig:semantic_heatmaps_with_MI_adv}). 

We also examined the relationship between what has been learned and functional connectivity by permuting the weights across edges within each layer and reapplying the functional connectivity analysis (See Fig. \ref{fig:shuffled_weights}). While the overall distribution of STTC weights and the proportion of statistically significant connections remain largely unchanged in early and intermediate layers, the characteristic response properties are lost. Moreover, deeper layers fail to form coherent 1FC groups altogether, underscoring the importance of functional connectivity for the emergence of meaningful ensembles.
This establishes 1FC ensembles as a functionally meaningful substrate for input encoding and information transfer. The response properties of \textit{targeted} neurons, selected based on their mutual information with their preferred class and 1FC size, at the onset of a strong-response regime, can be used to provide alerts about the information flow.
This framework thus, establishes a conceptual foundation for conducting diagnostics on the information transfer. 
\clearpage
\begin{ack}
This work has received funding from the European Union’s Horizon 2020 research and innovation program under the Marie Skłodowska-Curie grant agreement No 101007926 as well as from the Hellenic Foundation Research Institute (HFRI) with the neuron-AD project number 04058 and neuronXnet project number 2285 (PI: Maria Papadopouli). It has been partially supported by project MIS 5154714 of the National Recovery and Resilience Plan Greece 2.0 funded by the European Union under the NextGenerationEU Program. Finally, this research was also supported by R01 NS113890, and R21 NS127299 (PI: Stelios Smirnakis). AWS resources were provided by the National Infrastructures for Research and Technology GRNET and funded by the EU Recovery and Resiliency Facility.
\end{ack}

\bibliographystyle{unsrtnat}
\bibliography{references}
\clearpage

\appendix
\clearpage
\renewcommand{\thefigure}{A\arabic{figure}}
\renewcommand{\theHfigure}{A\arabic{figure}}  
\setcounter{figure}{0}
\section{Technical appendices and supplementary material}
\subsection{SNN Architecture and Training Details}\label{sec:SNN_arch_train}
From a practical standpoint, hierarchical representation learning necessitates sufficient network depth—something shallow or linear models cannot provide without a significant loss in classification performance. ResNet-18 offers a well-characterized hierarchy that is deep enough to enable systematic layer-by-layer analysis, while remaining computationally tractable.
From a theoretical perspective, convolutional architectures embody an inductive bias—fixed local connectivity combined with learned weight sharing—that serves as a structural analogue to the organization of the visual cortex. In biological systems, feedforward anatomical wiring gives rise to functionally distinct cell types. This parallel between imposed structural connectivity and emergent functional organization is central to our analysis.
Moreover, the hierarchical stages of CNNs and ResNet have been extensively mapped onto the ventral visual stream in prior work\cite{Yamins2016-ry, Wen2018-ho}, further supporting the use of SEW-ResNet as a suitable model system for probing the development of spiking representations. Our analysis focuses on the residual stream, tracing the direct flow of information through successive layers.

\textbf{Training Details} : 
The SEW-ResNet-18 was trained on CIFAR-100, which comprises 60,000 32×32 color images across 100 classes, partitioned into 50,000 training and 10,000 test images. The stem convolution was adapted from a $7\times7$, 128-channel configuration to a $3\times3$, 64-channel kernel to suit CIFAR-100 input dimensions. The feature of SEW ResNet is that the dropout layers also have the same structure, where if the stride of a layer is more than 1, the residual connections have the block structure (CONV$\rightarrow$BN$\rightarrow$PLIF) to downsample the input. Training was conducted for 320 epochs with a batch size of 256 images, using the SGD optimizer with a learning rate of $5 \times 10^{-4}$. Gradients were propagated through the non-differentiable spiking nonlinearity via surrogate gradient backpropagation, using the sigmoid surrogate function. The model checkpoint from epoch 314 was selected as the final model on the basis of best test accuracy, achieving $72.18\%$ on the CIFAR-100 test set.
Each image was presented across $T=4$ timesteps during both training and inference. The following data augmentation strategies were applied during training: random cropping with padding of 4 pixels, random horizontal flipping, and Gaussian blur (kernel size $3\times3$, $\sigma \in [0.05, 0.15]$,) applied with a probability of 0.5.

\textbf{Neuron Sampling} : For each convolutional spiking layer, spike trains were recorded during inference over the full CIFAR-100 test set. Rather than recording all neurons, we sample a spatially contiguous column of approximately 8,000 neurons per layer to balance computational tractability with representational coverage. Specifically, for a layer with shape 
$(C,H,W)$, a square spatial patch of side length $s \approx \lfloor\sqrt{8000/C}$ is extracted from the center of each feature map, where $s$ is clipped to not exceed H or W. This yields a set of $C \times s \times s$  neurons, uniformly covering all channels while remaining spatially localized to the central region of each feature map. In the final four convolutional spiking layers, where the spatial dimensions are sufficiently small, all 8192 neurons are included. The recorded spike trains from this sampled population form the basis for all functional connectivity, mutual information, and robustness analyses reported in this paper.

\textbf{Exclusion Criteria.} Neurons with a mean firing rate below 0.01 spikes per sample are designated as \textit{silent} and excluded from all analyses ($\sim 9.5\%$ excluded per layer). This firing-rate threshold was chosen to ensure sufficient spiking activity for reliable estimation of functional connectivity, response functions, and information-theoretic measures. Of the sampled neurons per layer, approximately $90.5\%$ meet this activity criterion and are designated as \textit{active} neurons; all subsequent analyses are conducted exclusively on this active population.
For analyses involving response-function fitting, additional $R^2$ thresholds were applied where indicated in the corresponding figures. These thresholds were selected to ensure reliable characterization of neuronal response regimes and were chosen to remain consistent with prior analyses performed in mouse visual cortex datasets.

\begin{figure}[h]
    \centering
    \begin{overpic}[width=\linewidth]{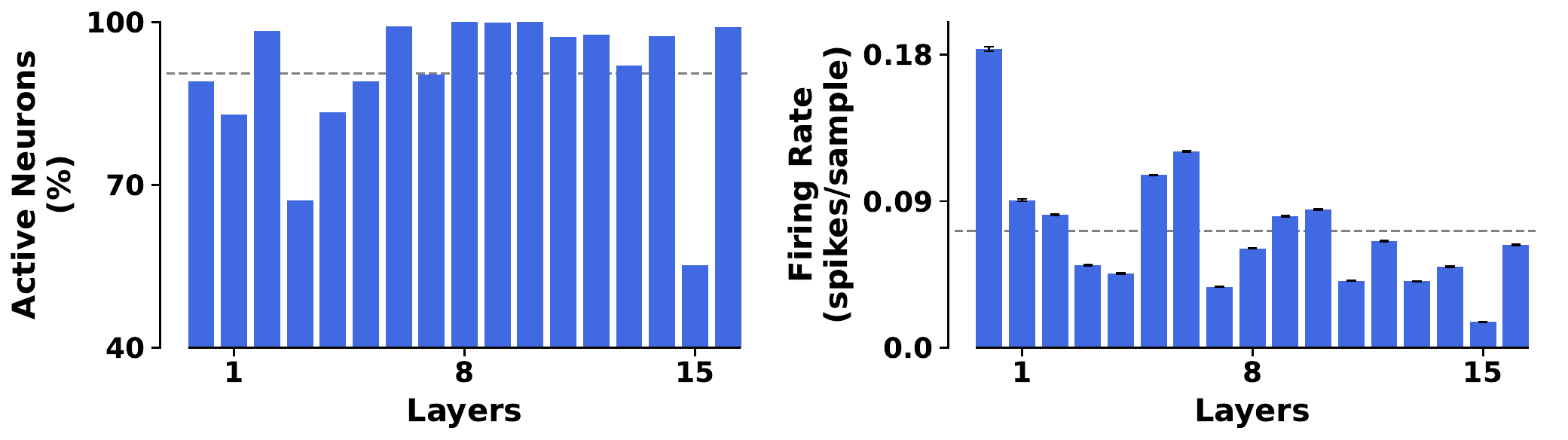}
        \put(2,28){\textbf{A}}
        \put(53,28){\textbf{B}}
    \end{overpic}
    \caption{\textbf{Neuronal activity per layer.} \textbf{(A)} Percentage of active neurons (neurons having firing rate > 0.01 spikes per sample) among all sampled neurons across layers. The grey dashed line indicates the mean activity level across layers. \textbf{(B)} Firing rate of active neurons, defined as the proportion of spikes per input sample for each layer. The grey dashed line indicates the mean firing rate across layers ($\sim 0.072$ spikes per sample). Approximately $9.5 \%$ of neurons (out of $\sim 8300$ per layer) are excluded due to inactivity. We observe that neuronal responses are inherently sparse, despite no explicit sparsity constraint during training.}
    \label{fig:firing_rates_SNN}
\end{figure}

\subsection{STTC}
\label{sec:A_STTC}
We use the spike time tiling coefficient (STTC), introduced by \citet{cutts2014detecting}, which performs favorably against 33 other commonly-used measures and is less sensitive to firing rate. The STTC of neuron A relative to B is defined as:
\begin{equation}
    \text{STTC}(\Delta t) = \frac{1}{2} \left ( \frac{P_A - T_B}{1 - P_A T_B} + \frac{P_B - T_A}{1 - P_B T_A}\right )
\end{equation}
where $P_A$ is the proportion of firing events of neuron $A$ found within an interval ($\pm \Delta t$) around each firing event of
neuron $B$, $T_B$ is the proportion of the total observed spikes that falls within an interval $\pm \Delta t$ around each firing event
of neuron $B$, and likewise for $T_A$ and $P_B$. STTC is a symmetric measure and is normalized to lie within $[-1, 1]$.
STTC is defined with respect to a measurement window of synchronization $\Delta t$. Since we use an artificial SNN where we know the exact response timing, we set $\Delta t$= 0. 

\begin{figure}[H]
    \centering
    \begin{overpic}[width=\linewidth]{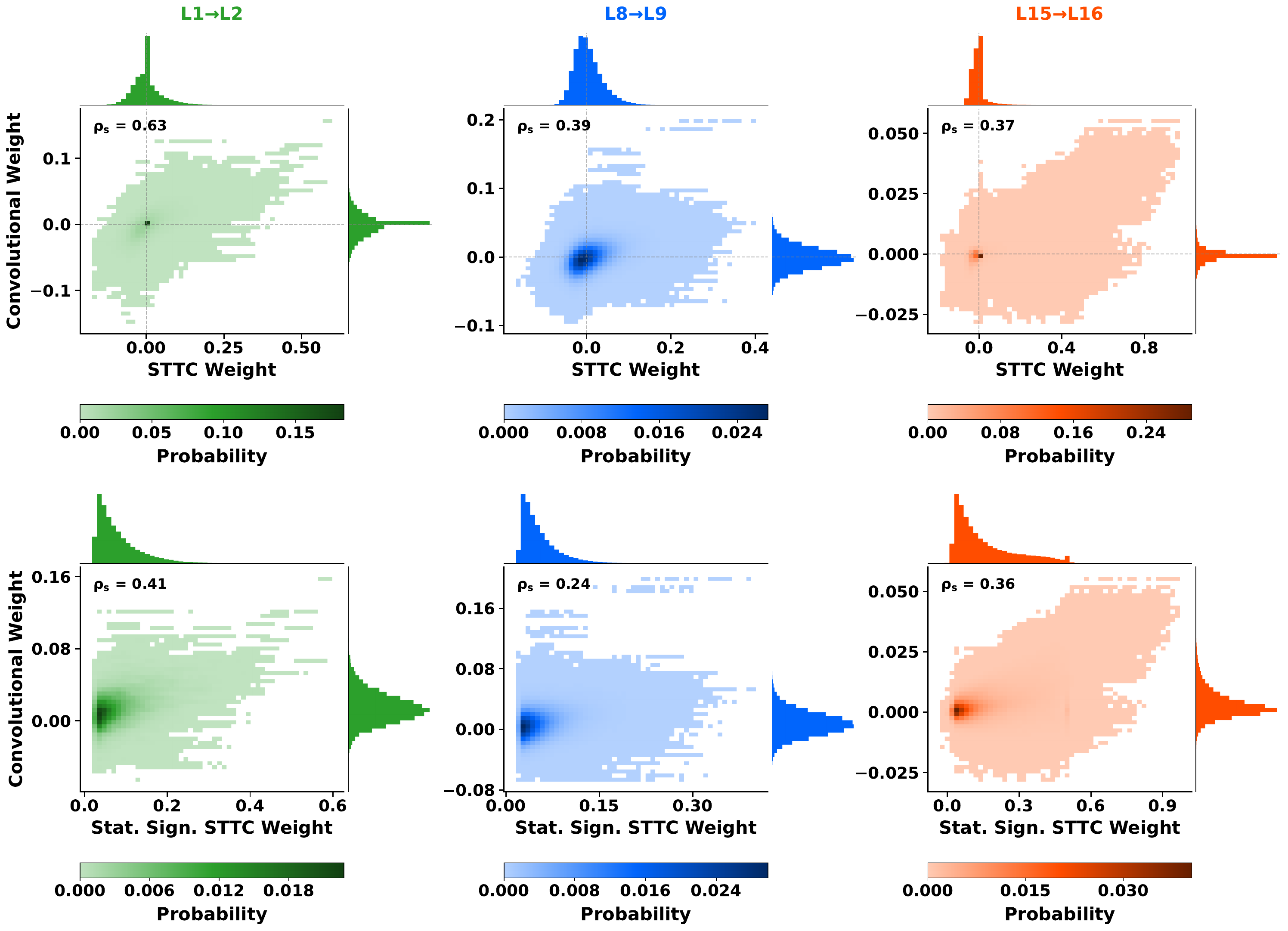}
        \put(3,72){\textbf{A}}
        \put(35,72){\textbf{B}}
        \put(67,72){\textbf{C}}
        \put(3,34){\textbf{D}}
        \put(35,34){\textbf{E}}
        \put(67,34){\textbf{F}}
\end{overpic}
    \caption{\textbf{Structural Connections: Convolutional kernel weight vs. inter-layer STTC weight}. \textbf{(A)} Two-dimensional histogram of convolutional weights versus corresponding inter-layer STTC weights for Layer 1$\rightarrow$2; shading (light to dark) indicates probability of occurrence. Marginal histograms (top and right) show the distributions of STTC and convolutional weights, respectively. \textbf{(B)}, \textbf{(C)} Same as \textbf{(A)} for Layer pairs 8$\rightarrow$9 and 15$\rightarrow$16. \textbf{(D-F)} Same as \textbf{(A-C)}, respectively, restricted to positive statistically significant edges (z > 4). Inset: Spearman's rank correlation between STTC weight and convolutional weight.Across all examined layer pairs, the joint distributions reveal a consistent alignment between STTC weights and the corresponding convolutional kernel weights; however, this correspondence is modestly attenuated when restricted to statistically significant positive edges \textbf{(D-F)}, suggesting that strong functional correlations are not exclusively confined to connections with large learned weights.}

    \label{fig:BP_vs_STTC_weights}
\end{figure}

\begin{figure}[H]
    \centering
    \begin{overpic}[width=\linewidth]{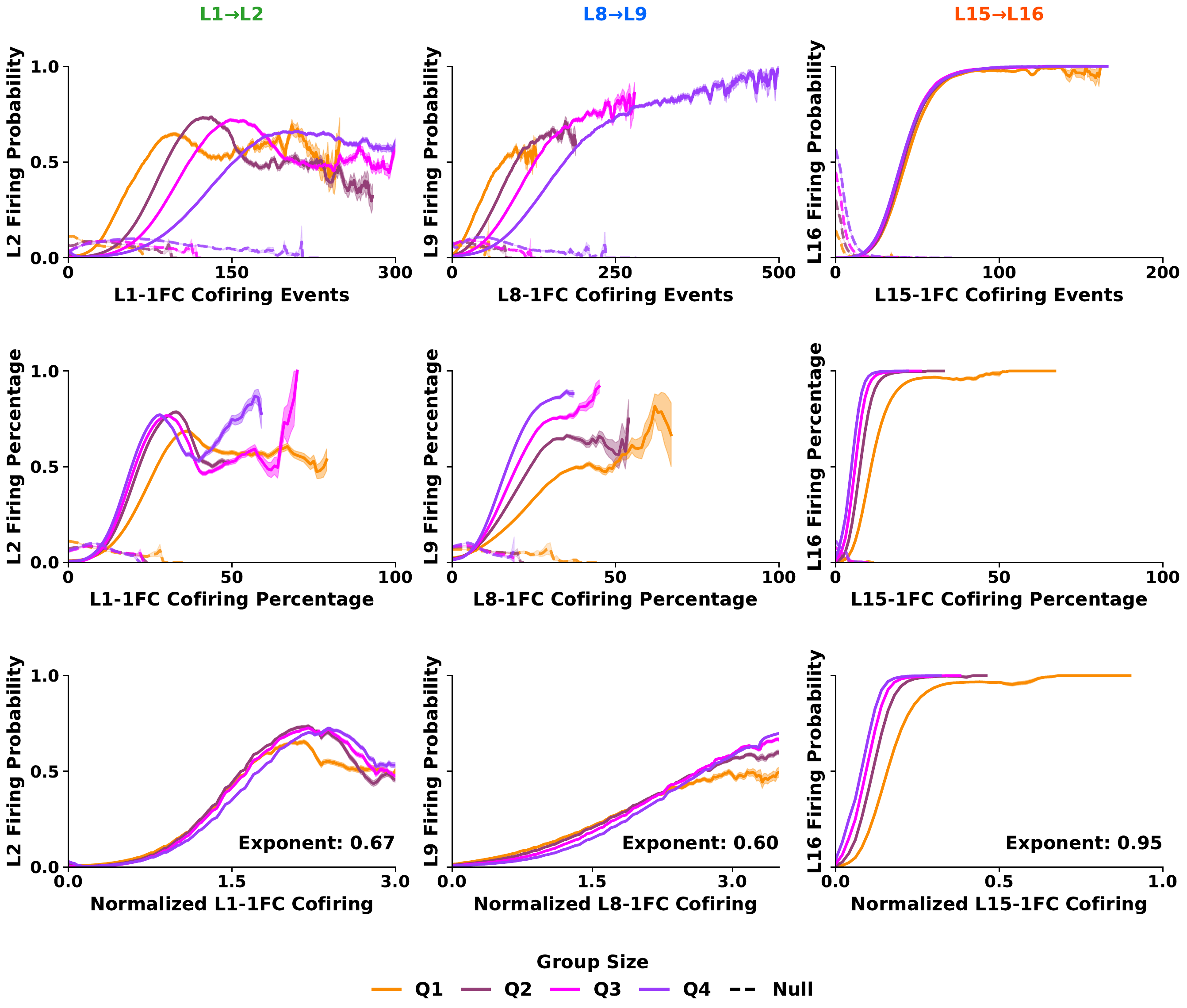}
        \put(3,84){\textbf{A}}
        \put(35,84){\textbf{B}}
        \put(68,84){\textbf{C}}
        \put(3,58){\textbf{D}}
        \put(35,58){\textbf{E}}
        \put(68,58){\textbf{F}}
        \put(3,32){\textbf{G}}
        \put(35,32){\textbf{H}}
        \put(68,32){\textbf{I}}
\end{overpic}
    \caption{\textbf{Response functions across group size conditions.} \textbf{(A--C)} Aggregate firing probability of target-layer neurons as a function of input 1FC co-firing events, stratified by group size quartiles (Q1–Q4), for layer pairs L2$\rightarrow$L3 \textbf{(A)}, L8$\rightarrow$L9 \textbf{(B)}, and L15$\rightarrow$L16 \textbf{(C)}. Functional groups (1FC) are defined once using the full test dataset and held fixed; the null condition (dashed gray) shows the mean $\pm$ SEM of size-matched control groups without significant functional correlations. \textbf{(D--F)} Same as \textbf{(A--C)}, with responses expressed as a function of the percentage of co-firing events (normalized by total group size). \textbf{(G--I)} Same as \textbf{(A--C)}, but co-firing is normalized by a normalization factor - $ N^{\alpha}$, with N as the group size and $\alpha = 0.67$, $0.6$, and $0.95$ for L2$\rightarrow$L3, L8$\rightarrow$L9, and L15$\rightarrow$L16, respectively.}
    \label{fig:Resp_Group_Sizes}
\end{figure}

\subsection{Temporal Robustness of Responses} \label{sec:temp_robustness}
\begin{figure}[H]
    \centering
    \begin{overpic}[width=\linewidth]{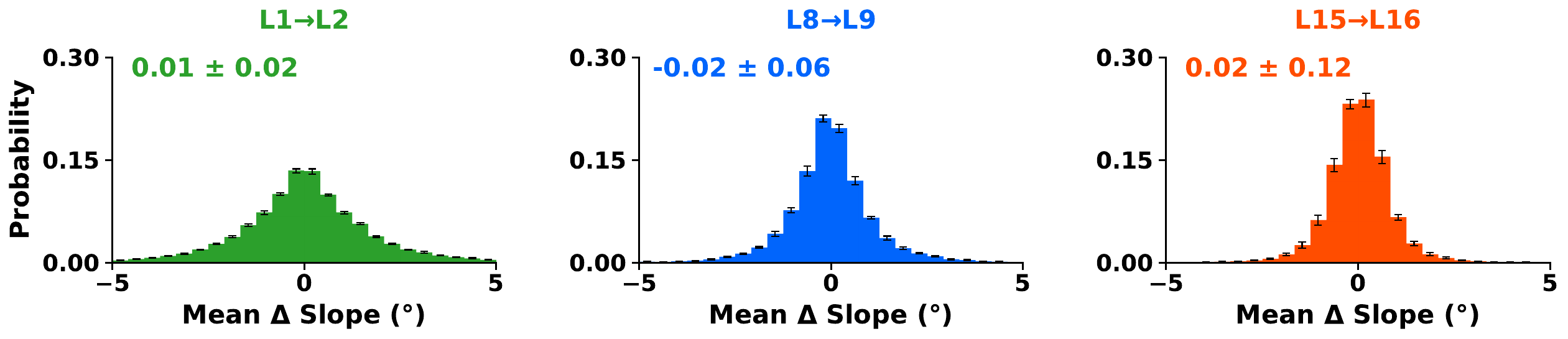}
        \put(3,20){\textbf{A}}
        \put(35,20){\textbf{B}}
        \put(68,20){\textbf{C}}
\end{overpic}
    \caption{\textbf{Stability of response function slopes across batches.} For layer pairs L1$\rightarrow$L2 \textbf{(A)}, L8$\rightarrow$L9 \textbf{(B)}, and L15$\rightarrow$L16 \textbf{(C)}, we quantify variability in neuron-wise response function slopes across input batches. 1FC input functional groups are defined on the full dataset and held fixed. For each neuron, slopes are computed independently for each of four batches (B1,...,B4). Pairwise differences in slopes are then evaluated across all 6 batch combinations (e.g., B2$-$B1, B3$-$B1). For each batch pair, we obtain a distribution of slope differences across neurons; panels show the mean $\pm$ SEM of these distributions aggregated over all batch pairs. Inset text shows the mean ($\pm$ SD) of the distributions across batch pairs. Small mean differences indicate that response slopes are stable across batches. Neurons whose response-curve fits yielded $R^2 \leq 0.8$ were excluded from the analysis.(2424, 5946 and 11 neurons were excluded from the analysis for L1$\rightarrow$L2, L8$\rightarrow$L9, L15$\rightarrow$L16 respectively.)}
    \label{fig:Slopes_Temp_Robust}
\end{figure}

To verify that the 1FC groups are not an artifact of the specific images used to estimate functional connectivity, we assess the temporal robustness of neuronal firing response functions across held-out input samples. For each neuron, we compute firing response functions over four independent batches of test images and compare the resulting slopes, retaining the same 1FC group assignments throughout. The distribution of slope differences across all four batches remains tightly concentrated around zero, with a negligible proportion of outliers (Fig.~\ref{fig:Slopes_Temp_Robust}). The consistency of response function slopes across batches and across the full range of input images confirm that the 1FC groups reflect stable functional structure rather than sample-specific co-activation.
\begin{figure}[H]
    \centering
    \begin{overpic}[width=\linewidth]{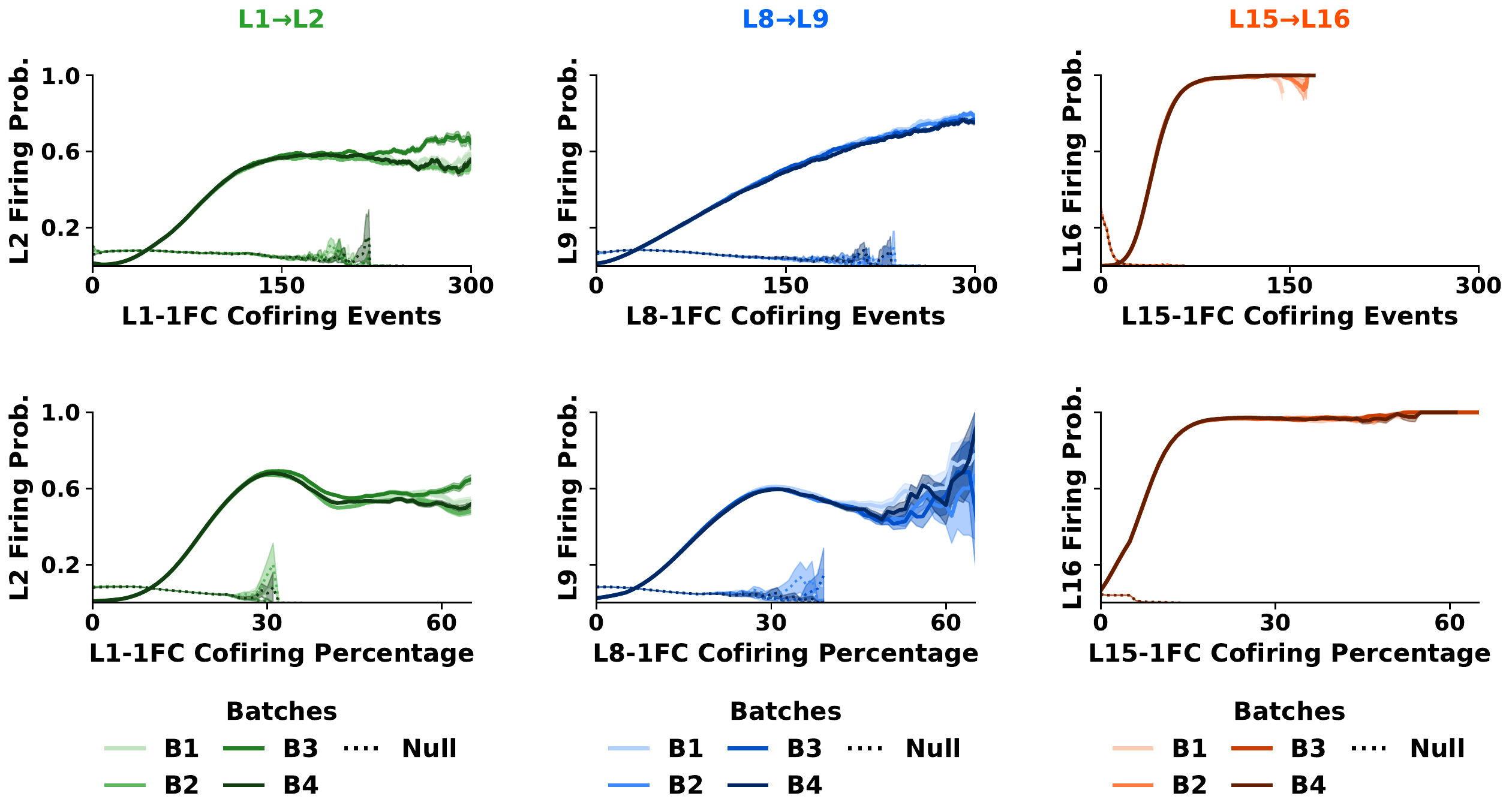}
        \put(3,51){\textbf{A}}
        \put(35,51){\textbf{B}}
        \put(68,51){\textbf{C}}
        \put(3,28){\textbf{D}}
        \put(35,28){\textbf{E}}
        \put(68,28){\textbf{F}}
\end{overpic}
    \caption{\textbf{Temporal robustness of spiking responses.} \textbf{(A--C)} Aggregate firing probability of target-layer neurons as a function of input 1FC co-firing events, evaluated for layer pairs L2$\rightarrow$L3 \textbf{(A)}, L8$\rightarrow$L9 \textbf{(B)}, and L15$\rightarrow$L16 \textbf{(C)}. Functional groups (1FC) are defined once using the full test dataset and held fixed; firing responses are then computed separately across four input batches (light to dark shading). Shaded regions denote SEM. The null condition (dotted gray) shows the mean $\pm$ SEM of size-matched control groups drawn from neurons without statistically significant functional correlations. \textbf{(D--F)} Same as \textbf{(A--C)}, shown as a function of the percentage of co-firing events. Consistent responses across batches indicate that firing probabilities are not biased by input composition or batch order.}
    \label{fig:Resp_Temp_Robust}
\end{figure}

\subsection{Mutual Information}
A \emph{one-hot encoding} is a mapping from a categorical set to binary vectors.
Let $\mathcal{C} = \{c_1, \dots, c_K\}$ be a set of $K$ categories (here classes of input images). The one-hot encoding is the function
\[
\phi : \mathcal{C} \to \{0,1\}^K
\]
defined by
\[
\phi(c_i) = e_i,
\]
where $e_i \in \mathbb{R}^K$ is the $i$-th standard basis vector, i.e.,
\[
(e_i)_j =
\begin{cases}
1 & \text{if } j = i, \\
0 & \text{otherwise.}
\end{cases}
\]
To quantify the informativeness of individual neurons, we computed the mutual information (MI) between each neuron's spike train and the true class labels across the test set. 
MI was calculated between the binary spiking activity of each neuron (spike or no spike) and both the categorical class labels and their one-hot encoded representations, the latter yielding a class-wise MI profile for each neuron. MI is defined as 
\begin{equation}
I(X;Y) = H(X) - H(X|Y),
\end{equation} where $H(X)$ denotes the Shannon entropy of the neuron's spiking activity and $H(X|Y)$ is the conditional entropy given the class label, such that MI reflects the reduction in uncertainty about a neuron's firing upon observation of the true class. 

\begin{figure}[H]
    \centering
    \begin{overpic}[width=\linewidth]{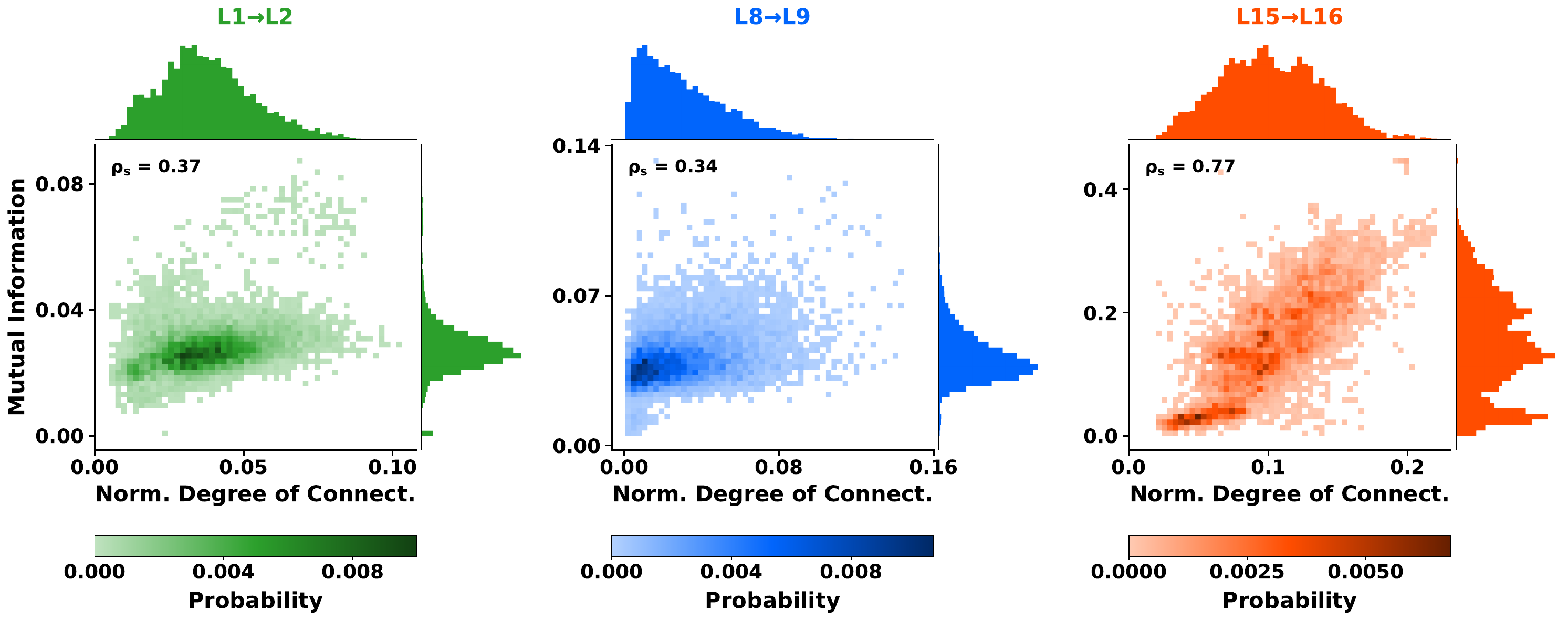}
        \put(3,38){\textbf{A}}
        \put(35,38){\textbf{B}}
        \put(67,38){\textbf{C}}
\end{overpic}
    \caption{\textbf{Relationship between normalized degree of connectivity and mutual information}. \textbf{(A)} Two-dimensional histogram of mutual information versus corresponding normalized degree of connectivity for Layer 1$\rightarrow$2; shading (light to dark) indicates probability of occurrence. Marginal histograms (top and right) show the distributions of normalized degree of connectivity and mutual information, respectively. \textbf{(B)}, \textbf{(C)} Same as \textbf{(A)} for Layer pairs 8$\rightarrow$9 and 15$\rightarrow$16. Especially in L16, the higher the degree of connectivity, the larger the MI. A cluster of hub neurons with high degree of connectivity and MI is prominent. Spearman correlations are reported in each plot. Positive correlation between MI and DoC is most pronounced in the final layer pair \textbf{(C)}, suggesting that deeper layers progressively concentrate class-discriminative information within a subset of highly interconnected neurons.}
    \label{fig:MI_vs_DoC}
\end{figure}

\subsection{Differences in functional edge types: structural and non-structural edges.}
\begin{figure}[H]
\centering
\begin{overpic}[width=0.8\linewidth]{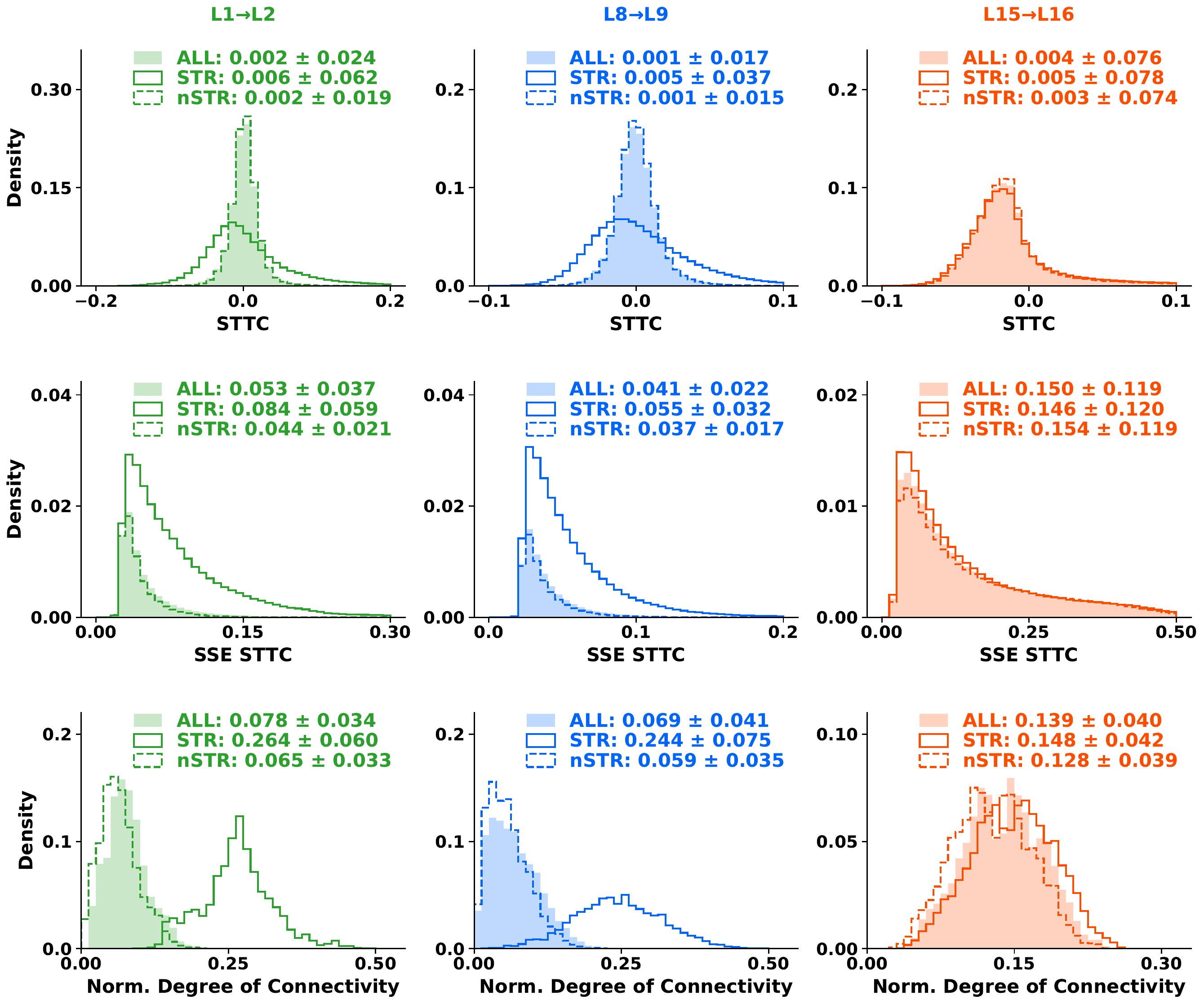}
  \put(2,82){\textbf{A}}
  \put(34,82){\textbf{B}}
  \put(67,82){\textbf{C}}
  \put(2,53){\textbf{D}}
  \put(34,53){\textbf{E}}
  \put(67,53){\textbf{F}}
  \put(2,25){\textbf{G}}
  \put(34,25){\textbf{H}}
  \put(67,25){\textbf{I}}
\end{overpic}
\caption{\textbf{Inter-Layer First-Order Functionally Connected (1FC) Groups.} \textbf{(A-C)} show pair-wise STTC weights (ALL; filled), structural edges (STR; solid line), and non-structural edges (nSTR; dashed line) for layer-pairs L2$\rightarrow$L3, L9$\rightarrow$L10, and L16$\rightarrow$L17 respectively. \textbf{(D-F)} show the STTC weights of statistically significant edges (SSEs) having z-score $> 4$ for the three connection types. \textbf{(G-I)} shows the distribution of the normalized degree of connectivity (DoC) of each output neuron. For ALL edges, degree is expressed as a fraction of the total number of neurons in the source layer. For structural edges, normalization is with respect to the total number of possible structural connections defined by the convolutional kernel ($c_{\text{in}} \times k \times k$). For non-structural edges, normalization is with respect to the remaining possible connections (ALL $-$ STR). Early and middle layer pairs show higher structural connectivity, whereas in the final layer pair structural and non-structural contributions become comparable, indicating increased purely functional connectivity in deeper layers.}
\label{fig:ALL_STR_nSTR}
\end{figure}
\begin{figure}[H]
    \centering
    \begin{overpic}[width=\linewidth]{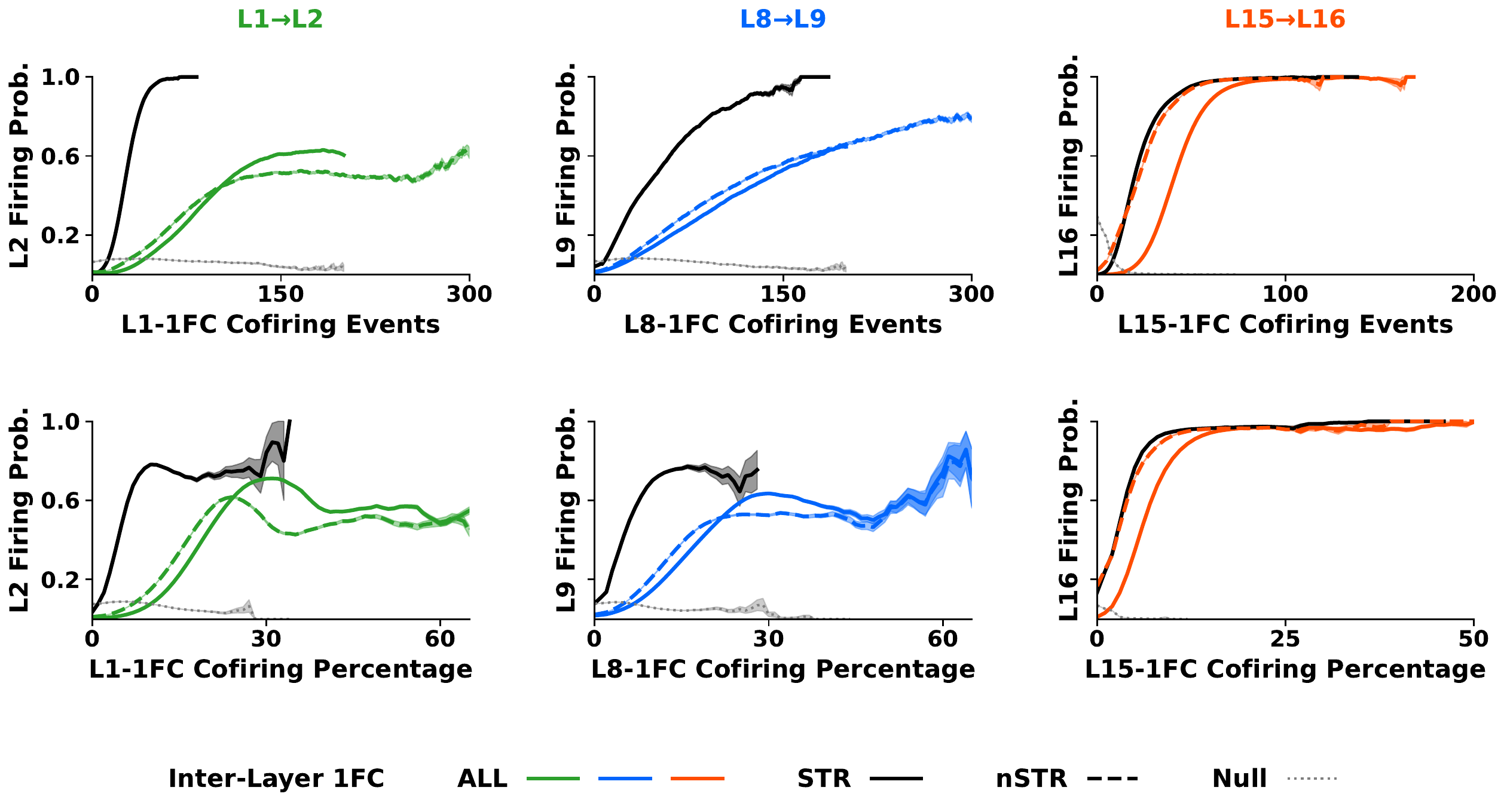}
        \put(3,51){\textbf{A}}
        \put(35,51){\textbf{B}}
        \put(68,51){\textbf{C}}
        \put(3,28){\textbf{D}}
        \put(35,28){\textbf{E}}
        \put(68,28){\textbf{F}}
\end{overpic}
    \caption{\textbf{ 
    Aggregate neuronal responses as a function of each neuron’s 1FC neighbors from the previous layer, under three cases: all its 1FC neighbors (ALL), only those with structural connections (STR), and the remaining ones (nSTR).}
\textbf{(A)} For each L2 neuron, we estimated its firing probability as a function of co-firing events among its all its L1-FC neighbors ALL; solid green), only its STR L1-1FC neighbors (solid black), and only its nSTR (dashed green). Results are aggregated across all L2 neurons. Shaded regions indicate the SEM. The null condition (dotted gray) represents the mean ± SEM of a control group matched in size to the 1FC set, sampled from neurons without statistically significant functional correlations. \textbf{(D)} Same as \textbf{(A)}, but plotted as a function of the percentage of co-firing events. \textbf{(B--C, E--F)} Corresponding analyses for layer pairs L8$\rightarrow$L9 layer connectivity and L15$\rightarrow$L16 layer connectivity, respectively. All neurons were included.}
    \label{fig:Resp_STR_nSTR}
\end{figure}

\subsection{Impact of Identity of Neurons in 1FC}\label{sec:membership-1FC}
\begin{figure}[h!]
\centering
\begin{tabular}{cccc}
\begin{overpic}[width=0.20\linewidth]{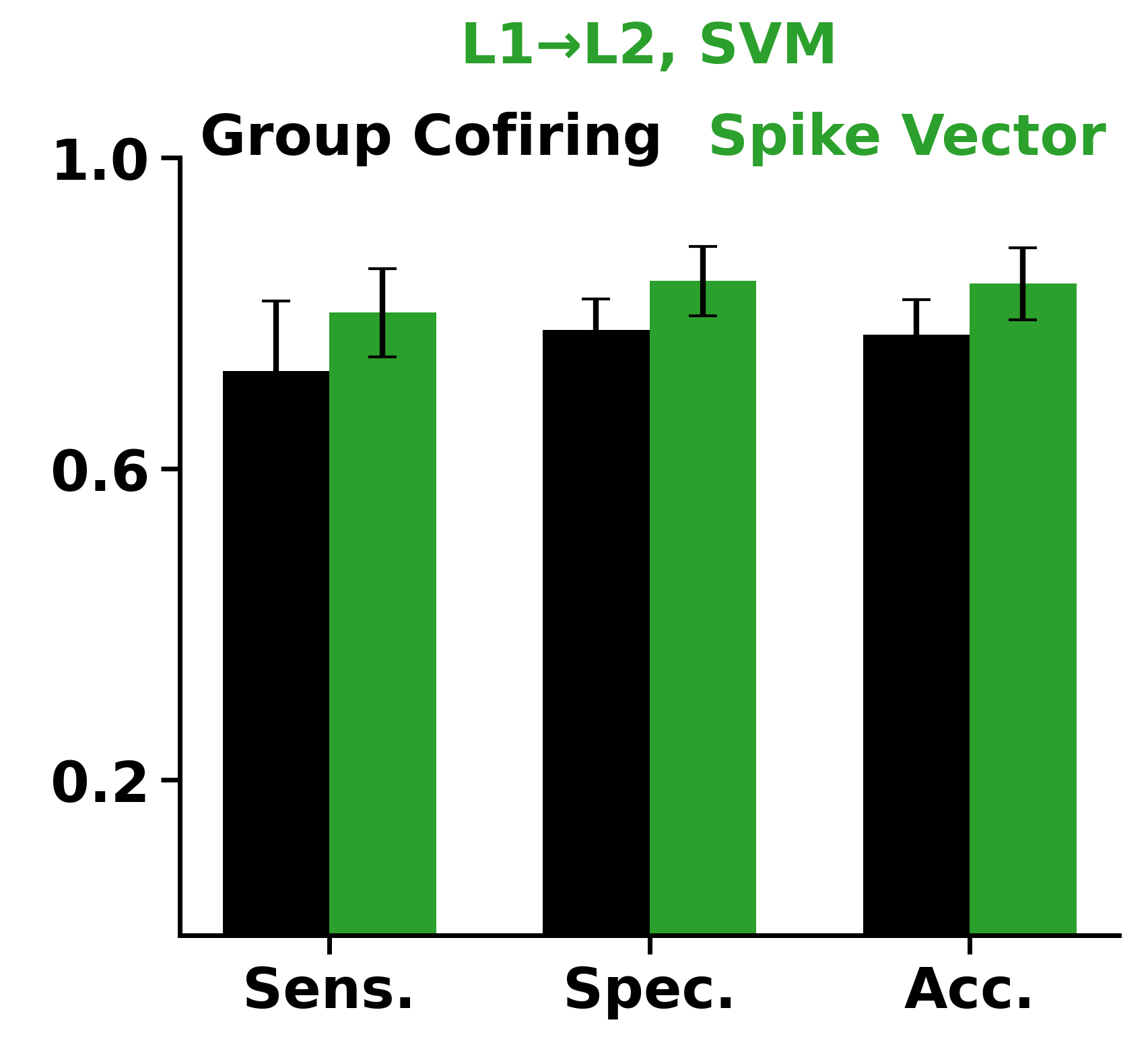}
\put(2,90){\textbf{A}}
\end{overpic} &
\begin{overpic}[width=0.20\linewidth]{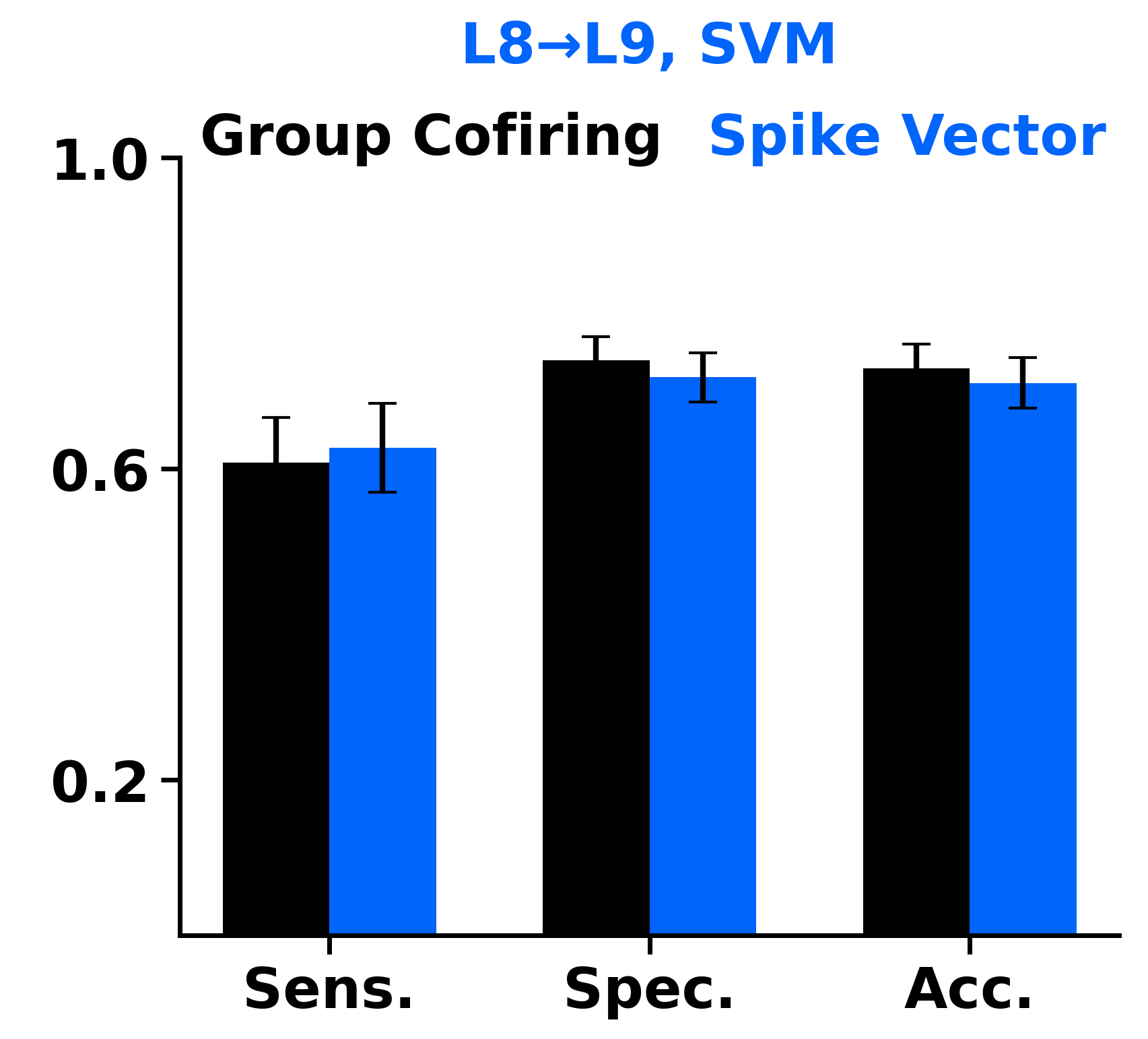}
\put(2,90){\textbf{B}}
\end{overpic} &
\begin{overpic}[width=0.20\linewidth]{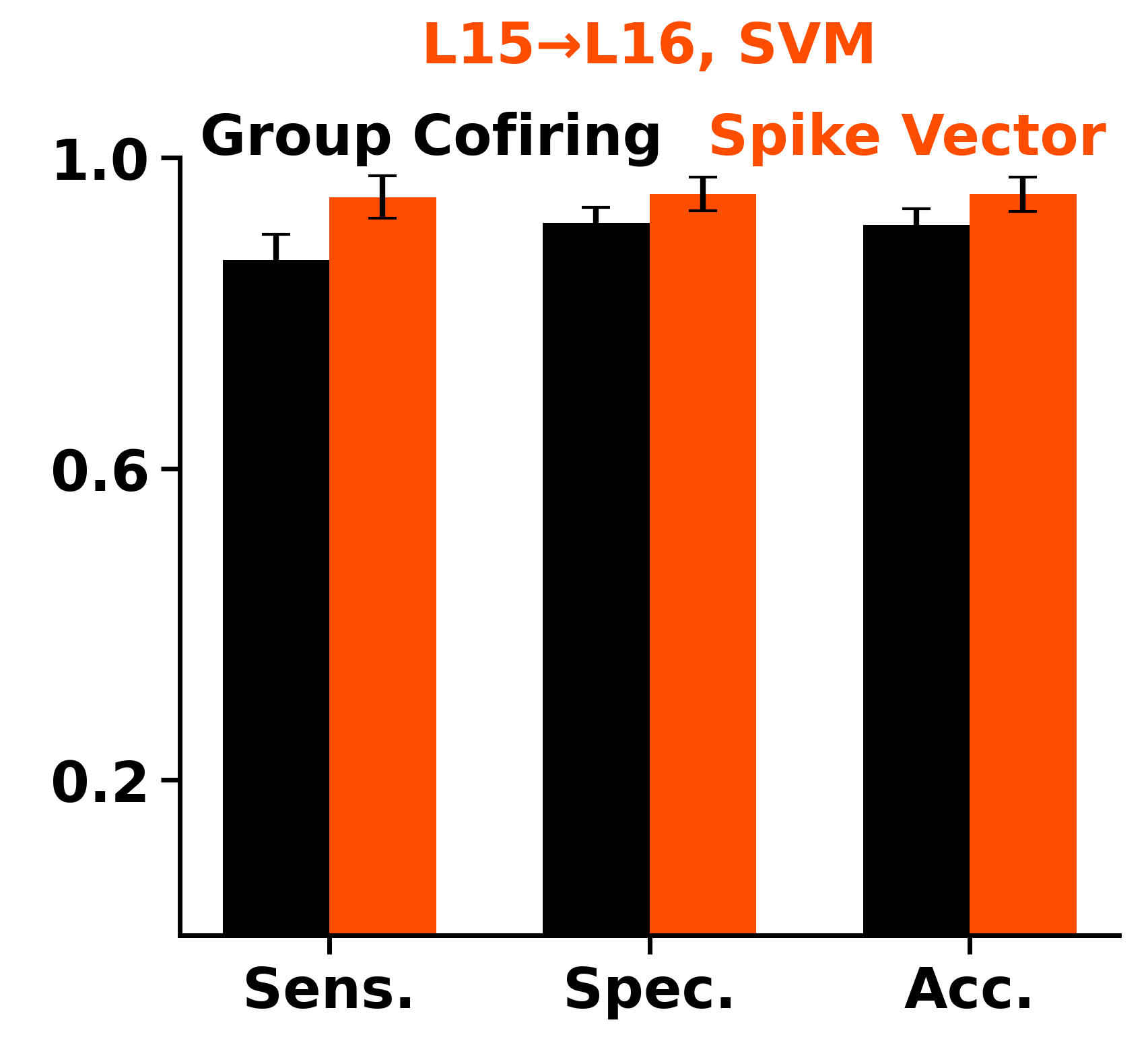}
\put(2,90){\textbf{C}}
\end{overpic} &
\begin{overpic}[width=0.20\linewidth]{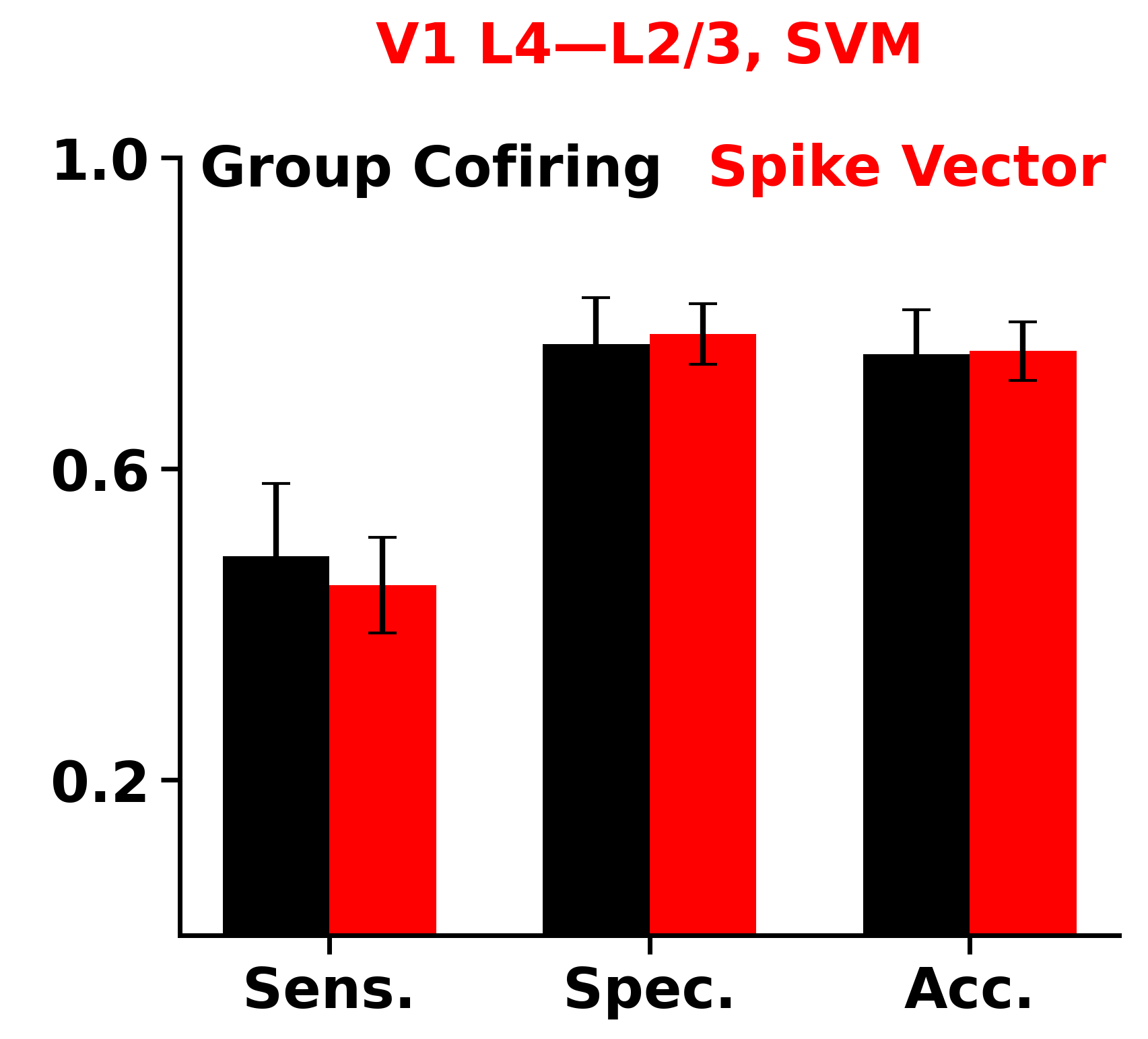}
\put(2,90){\textbf{D}}
\end{overpic}\\

\begin{overpic}[width=0.20\linewidth]{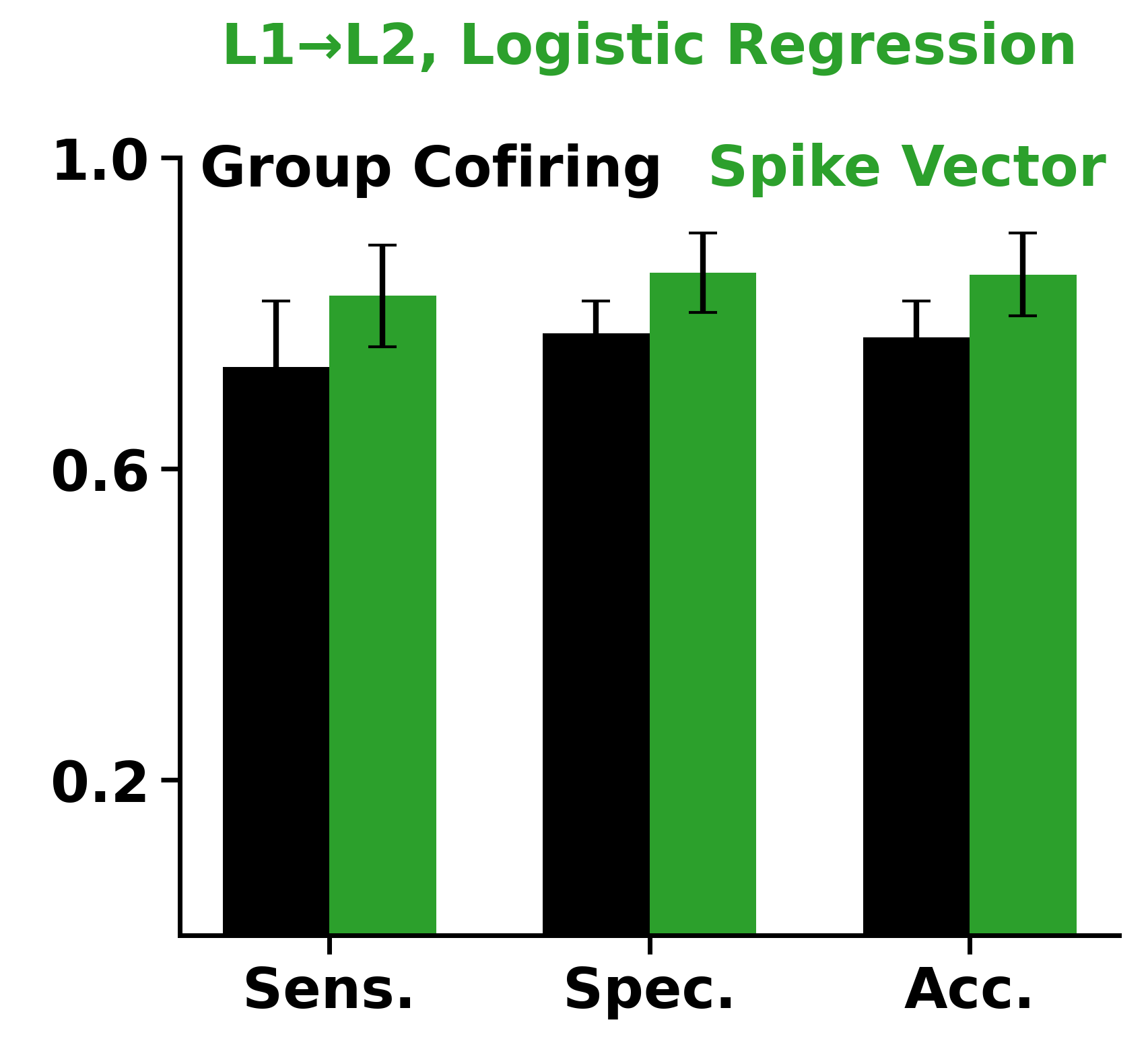}
\put(2,90){\textbf{E}}
\end{overpic} &
\begin{overpic}[width=0.20\linewidth]{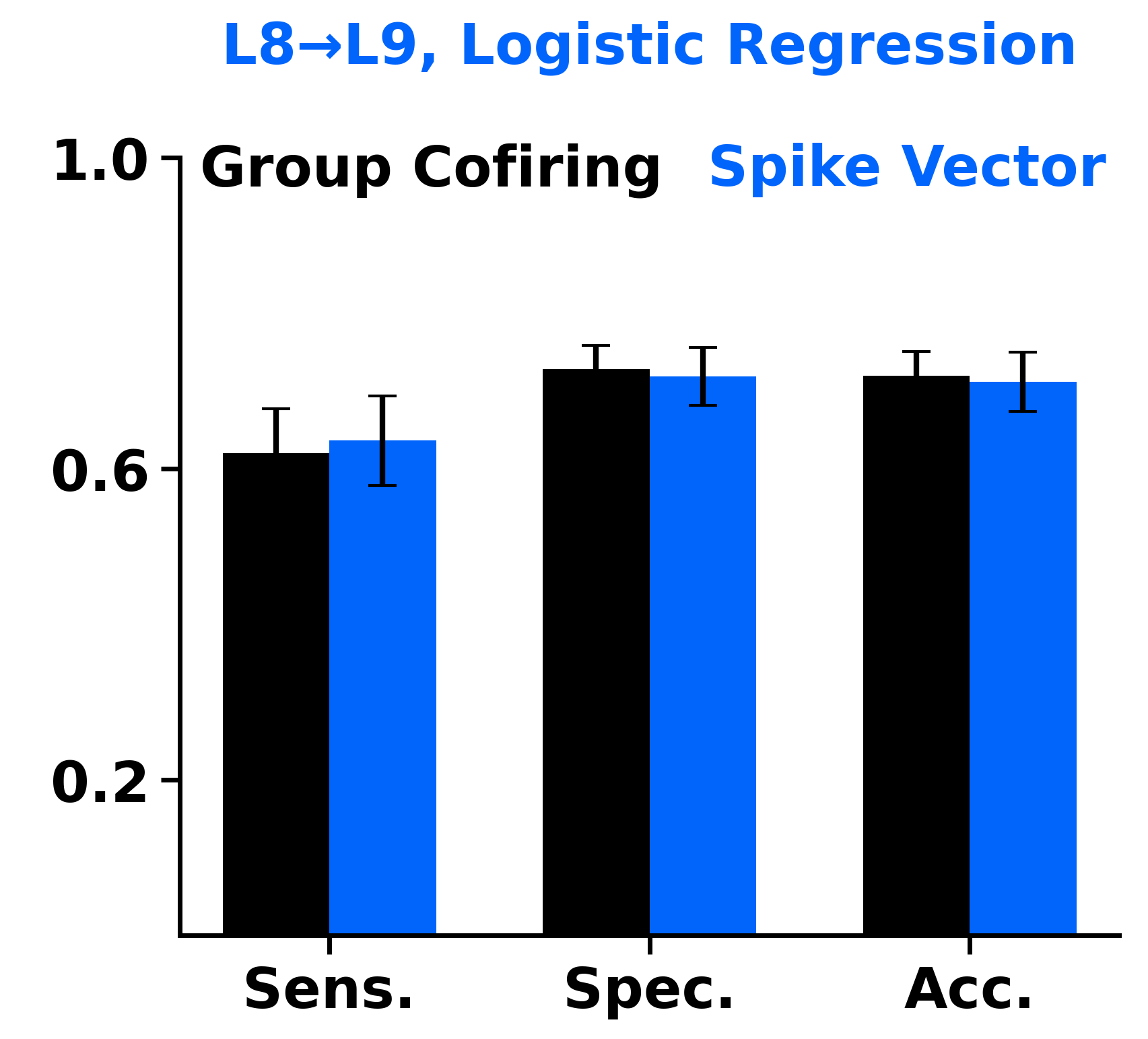}
\put(2,90){\textbf{F}}
\end{overpic} &
\begin{overpic}[width=0.20\linewidth]{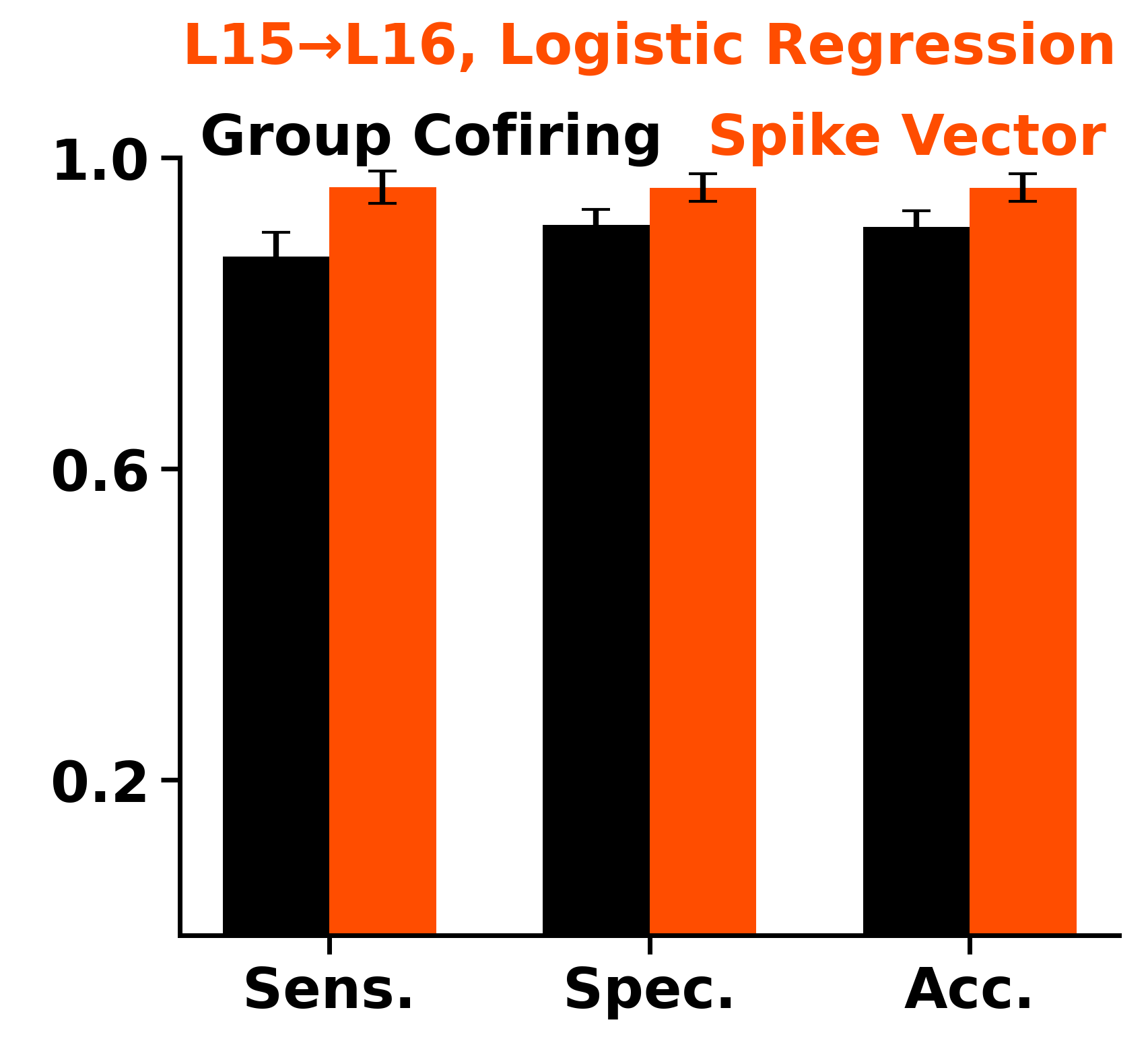}
\put(2,90){\textbf{G}}
\end{overpic} &
\begin{overpic}[width=0.20\linewidth]{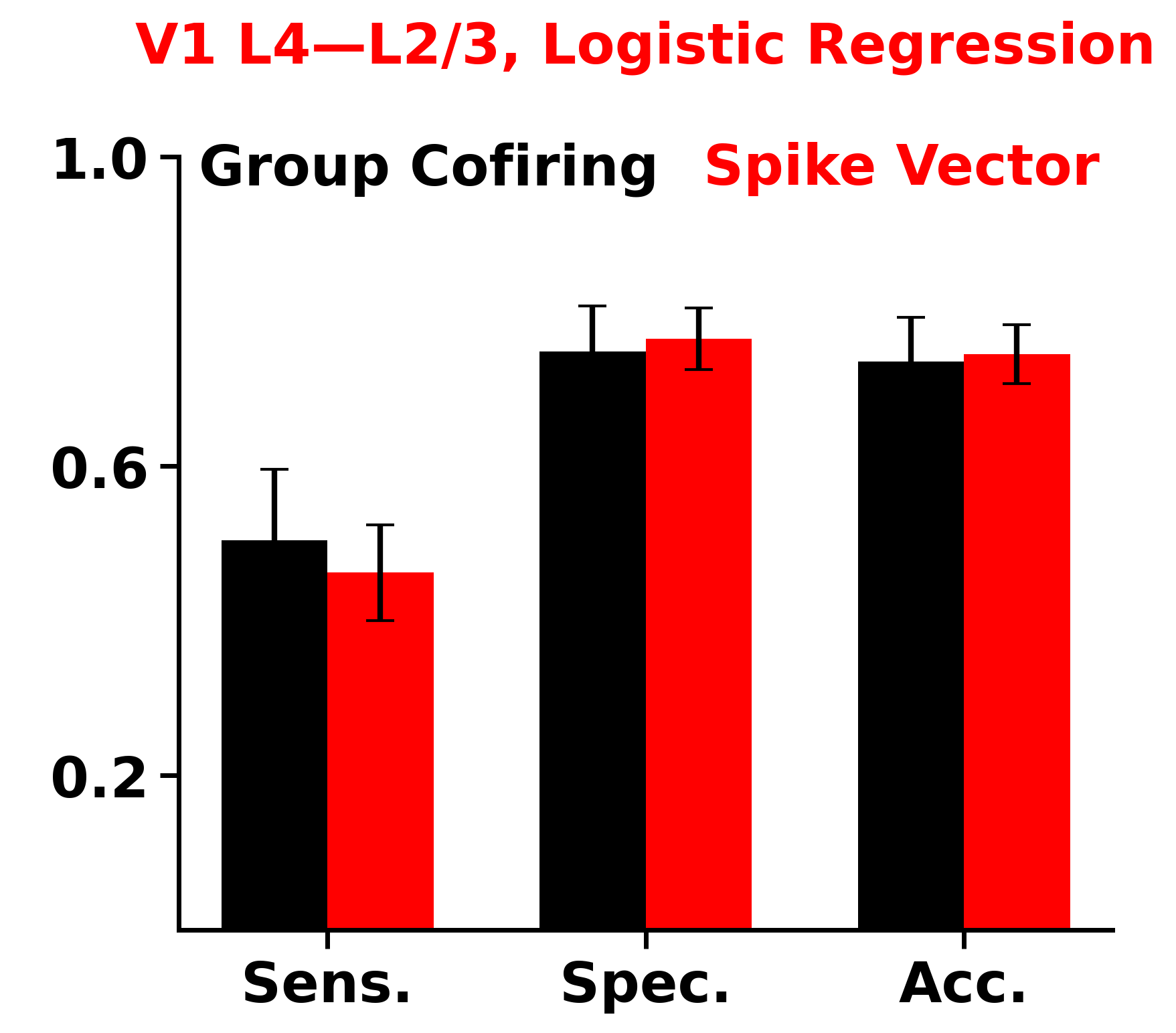}
\put(2,90){\textbf{H}}
\end{overpic} \\

\begin{overpic}[width=0.20\linewidth]{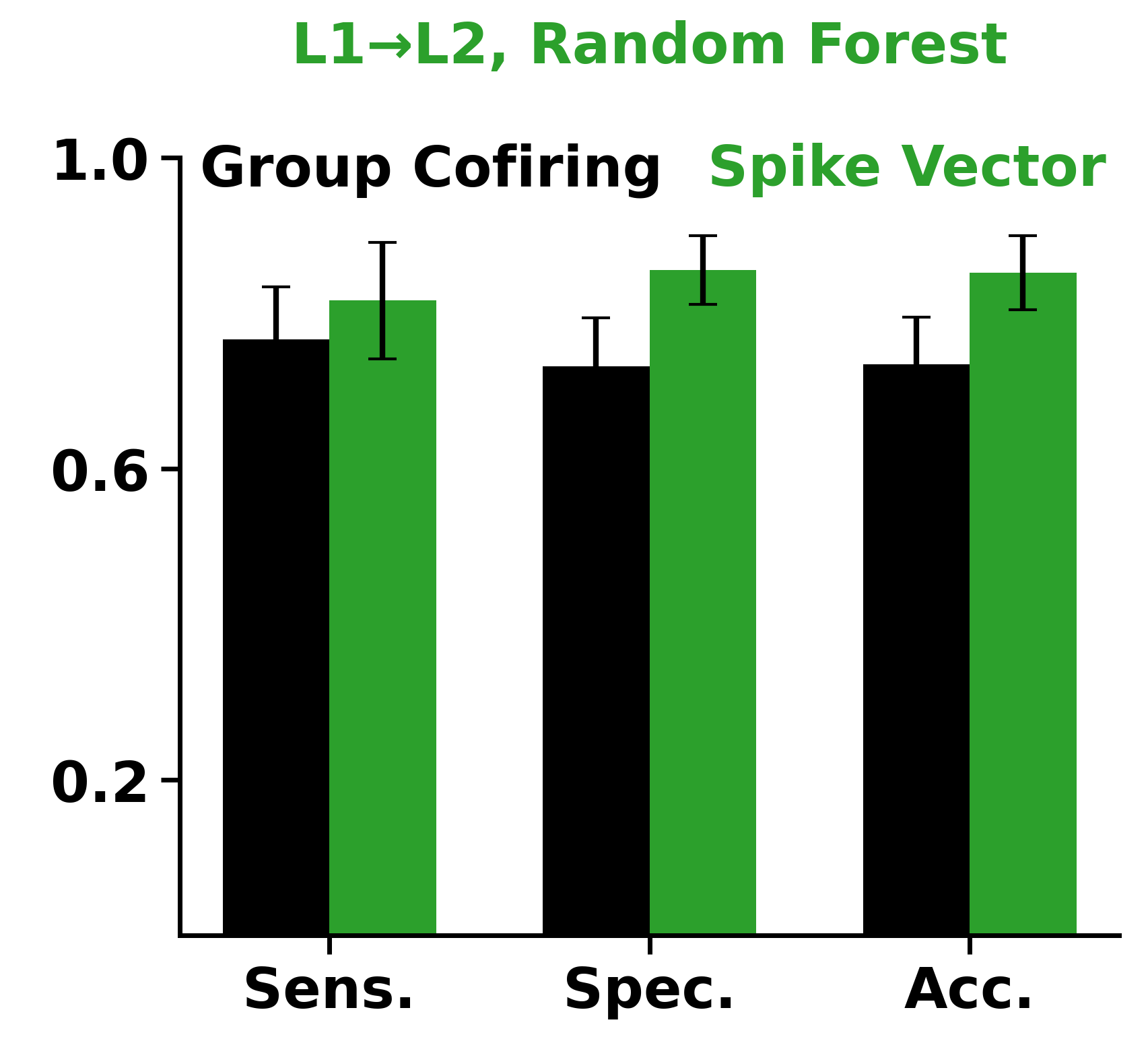}
\put(2,90){\textbf{I}}
\end{overpic} &
\begin{overpic}[width=0.20\linewidth]{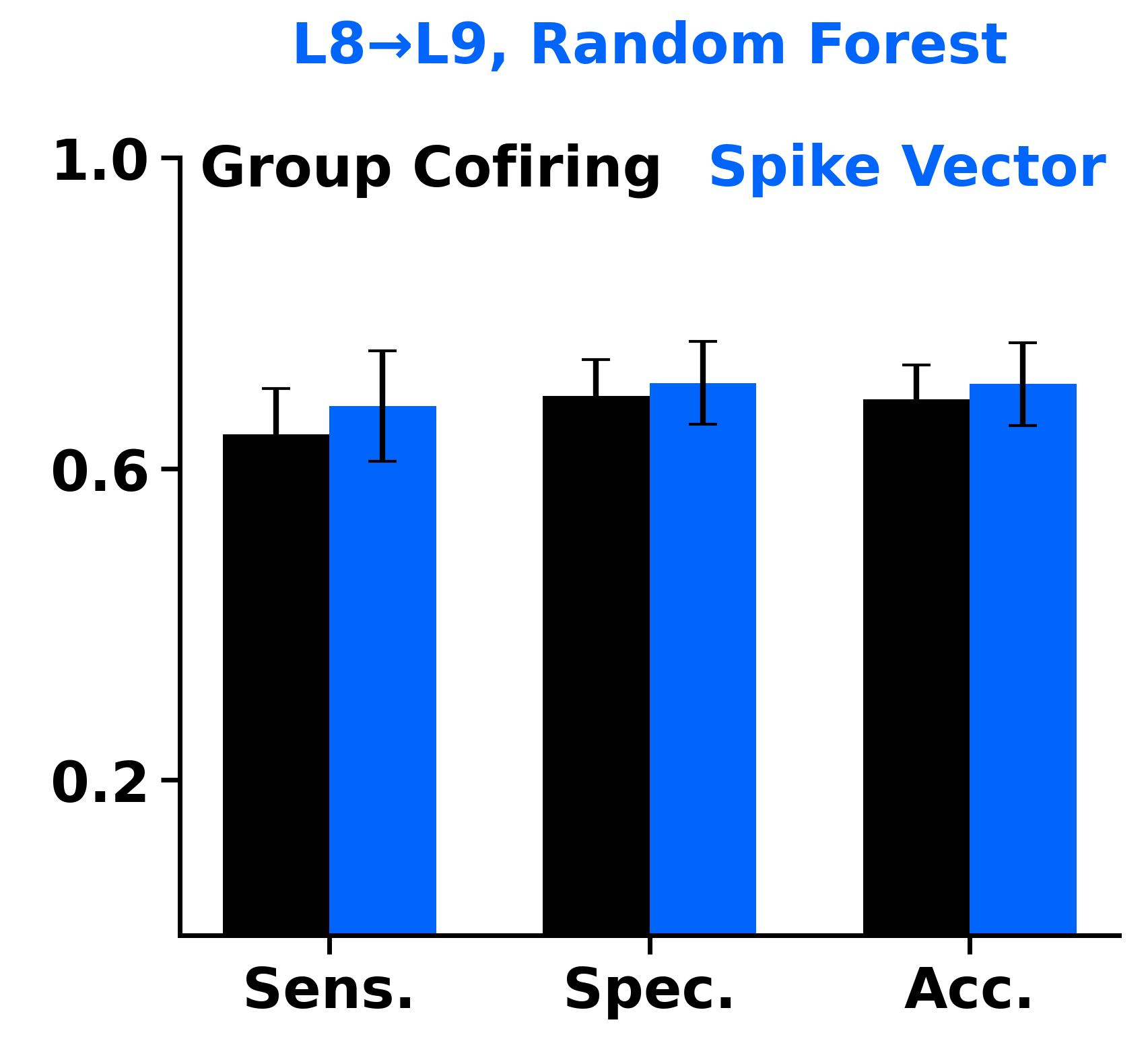}
\put(2,90){\textbf{J}}
\end{overpic} &
\begin{overpic}[width=0.20\linewidth]{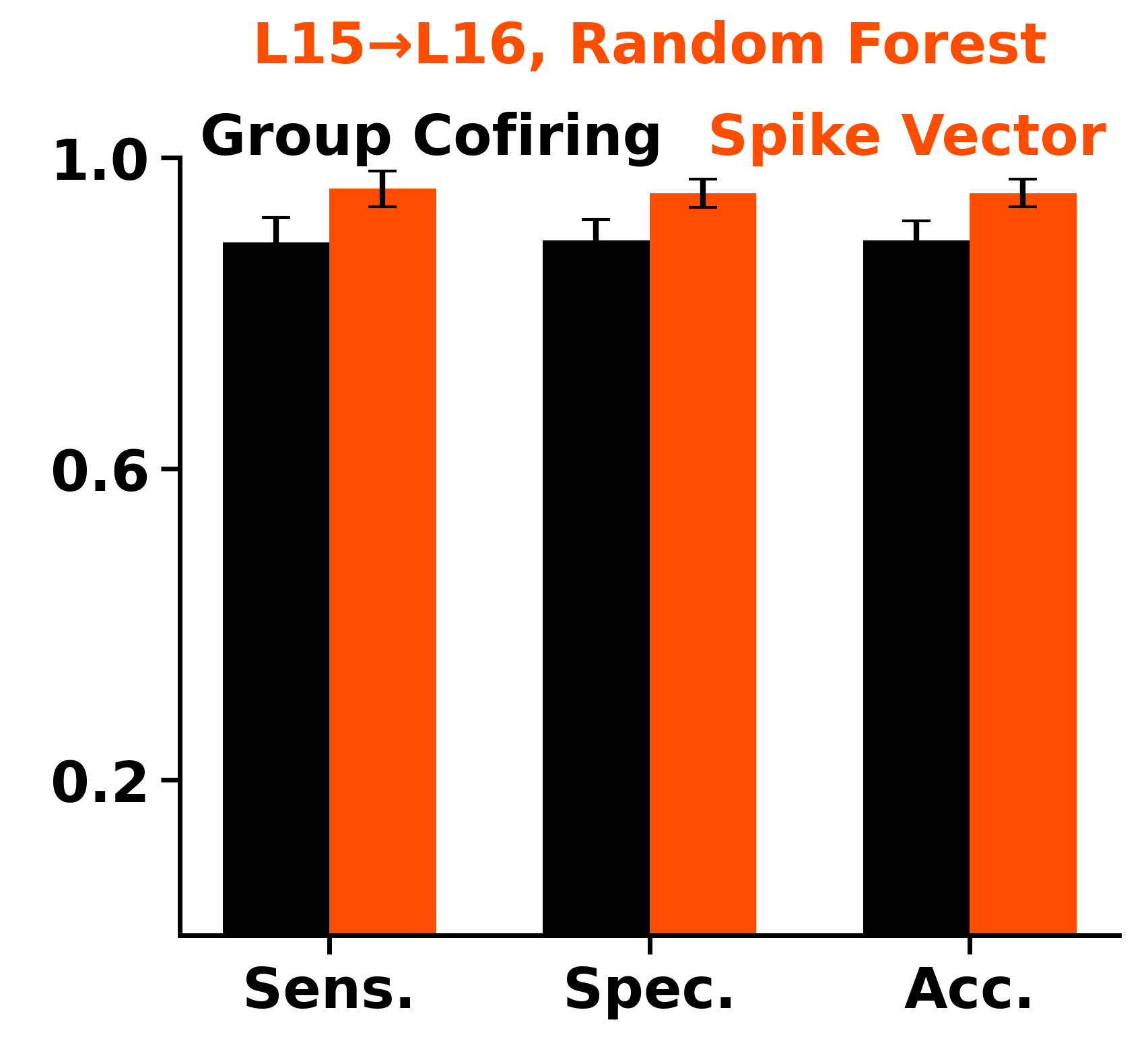}
\put(2,90){\textbf{K}}
\end{overpic} &
\begin{overpic}[width=0.20\linewidth]{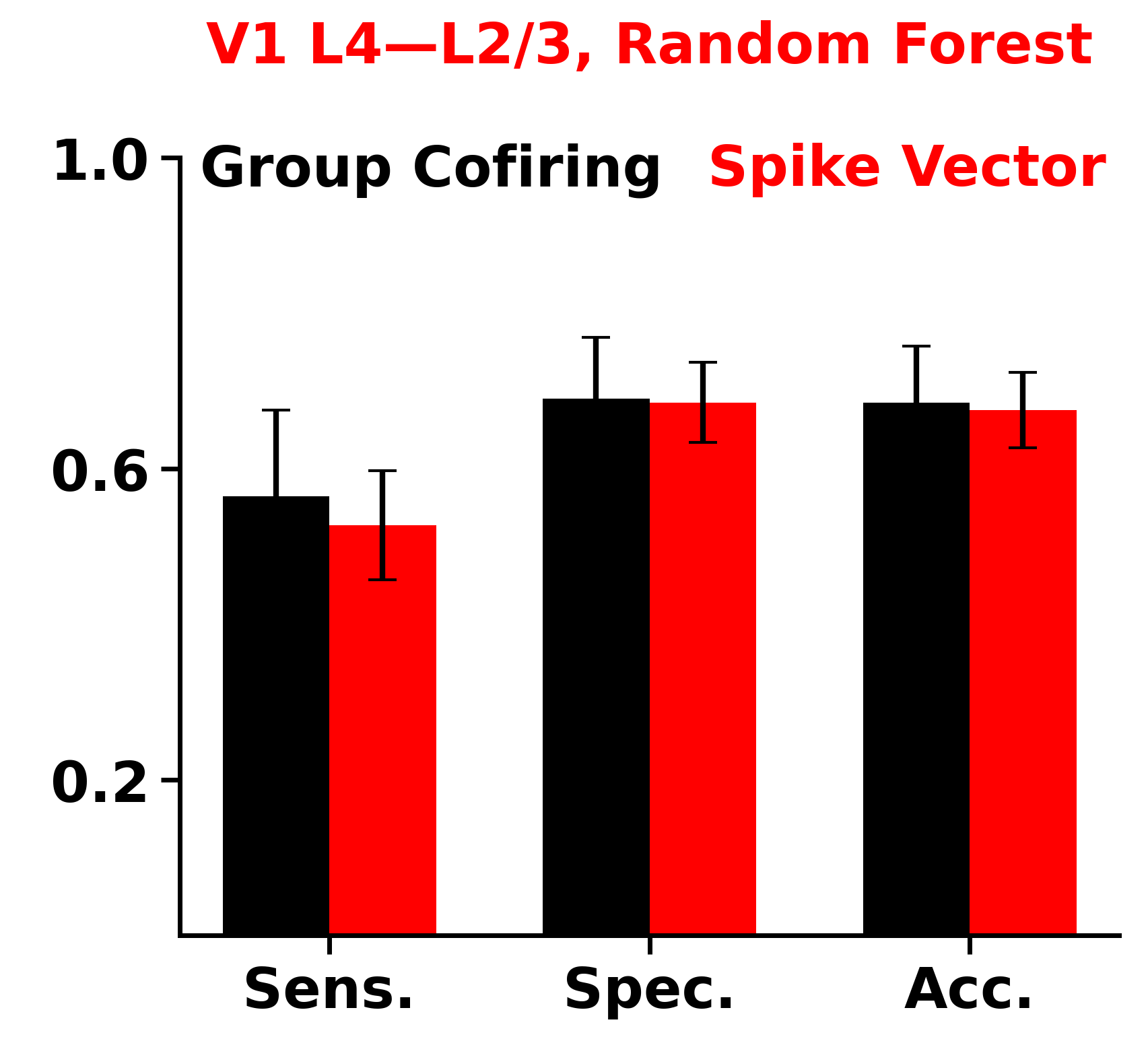}
\put(2,90){\textbf{L}}
\end{overpic}\\

\begin{overpic}[width=0.20\linewidth]{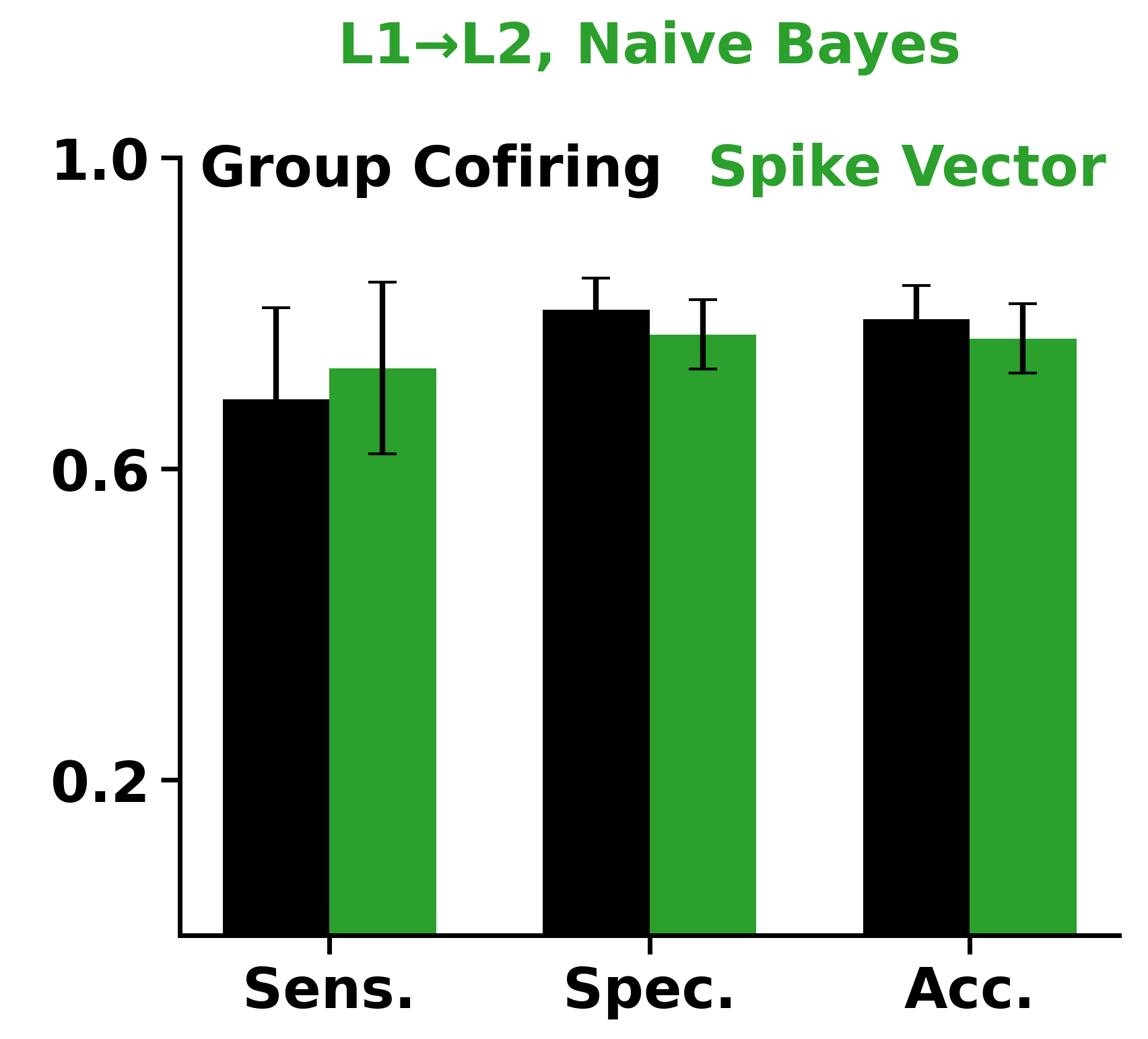}
\put(2,90){\textbf{M}}
\end{overpic} &
\begin{overpic}[width=0.20\linewidth]{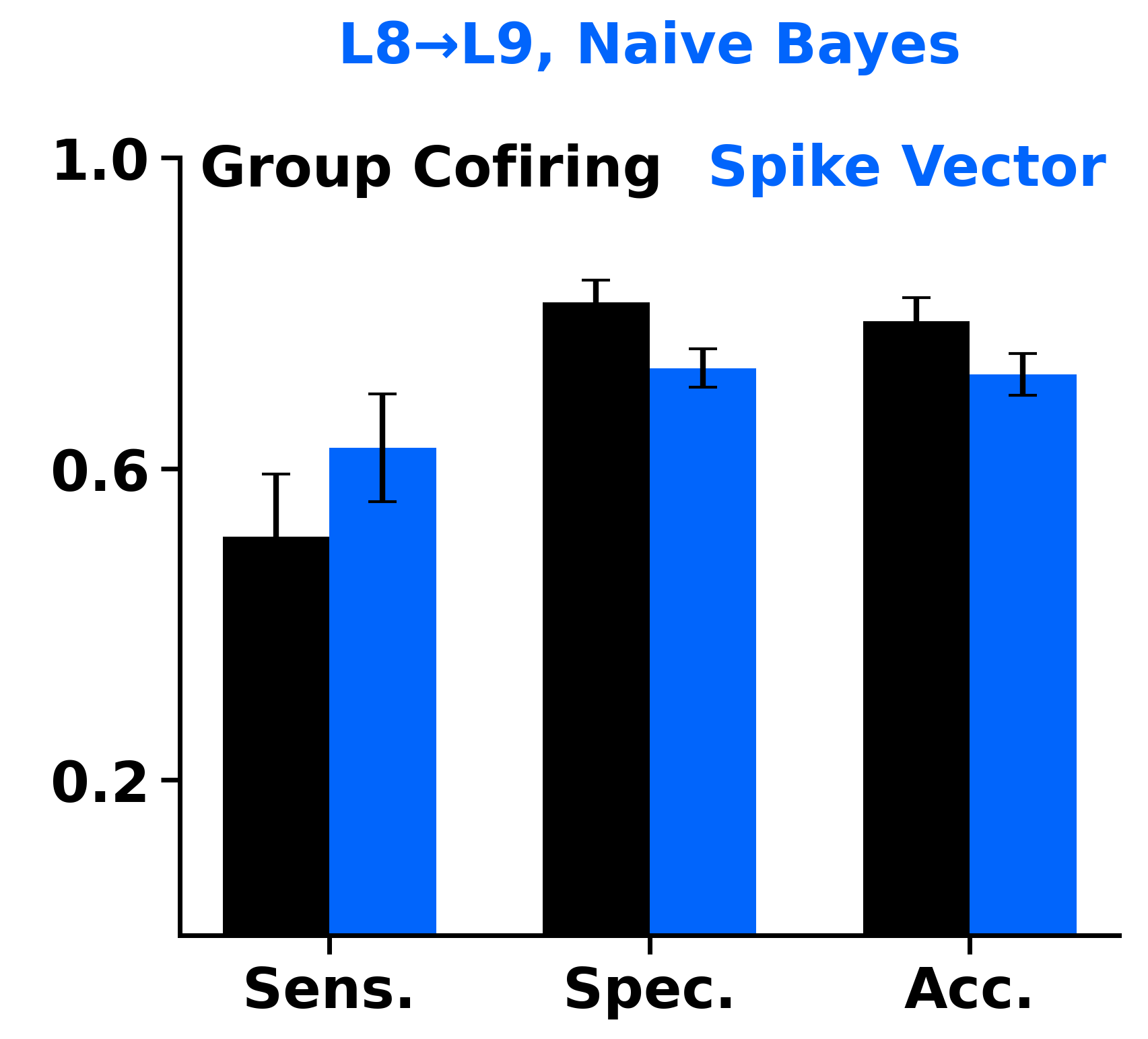}
\put(2,90){\textbf{N}}
\end{overpic} &
\begin{overpic}[width=0.20\linewidth]{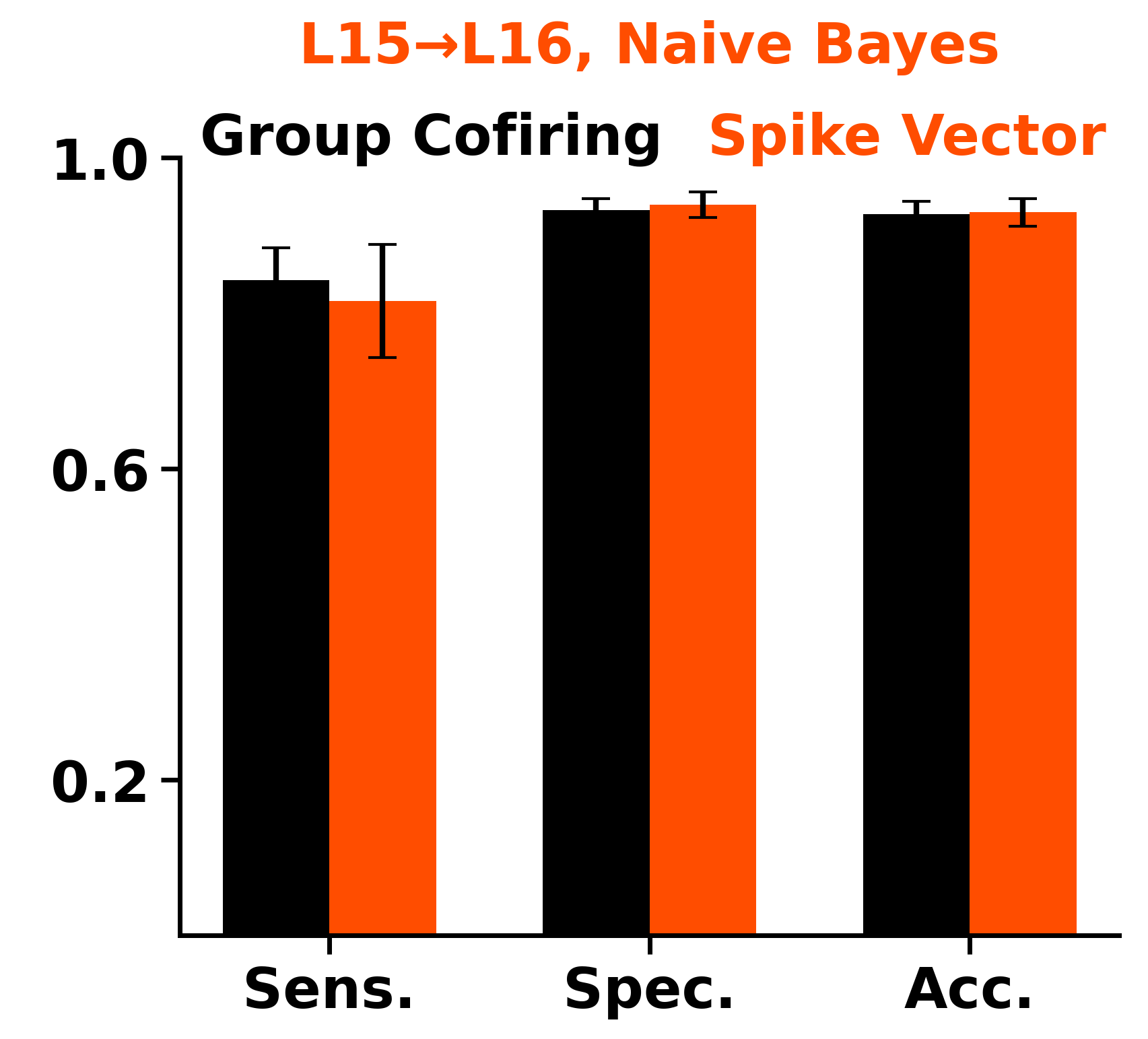}
\put(2,90){\textbf{O}}
\end{overpic} &
\begin{overpic}[width=0.20\linewidth]{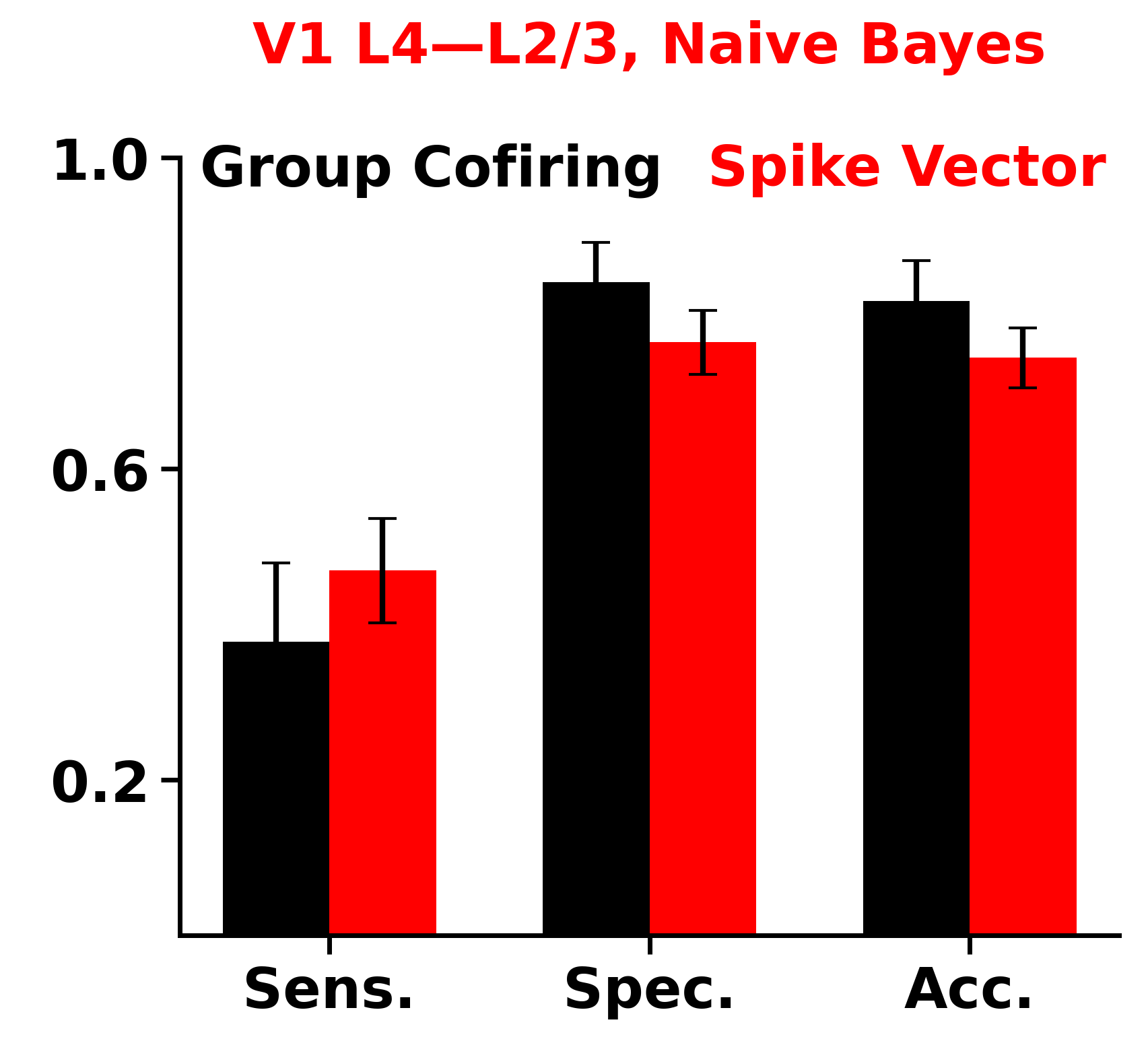}
\put(2,90){\textbf{P}}
\end{overpic}
\end{tabular}
\caption{\textbf{Predicting neural response using its -1FC group}. The knowledge of the identity of the 1FC neighbors of an index neuron, in comparison to the aggregate 1FC cofiring, has a relatively small impact on the classification accuracy in determining the firing of the index neuron. We employ Support Vector Machines \textbf{(A-D)}, Logistic Regression \textbf{(E-H)}, Random Forests \textbf{(I-L)}, and Naive Bayes Classifers \textbf{(M-P)} to predict the inferred firing of the index neuron during the original test dataset using as input: the number of cofiring of its 1FC groups (black), or the individual spike trains of the 1FC group of the index neuron (color; green, blue, orange for layer-pairs L1$\rightarrow$L2, L8$\rightarrow$L9 and L15$\rightarrow$L16 respectively). Neurons with 1FC group sizes < 15 were excluded from this analysis. Note that the input of the individual spike trains (color) includes the identity of the neurons that fire. We compared their performance in terms of sensitivity, specificity, and accuracy.}
\label{sfig:identity_barplots}
\end{figure}

Preserving the identities of individual neurons within input–1FC groups does not provide significant additional predictive value beyond the aggregate co-firing signal when modeling the activity of the corresponding output neuron. In particular, a linear-SVM kernel (Fig. \ref{sfig:identity_barplots} A-D) classifier that explicitly incorporated neuron identity exhibit a relatively small improved performance to models relying solely on total co-firing (consistent results also with Random Forest, Naive Bayes, and Logistic Regression; Fig. \ref{sfig:identity_barplots} E-P). 
These findings suggest that ensemble-level co-firing magnitude may play a more prominent role in shaping downstream responses than the specific identities of the contributing neurons. 

\clearpage
\subsection{Semantic Similarity}
We examined whether neurons preferentially encode classes that are semantically related in human-interpretable terms. To quantify semantic relationships among CIFAR-100 labels, we compute embeddings for each class using CLIP, acontrastive language-image pre-trained model \cite{OpenAI_CLIP},  and define semantic similarity via cosine similarity between embedding vectors. Formally, semantic distance is given by cosine distance, and semantic similarity is defined as $1 -$ cosine distance.

\begin{figure}[t]
\centering
\begin{tabular}{@{}c@{\hspace{0.02\linewidth}}c@{\hspace{0.02\linewidth}}c@{}}
  \begin{minipage}[c]{0.28\linewidth}
    \centering
    \begin{overpic}[width=\linewidth]{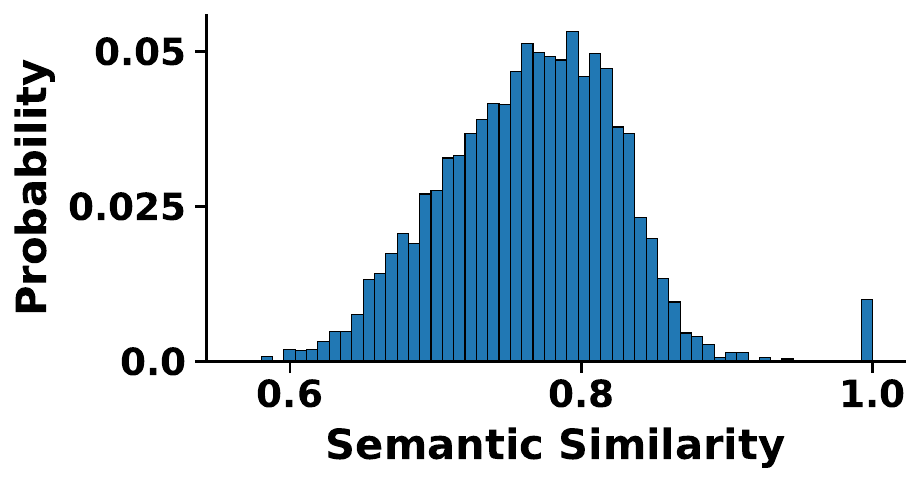}
      \put(1, 55){\textbf{A}}
    \end{overpic}\\[4pt]
    \begin{overpic}[width=\linewidth]{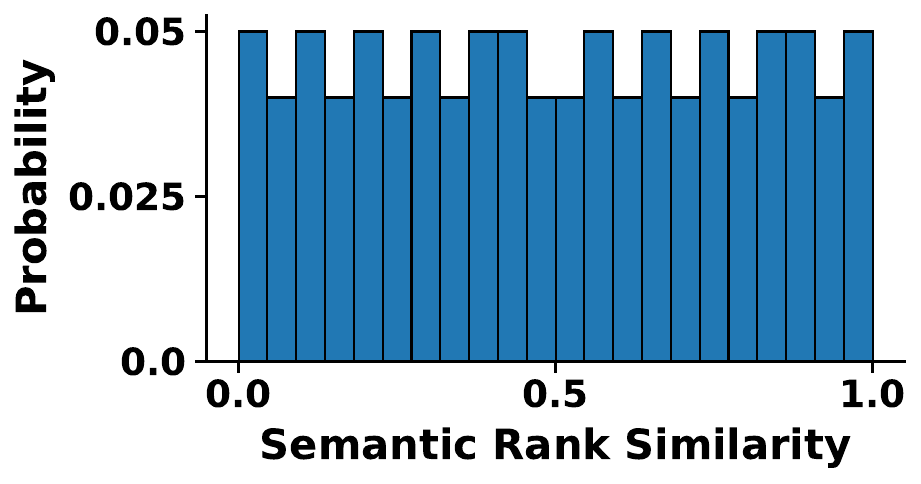}
      \put(1,53){\textbf{B}}
    \end{overpic}
  \end{minipage}
  &
  
  \begin{minipage}[c]{0.32\linewidth}
    \centering
    \begin{overpic}[width=\linewidth]{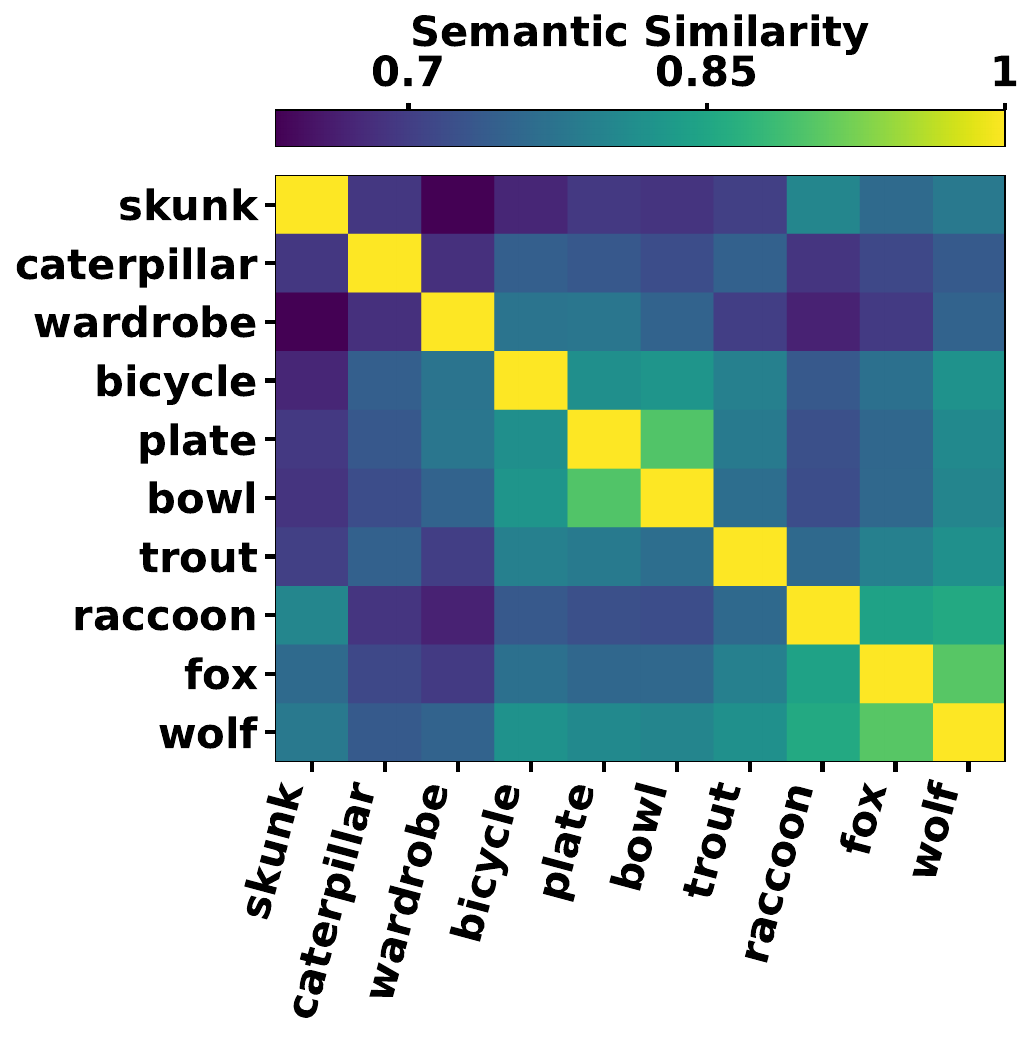}
      \put(2,96){\textbf{C}}
    \end{overpic}
  \end{minipage}
  &
  
  \begin{minipage}[c]{0.32\linewidth}
    \centering
    \begin{overpic}[width=\linewidth]{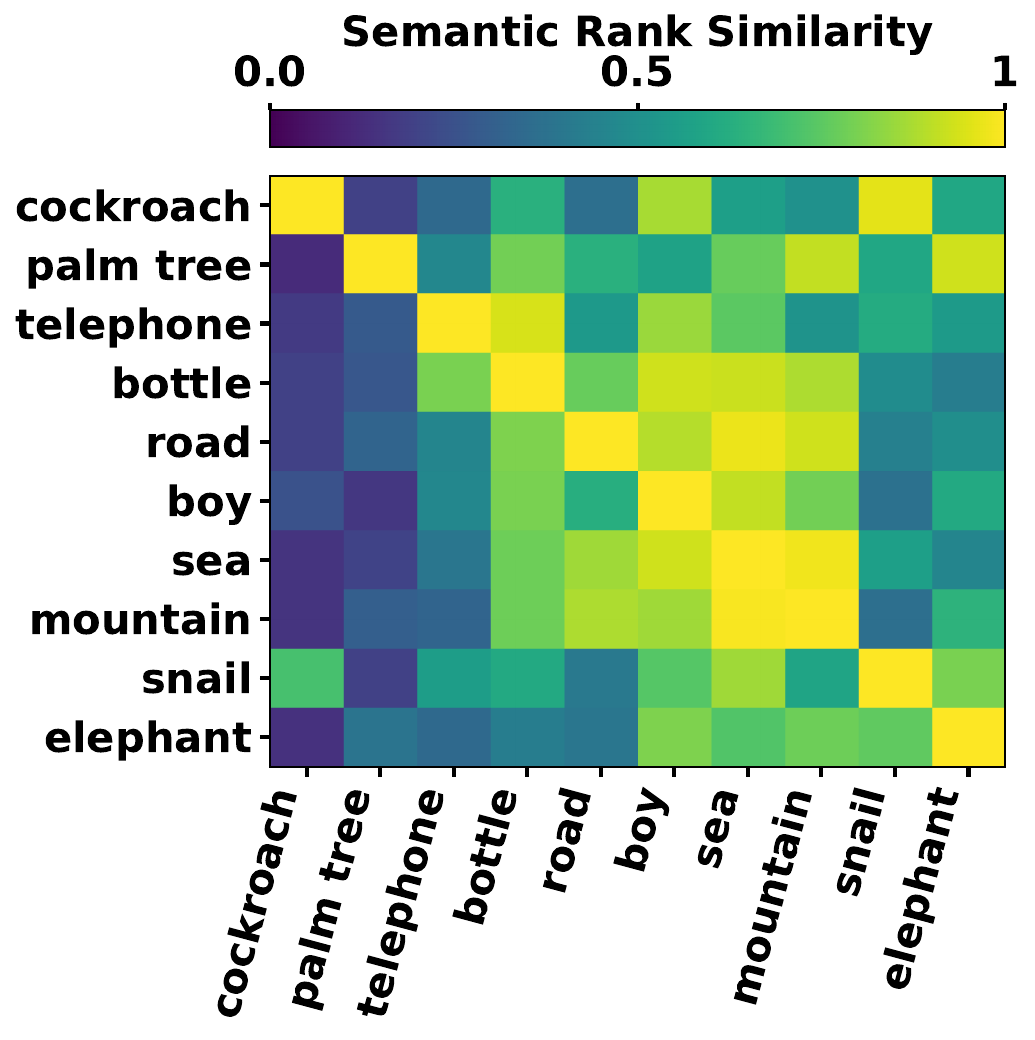}
      \put(2,96){\textbf{D}}
    \end{overpic}
  \end{minipage}
\end{tabular}
\caption{\textbf{Semantic Similarity of CIFAR100 classes} \textbf{(A, B)} Distributions of CLIP-based semantic similarity and semantic rank similarity across CIFAR-100 classes. \textbf{(C, D)} Heatmaps of pairwise semantic similarity and semantic rank similarity for a subset of 10 randomly selected classes. While both representations encode similar relational structure, rank similarity reduces distributional imbalance by emphasizing relative ordering over absolute cosine distances, resulting in a more uniform comparison across classes.}
\label{fig:semantic_rank_sim_metric}
\end{figure}

Because CIFAR-100 class definitions are not uniformly distributed in semantic space, raw cosine similarities exhibit a biased and uneven distribution. To mitigate this, we introduce a semantic rank similarity metric. For each class, all other classes are ranked according to their semantic distance, and ranks are normalized by the total number of classes. This yields a uniform representation of relative semantic proximity, where each class has maximal self-similarity and progressively lower similarity assigned according to rank order rather than absolute distance. (See Fig \ref{fig:semantic_rank_sim_metric} for distributions of semantic similarity and semantic rank similarity.)

\subsection{Additive Noise conditions}\label{sec:add_noise}
We analyze the functional connectivity under two types of additive noise, namely uniform random noise and adversarial noise via targeted input perturbation, based on the Fast Gradient Sign Method (FGSM)\cite{Goodfellow2014ExplainingAH}.

\begin{figure}[H]
\centering
\begin{tabular}{ccc}
\begin{overpic}[width=0.45\linewidth]{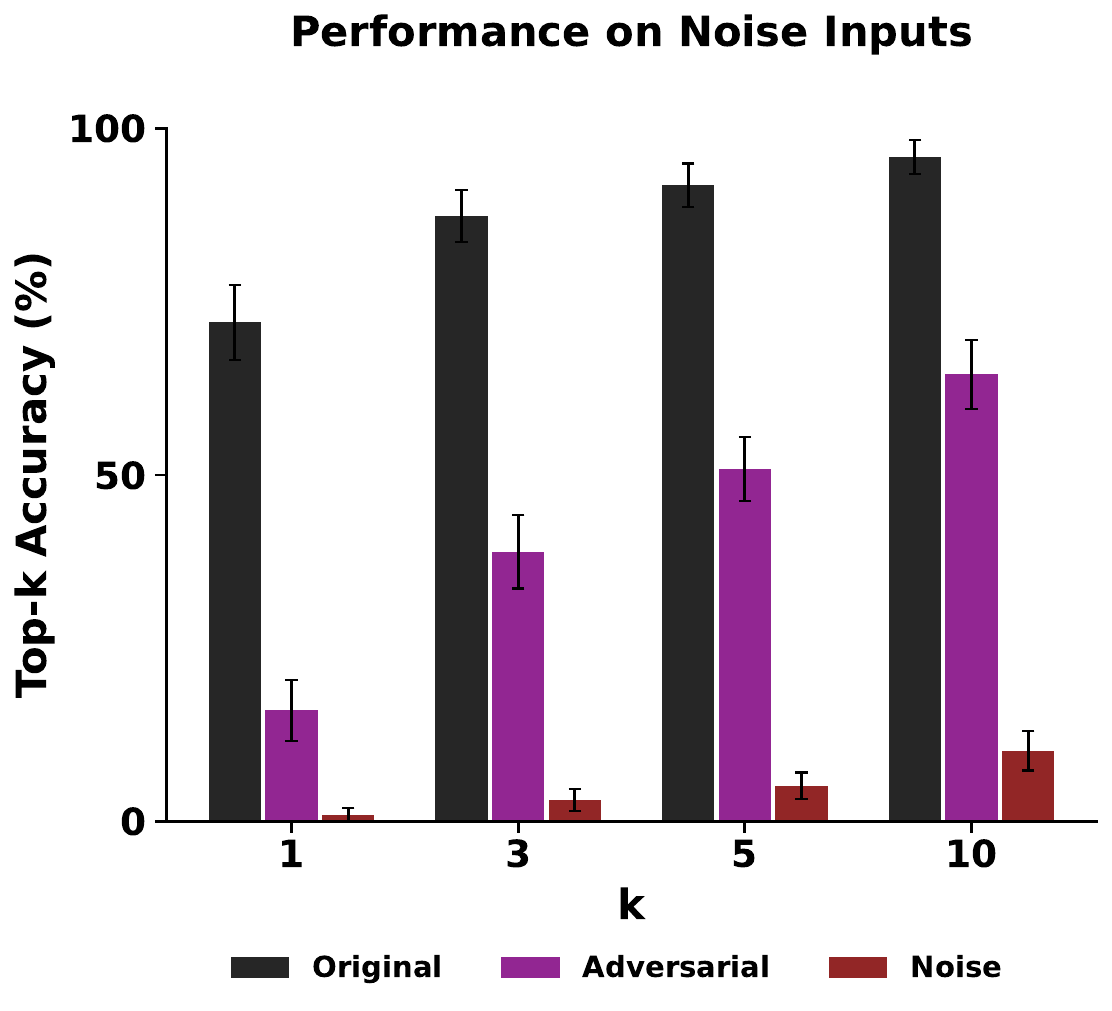}
\put(2,90){\textbf{A}}
\end{overpic} &
\begin{overpic}[width=0.49\linewidth]{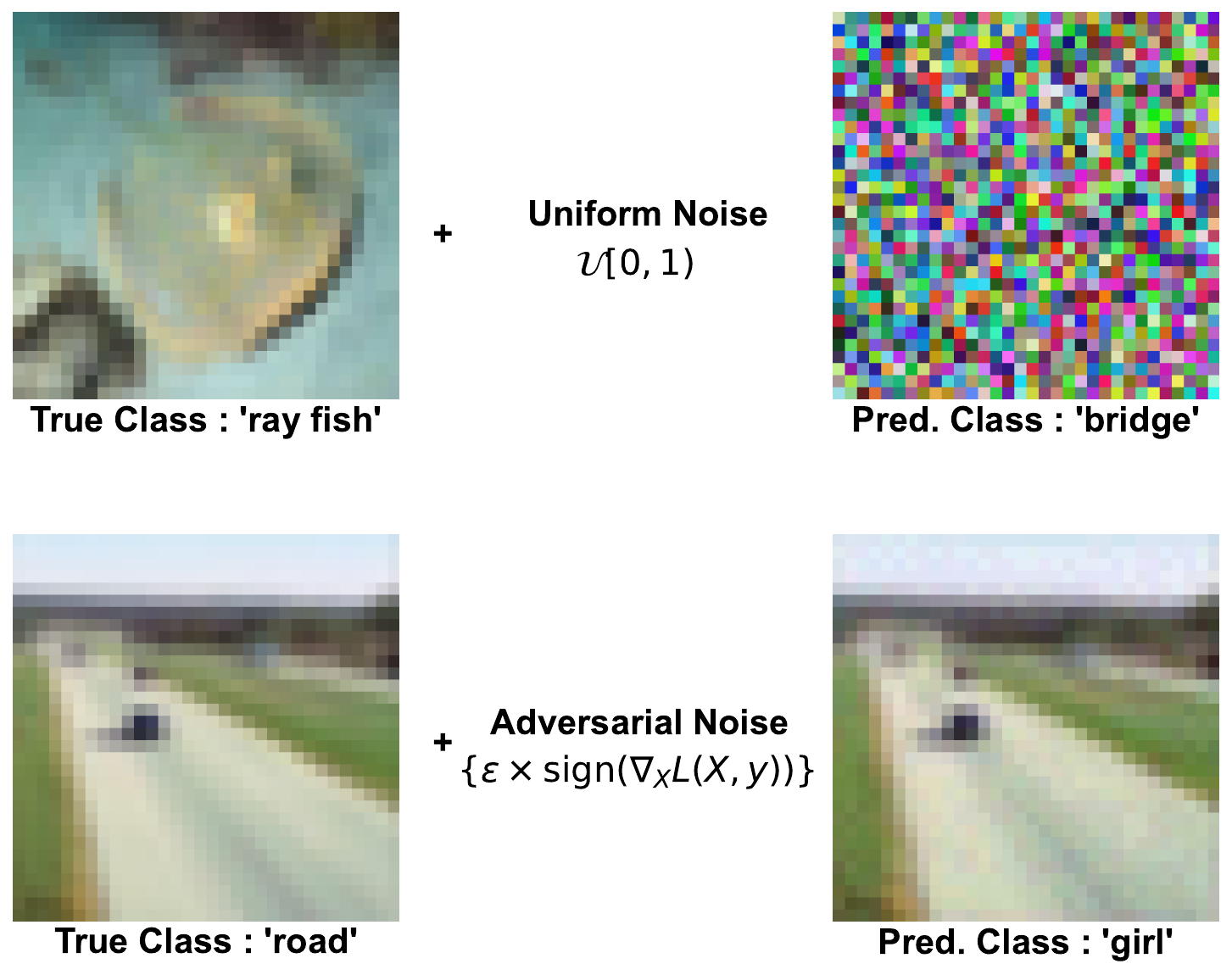}
\put(2,90){\textbf{B}}
\end{overpic} \\
\end{tabular}
\caption{\textbf{Performance of the network under additive noise and adversarial perturbations.} \textbf{(A)} Bars show top-$k$ accuracy: the proportion of samples for which the ground-truth label is contained within the model’s $k$ highest-scoring predictions—for $k = $ 1, 3, 5, and 10, evaluated on original, noise-corrupted, and adversarial inputs. Error bars denote standard deviation across batches of input.  \textbf{(B)} Example test images (“ray fish” and “road”) with added perturbations. Top row: uniform random noise. Bottom row: adversarial noise. The perturbed images are misclassified as “bridge” and “girl,” respectively.}
\label{fig:noise_example_images}
\end{figure}

\begin{figure}[H]
\centering
\begin{overpic}[width=\linewidth]{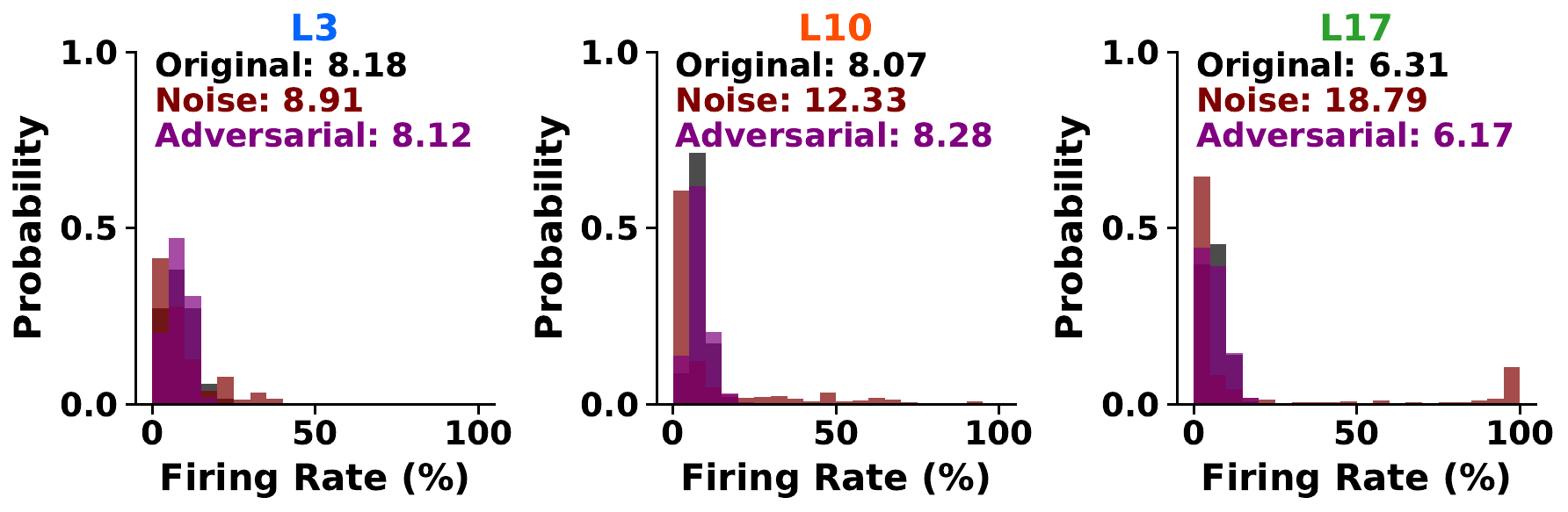}
        \put(4,32){\textbf{A}}
        \put(34,32){\textbf{B}}
        \put(62,32){\textbf{C}}
\end{overpic}
\caption{\textbf{Firing rates under original test, noisy and adversarial input.} \textbf{(A)}. The histogram of the event rate of each neuron in L2 under the noisy test input (maroon), adversarial noise (purple) and original test input (black). \textbf{(B-C)}. Same, for neurons in L9 and L16, respectively. Each panel shows active neurons with firing rate > 0.01 spikes/sample. Inset: Mean event rate, for each layer.}
\label{sfig:noise_firing_rates}
\end{figure}

\begin{figure}
    \centering
    \begin{overpic}[width=\linewidth]{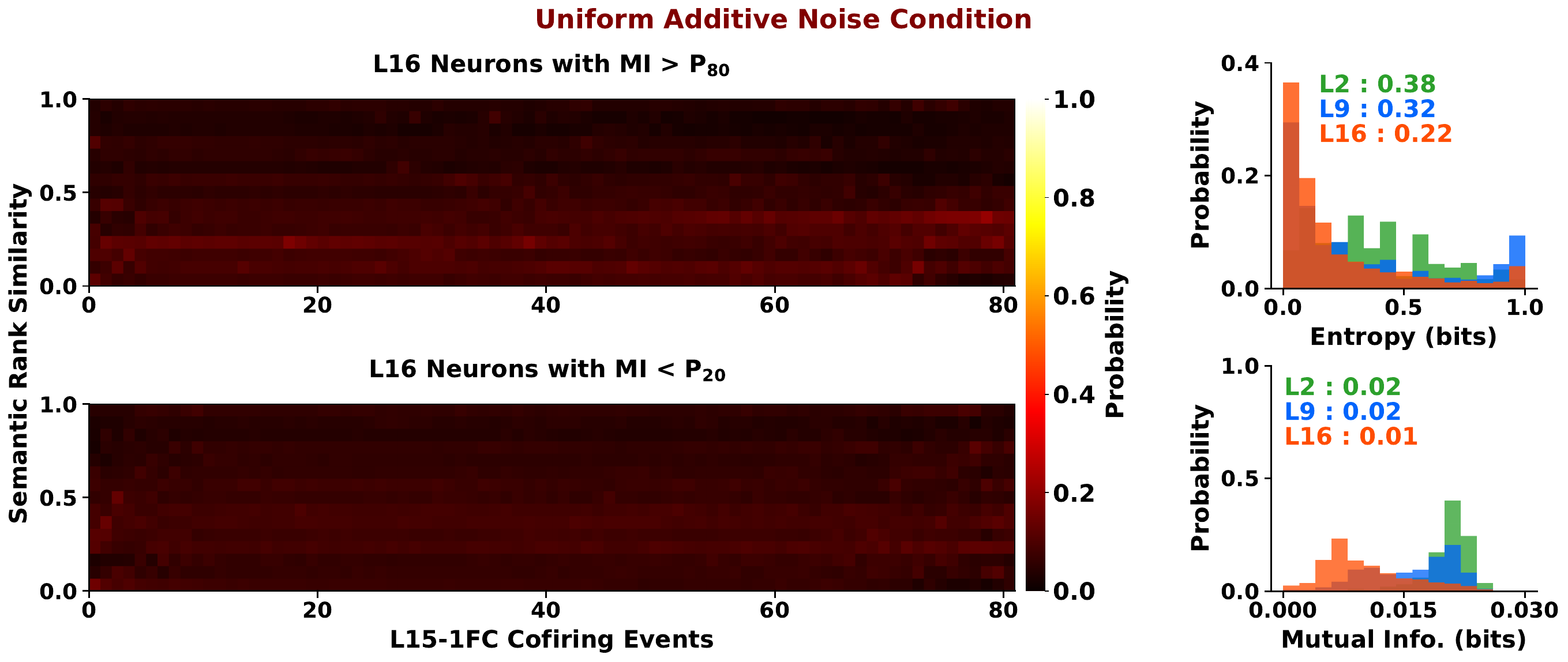}
        \put(3,38){\textbf{A}}
        \put(74,38){\textbf{B}}
        \put(74,19){\textbf{C}}
\end{overpic}
    \caption{\textbf{Information encoding within output neurons and their 1FC groups under Uniform Additive Noise conditions} \textbf{(A)} Top: heatmap of semantic rank similarity for L16 neurons whose class-wise mutual information lies in the top 20th percentile, shown as a function of L15-1FC group co-firing events under uniform additive noise conditions. Each bin reflects the probability that the class of frames with N co-firing events matches the L16 neuron’s preferred class (defined as the class with maximum class-wise mutual information). L15-1FC group identities and L16 neuron's preferred classes are kept fixed from the original input without additive noise. Bottom: same analysis for neurons in the lowest 20th percentile. For each co-firing bin, probabilities are normalized to sum to 1 across semantic rank similarity. \textbf{(B)} Entropy of L2, L9, and L16 spike trains under uniform additive noise conditions. \textbf{(C)} Total mutual information of L2, L9, and L16 neurons with the output under uniform additive noise conditions, quantifying the ability of their spike trains to distinguish among output classes.}
    \label{fig:semantic_heatmaps_with_MI_noise}
\end{figure}

\textbf{Uniform Random Noise.} 
We probed the robustness of the neuronal responses under uniform random noise added to the image. (See  Fig. \ref{fig:noise_example_images})
\begin{equation}
X' = 0.1\times X + 0.9\times\mathcal{U}\left[0, 1\right).   
\end{equation}
Under uniform additive noise, firing rates were substantially lower than under original input conditions (Fig. \ref{sfig:noise_firing_rates}). Information encoding was also altered: although L15-1FC groups of L16 neurons could still exhibit high co-firing events, these events occurred less frequently and were largely associated with images unrelated to the neuron’s preferred class (Fig. \ref{fig:semantic_heatmaps_with_MI_noise} A). Consequently, the spiking activity of L16 neurons exhibited reduced entropy and lower class-specific information content (Fig. \ref{fig:semantic_heatmaps_with_MI_noise} B, C).

\textbf{Adversarial noise under targeted input perturbation.}
We applied the Fast Gradient Sign Method (FGSM) \citet{Goodfellow2014ExplainingAH} adversarial attack to the test set. FGSM generates adversarial examples by perturbing the input in the direction of the gradient of the loss with respect to the input, scaled by a perturbation magnitude $\epsilon$: 
\begin{equation}
X' = X + \epsilon\times \text{sign} (\nabla_XL(X, y)).     
\end{equation}
where, $x$ is the input image, $y$ is the network's prediction for $x$ and $(\nabla_xL(\theta, x, y))$ is the gradient of the loss function with respect to $x$.
Since spiking neural networks are non-differentiable by nature, gradient computation is non-trivial; however, the use of surrogate gradient backpropagation during training, specifically the sigmoid surrogate function, afforded a straightforward extension of FGSM to our SNN, as the surrogate gradients with respect to the input were directly available. Adversarial examples were generated at $\epsilon$ = 4, and neuronal firing probabilities were measured as a function of the 1FC defined using the original test images.

\begin{figure}[H]
    \centering
    \begin{overpic}[width=\linewidth]{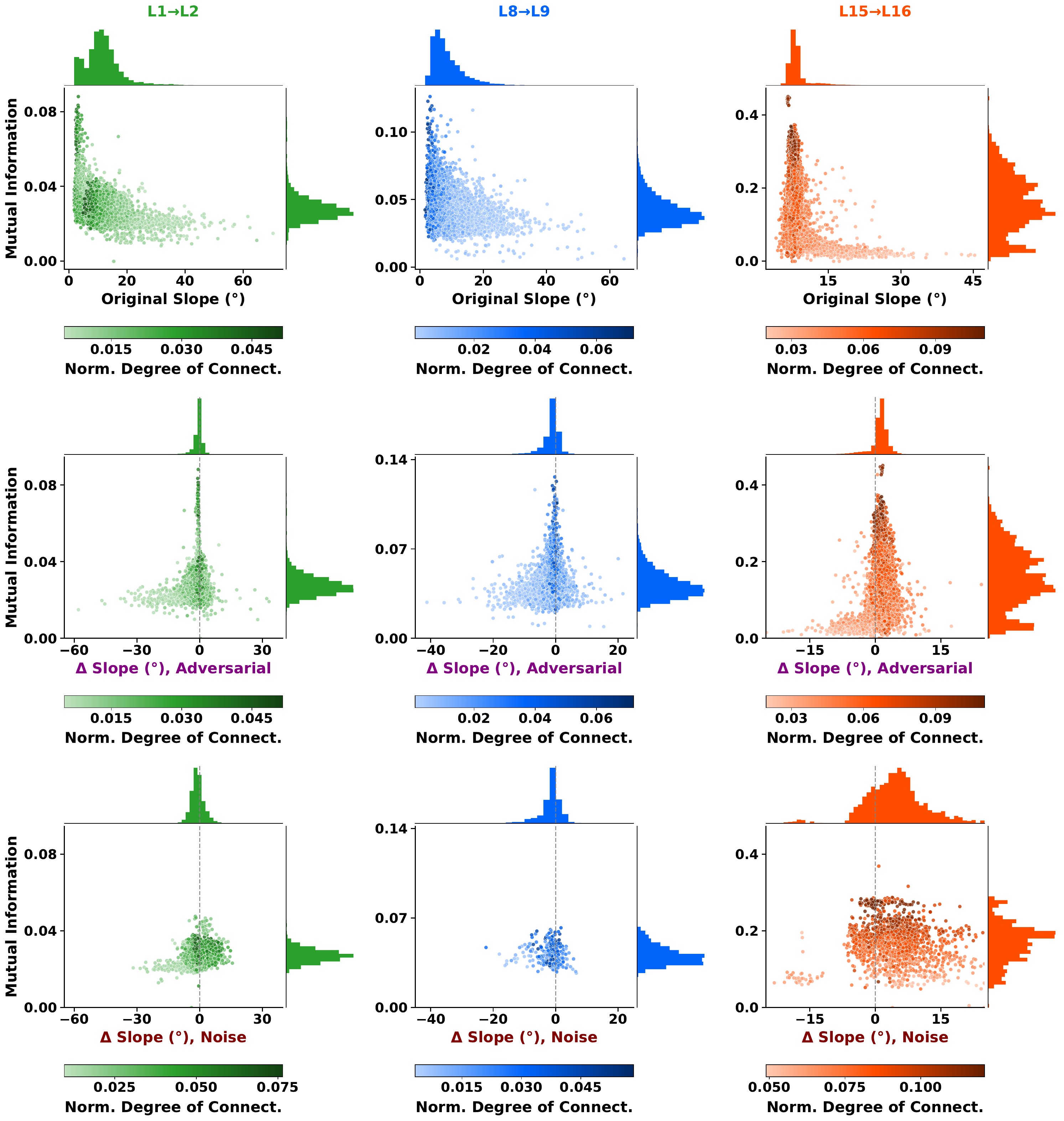}
        \put(3,93){\textbf{A}}
        \put(34,93){\textbf{B}}
        \put(66,93){\textbf{C}}
        \put(3,61){\textbf{D}}
        \put(34,61){\textbf{E}}
        \put(66,61){\textbf{F}}
        \put(3,28){\textbf{G}}
        \put(34,28){\textbf{H}}
        \put(66,28){\textbf{I}}
\end{overpic}
    \caption{\textbf{Differences in the slope of a neuron's response under additive noise input relative to its response under original input.} \textbf{(A–C)} Scatter plots of mutual information (MI) versus the slope of neuronal response functions for layer pairs L1$\rightarrow$L2, L8$\rightarrow$L9, and L15$\rightarrow$L16, respectively, where slope is computed as a function of input-layer 1FC group cofiring events under original input. \textbf{(D-F)} MI versus the change in slope under adversarial noise for the same layer pairs. \textbf{(G-I)} MI versus the change in slope under uniform additive noise. Neurons with lower MI and connectivity exhibit greater sensitivity to noise, whereas higher-MI neurons show more robust (less sensitive) responses. Neurons with response-curve fits yielding $R^2 \leq 0.8$ were excluded from the analysis.}
    \label{fig:MI_Adv_Scatter}
\end{figure}

\begin{figure}
    \centering
    \begin{overpic}[width=\linewidth]{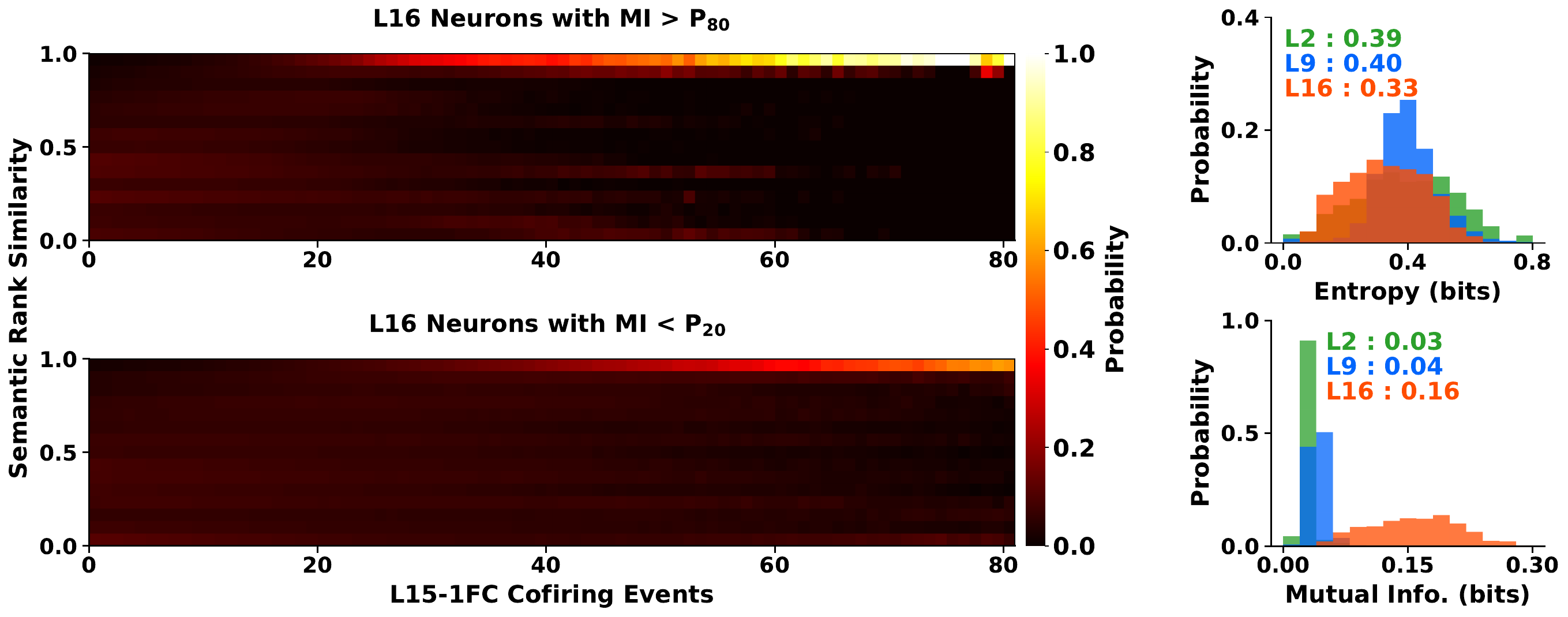}
        \put(3,38){\textbf{A}}
        \put(74,38){\textbf{B}}
        \put(74,19){\textbf{C}}
\end{overpic}
    \caption{\textbf{Information encoding within output neurons and their 1FC groups} \textbf{(A)} Top: heatmap of semantic rank similarity for L16 neurons whose class-wise mutual information lies in the top 20th percentile, shown as a function of L15-1FC group co-firing events. Each bin reflects the probability that the class of frames with N co-firing events matches the L16 neuron’s preferred class (defined as the class with maximum class-wise mutual information). Bottom: same analysis for neurons in the lowest 20th percentile. For each co-firing bin, probabilities are normalized to sum to 1 across semantic rank similarity. \textbf{(B)} Entropy of L2, L9, and L16 spike trains. \textbf{(C)} Total mutual information of L2, L9, and L16 neurons with the output, quantifying the ability of their spike trains to distinguish among output classes.}
    \label{fig:semantic_heatmaps_with_MI_original}
\end{figure}

\begin{figure}
    \centering
    \begin{overpic}[width=\linewidth]{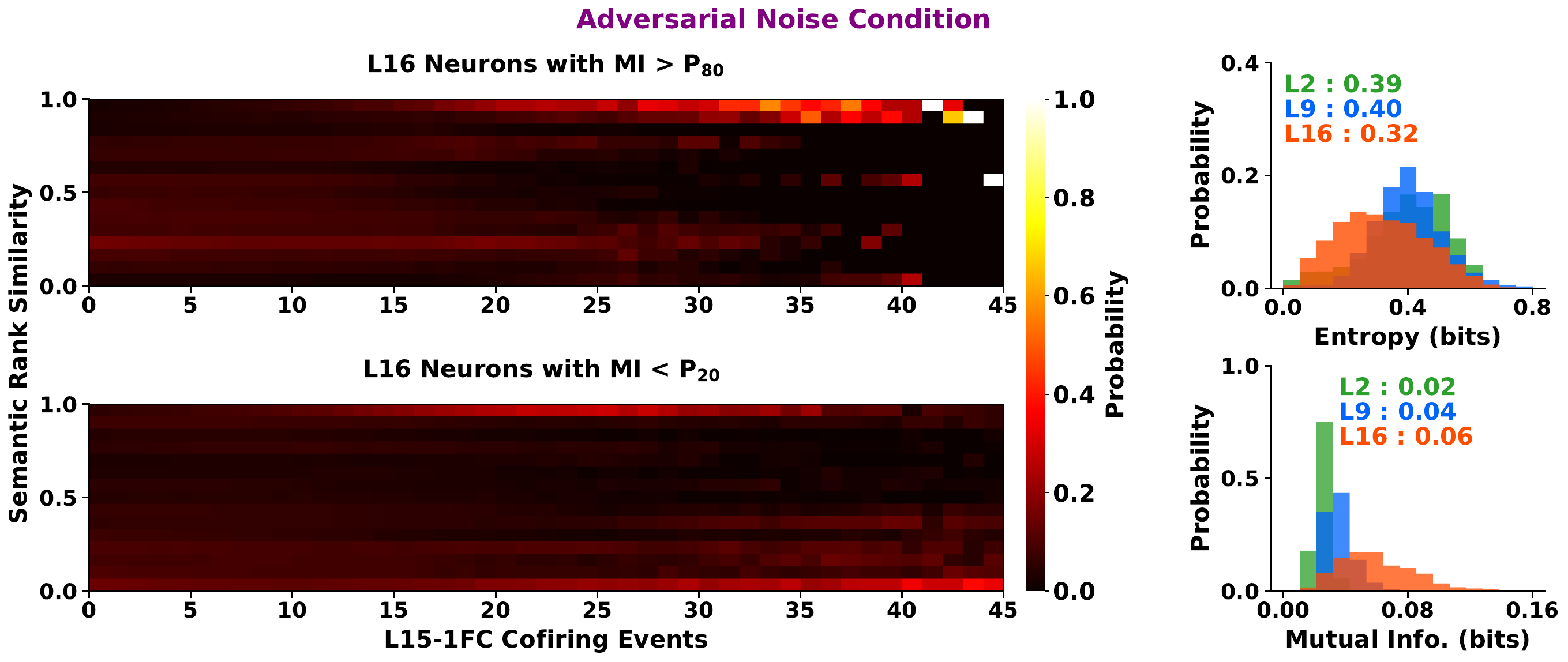}
        \put(3,38){\textbf{A}}
        \put(74,38){\textbf{B}}
        \put(74,19){\textbf{C}}
\end{overpic}
    \caption{\textbf{Information encoding within output neurons and their 1FC groups under Adversarial Noise conditions} \textbf{(A)} Top: heatmap of semantic rank similarity for L16 neurons whose class-wise mutual information lies in the top 20th percentile, shown as a function of L15-1FC group co-firing events under adversarial noise conditions. Each bin reflects the probability that the class of frames with N co-firing events matches the L16 neuron’s preferred class (defined as the class with maximum class-wise mutual information). L15-1FC group identities and L16 neuron's preferred classes are kept fixed from the original input without adversarial noise. Bottom: same analysis for neurons in the lowest 20th percentile. For each co-firing bin, probabilities are normalized to sum to 1 across semantic rank similarity. \textbf{(B)} Entropy of L2, L9, and L16 spike trains under adversarial noise conditions. \textbf{(C)} Total mutual information of L2, L9, and L16 neurons with the output under adversarial noise conditions, quantifying the ability of their spike trains to distinguish among output classes.}
    \label{fig:semantic_heatmaps_with_MI_adv}
\end{figure}

\begin{figure}
    \centering
    \begin{overpic}[width=0.6\linewidth]{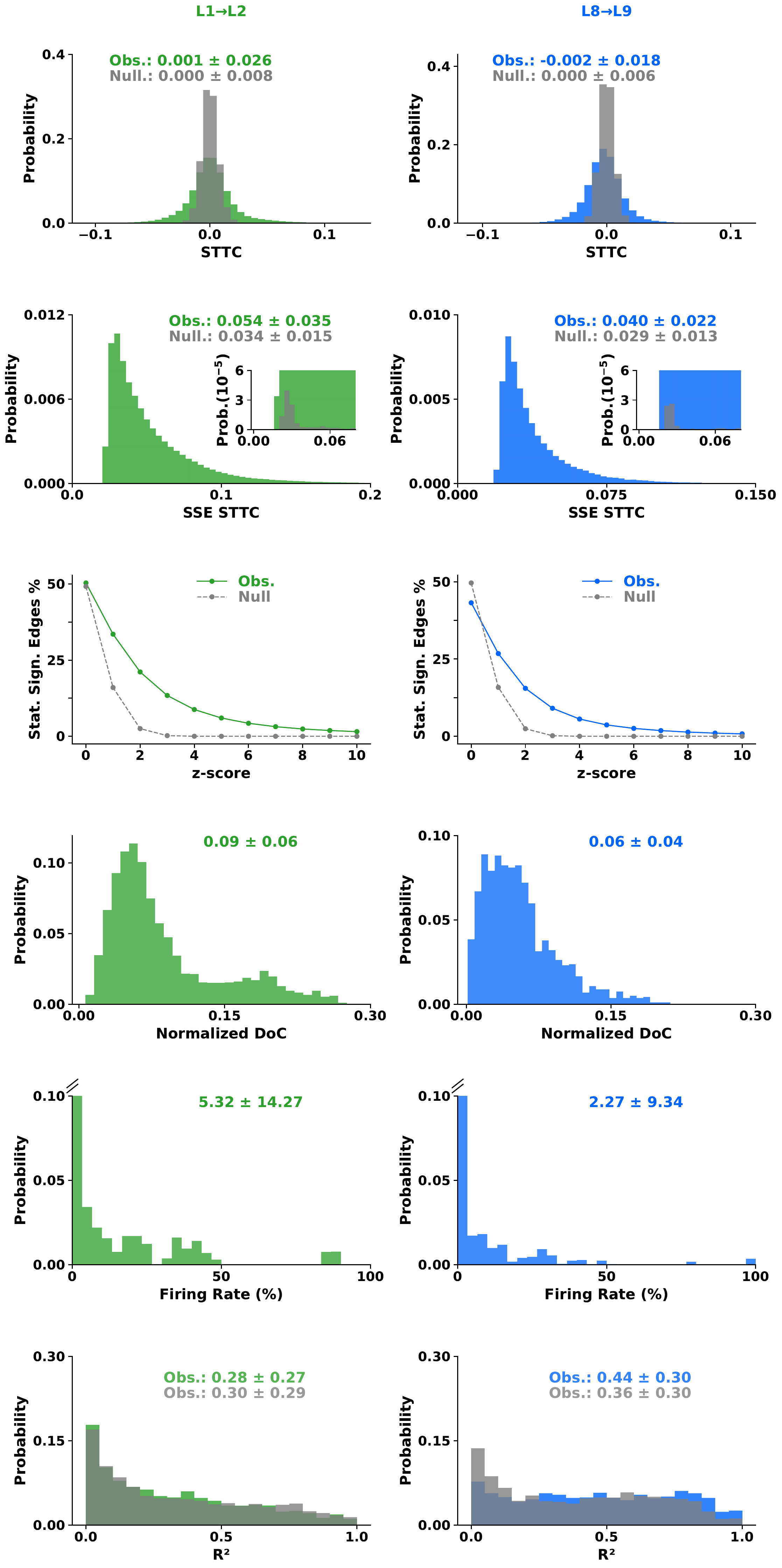}
        \put(1,98){\textbf{A}}
        \put(25,98){\textbf{B}}
        \put(1,81){\textbf{C}}
        \put(25,81){\textbf{D}}
        \put(1,64){\textbf{E}}
        \put(25,64){\textbf{F}}
        \put(1,48){\textbf{G}}
        \put(25,48){\textbf{H}}
        \put(1,31){\textbf{I}}
        \put(25,31){\textbf{J}}
        \put(1,15){\textbf{K}}
        \put(25,15){\textbf{L}}
\end{overpic}
    \caption{\textbf{Inter-layer functional connectivity under shuffled-weight conditions.} Pairwise STTC was computed between neurons in consecutive layers during inference on CIFAR100 using the trained network with shuffled weights. Columns correspond to layer pairs L2$\rightarrow$L3 (green) and L9$\rightarrow$L10 (blue). A null distribution (grey) was generated via 150 circular shifts of spike trains. \textbf{(A, B)} STTC distributions with corresponding nulls. \textbf{(C, D)} Distributions of statistically significant positive edges (z > 4). \textbf{(E, F)} Fraction of significant edges as a function of z-score. \textbf{(G, H)} Degree of connectivity distributions (normalized by total possible edges). \textbf{(I, J)} Firing rate distributions of active neurons (input and output combined). \textbf{(K, L)} Distributions of $R^2$ values from response-function fits. Neurons were considered active if firing rate > 0.01 spikes/sample (yielding $<40\%$ and $<10\%$ active neurons across layers).  z-scores were computed relative to the circular-shift null distributions.}
    \label{fig:shuffled_weights}
\end{figure}
\subsection{Limitations}
So far we tested this approach in
ResNet18 SNNs and spiking-neural units of spiking-layers that extend a ResNet32 feature extractor\cite{He2016:Deep}, pre-trained on CIFAR100 and ResNet9 SNNs trained on DVS Gesture Dataset. We also have only tested the framework for the Fast Gradient Sign Method (FGSM) adversarial attack\cite{Goodfellow2014ExplainingAH} and additive noise. It is part of our ongoing effort to test it with other architectures, activation functions, and hyperparameters. The impact of T=4 (in the image presentation of the SNN model) on STTC will be further examined.


Assessing the accuracy of the diagnostics for test datasets of different size, distribution of class images, diversity/richness of the images, types of distortion (e.g., noise levels, blurriness, adversarial attacks) is part of our ongoing research.
We plan to explore the required size of test set (batch size) and size of the population of neurons to be sampled for detecting reliably when abnormalities in the input or in the behaviour of the architecture occur.

Going forward, integrating this framework with causal perturbations will be key to clarifying the mechanistic role of specific ensembles in learning, as well as to probing how these mechanisms fail under adversarial conditions. 
\subsection{Computational Efficiency}
One of the most computationally intensive steps is estimating the statistical significance of pairwise STTC correlations for all neuron pairs (considered in the analysis), as it requires generating 150 circularly shifted null spike trains to compute the control STTC weight.

\textbf{Compute Resources.} Model training was performed on local GPU servers and the CINECA Leonardo Booster HPC facility (NVIDIA A100 GPUs). Training the SEW-ResNet-18 for 320 epochs on CIFAR-100 required approximately 1.5 GPU hours on 1 GPU.
Post-training analyses were considerably more computationally demanding. Pairwise STTC computation is memory-intensive despite algorithmic optimization via vectorization \cite{aravind2024sttcpy}. It achieves near-linear time complexity $\mathcal{O}(N)$, where \textit{N }is the length of the spike trains being compared; statistical validation further requires 150 circular shift controls per neuron pair. Computing STTC and all controls across all examined layers for a single model required approximately 1 GPU hour on the Leonardo Booster. Mutual information estimation and piecewise linear response function fitting, both performed at the single-neuron level across the sampled population of $\sim$8,000 neurons per layer, were executed on local servers and required approximately 2–3 hours per layer. All inference passes for spike train recording, including clean, noisy, and adversarial conditions, were performed on 1 GPU with a total inference time of approximately <10 minutes for the full test set. 
\subsection{Broader Implications}
A potential implication of this work involves the design of diagnostic frameworks for spiking neural networks (SNNs), for assessing the sequence of input images for testing/inference.

In general, by emulating fundamental features of biological computation—such as sparse and event-driven activity—SNN models offer a compelling paradigm for efficient information processing, which neuromorphic hardware systems exploit to deliver substantial gains in energy efficiency. Moreover, the shared features of functional connectivity between SNNs and the mouse visual cortex may enable the use of SNNs as testbeds for evaluating neurocomputational hypotheses about cortical connectivity under diverse conditions.
\clearpage

\end{document}